\documentclass[letterpaper, 11 pt, journal, onecolumn]{ieeeconf}

\usepackage{graphicx} 
\usepackage{xcolor} 
\usepackage{verbatim}
\usepackage{cite}
\usepackage{booktabs}

\RequirePackage{import}

\makeatletter
\let\NAT@parse\undefined
\makeatother
\usepackage{hyperref}
\usepackage{fancyhdr}

\graphicspath{ {images/} }

\title{
\textbf{Semantics for Robotic Mapping, Perception and Interaction: A Survey}}

\author{Sourav Garg$^{1*}$, Niko S\"underhauf$^{1}$, Feras Dayoub$^{1}$, Douglas Morrison$^{1}$, \\Akansel Cosgun$^{2}$, Gustavo Carneiro$^{3}$, Qi Wu$^{3}$, Tat-Jun Chin$^{3}$, \\Ian Reid$^{3}$, Stephen Gould$^{4}$, Peter Corke$^{1}$, Michael Milford$^{1}$ 
\thanks{$^{1}$QUT Centre for Robotics and School of Electrical Engineering and Robotics, Queensland University of Technology, Brisbane, Australia}%
\thanks{$^{2}$Department of Electrical and Computer Systems Engineering, Monash University, Melbourne, Australia}%
\thanks{$^{3}$School of Computer Science, University of Adelaide, Adelaide, Australia}%
\thanks{$^{4}$College of
Engineering and Computer Science, Australian National University, Canberra, Australia}%
\thanks{All authors are with the Australian Research Council (ARC) Centre of Excellence for Robotic Vision, Australia (Grant: CE140100016)}%
\thanks{$^{*}$Corresponding Author, {\tt\small Email: s.garg@qut.edu.au}}%
}

\begin{document}

\begin{titlepage}
\maketitle
\thispagestyle{fancy}
\setlength{\headheight}{40pt}
\chead{This is the author's version of an article that has been published in Foundations and Trends\textsuperscript{\textregistered} in Robotics (2020), Vol. 8: No. 1–2, pp 1-224. \\ \textcolor{blue}{The final version of record is available at \url{http://dx.doi.org/10.1561/2300000059}}}
\cfoot{}
\end{titlepage}

\pagestyle{plain}
\setlength{\headheight}{12pt}
\setlength{\textheight}{696pt}

\setcounter{tocdepth}{4}
\tableofcontents

\maketitle
\begin{abstract}

For robots to navigate and interact more richly with the world around them, they will likely require a deeper \textit{understanding} of the world in which they operate. In robotics and related research fields, the study of understanding is often referred to as semantics, which dictates \textit{what does the world `mean' to a robot}, and is strongly tied to the question of \textit{how to represent that meaning}. With humans and robots increasingly operating in the same world, the prospects of human-robot interaction also bring \textit{semantics and ontology of natural language} into the picture. Driven by need, as well as by enablers like increasing availability of training data and computational resources, semantics is a rapidly growing research area in robotics. The field has received significant attention in the research literature to date, but most reviews and surveys have focused on particular aspects of the topic: the technical research issues regarding its use in specific robotic topics like mapping or segmentation, or its relevance to one particular application domain like autonomous driving. A new treatment is therefore required, and is also timely because so much relevant research has occurred since many of the key surveys were published. This survey paper therefore provides an overarching snapshot of where semantics in robotics stands today. We establish a taxonomy for semantics research in or relevant to robotics, split into four broad categories of activity, in which semantics are extracted, used, or both. Within these broad categories we survey dozens of major topics including fundamentals from the computer vision field and key robotics research areas utilizing semantics, including mapping, navigation and interaction with the world. The paper also covers key practical considerations, including enablers like increased data availability and improved computational hardware, and major application areas where semantics is or is likely to play a key role. In creating this survey, we hope to provide researchers across academia and industry with a comprehensive reference that helps facilitate future research in this exciting field.
\end{abstract}

\section{INTRODUCTION}
For robots to move beyond the niche environments of fulfilment warehouses, underground mines and manufacturing plants into widespread deployment in industry and society, they will need to understand the world around them. Most mobile robot and drone systems deployed today make relatively little use of explicit higher level ``meaning'' and typically only consider geometric maps of environments, or three dimensional models of objects in manufacturing and logistics contexts. Despite considerable success and uptake to date, there is a large range of domains with few, if any, commercial robotic deployments; for example: aged care and assisted living facilities, autonomous on-road vehicles, and drones operating in close proximity to cluttered, human-filled environments. Many challenges remain to be solved, but we argue one of the most significant is simply that robots will need to better \textit{understand} the world in which they operate, in order for them to move into useful and safe deployments in more diverse environments. This need for \textit{understanding} is where \textit{semantics} meet robotics.

Semantics is a widely used term, not just in robotics but across fields ranging from linguistics to philosophy. In the robotics domain, despite widespread \textit{usage} of semantics, there is relatively little formal definition of what the term means. In this survey, we aim to provide a taxonomy rather than specific definition of semantics, and note that the surveyed research exists along a spectrum from traditional, non-semantic approaches to those which are primarily semantically-based. Broadly speaking, we can consider semantics in a robotics context to be about the \textit{meaning} of things: the meaning of places, objects, other entities occupying the environment, or even the language used in communicating between robots and humans or between robots themselves.  There are several areas of relevance to robotics where semantics have been a strong focus in recent years, including SLAM (Simultaneous Localization And Mapping), segmentation and object recognition.

Given the importance of semantics for robotics, how can they be equipped, or learn about meaning in the world? There are multiple methods, but they can be split into two categories. Firstly, \textit{provided semantics} describes situations where the robot is given the knowledge beforehand. \textit{Learnt semantics} describes situations where the robot, either beforehand or during deployment, learns this information. The learning mechanism leads to further sub categorisation: learning from observation of the entities of interest in the environment, and actual interaction with the entities of interest, such as through manipulation. Learning can occur in a supervised, unsupervised or semi-supervised manner.

Semantics as a research field in robotics has grown rapidly in the past decade. This growth has been driven in part by the opportunity to use semantics to improve the capabilities of robotic systems in general, but several other factors have also contributed. The popularization of deep learning over the past decade has facilitated much of the research in semantics for robotics, enabling capabilities like high performance general object recognition -- \textit{object class} is an important and useful ``meaning'' associated with the world. Increases in dataset availability, access to the cloud and compute resources have also been critical, providing a much richer source of information from which robots can learn, and the computational power with which to do so, rapidly and at scale. Given the popularity of the field, it has been the focus of a number of key review and survey papers, which we cover here. 

\subsection{Past Coverage Including Survey and Review Papers}

As one of the core fields of research in robotics, SLAM has been the subject of a number of reviews, surveys and tutorials over the past few decades, including coverage of semantic concepts. A recent review of SLAM by Cadena \textit{et al.}~\cite{cadena2016past} positioned itself and other existing surveys as belonging to either classical~\cite{durrant2006simultaneous, bailey2006simultaneous, thrun2002probabilistic, stachniss2016simultaneous}, algorithmic-analysis~\cite{dissanayake2011review} or the ``robust perception'' age~\cite{cadena2016past}. This review highlighted the evolution of SLAM, from its earlier focus on probabilistic formulations and analyses of fundamental properties to the current and growing focus on robustness, scalablity and high-level understanding of the environment - where semantics comes to the fore. Also discussed were robustness and scalability in the context of long-term autonomy and how the underlying representation models, metric \textit{and} semantic, shape the SLAM problem formulation. For the high-level semantic representation and understanding of the environment, \cite{cadena2016past} discussed literature where SLAM is used for inferring semantics, and semantics are used to improve SLAM, paving the way for a joint optimization-based semantic SLAM system. They further highlight the key challenges of such a system, including consistent fusion of semantic and metric information, task-driven semantic representations, dynamic adaptation to changes in the environment, and effective exploitation of semantic reasoning.

Kostavelis and Gasteratos~\cite{kostavelis2015semantic} reviewed semantic mapping for mobile robotics. They categorized algorithms according to scalability, inference model, temporal coherence and topological map usage, while outlining their potential applications and emphasizing the role of human interaction, knowledge representation and planning. More recently, \cite{davison2018futuremapping} reviewed current progress in SLAM research, proposing that SLAM including semantic components could evolve into ``Spatial AI'', either in the form of autonomous AI, such as a robot, or as Intelligent Augmentation (IA), such as in the form of an AR headset. A Spatial AI system would not only capture and represent the scene intelligently but also take into account the embodied device's constraints, thus requiring joint innovation and co-design of algorithms, processors and sensors. More recently, \cite{davison2019futuremapping} presented Gaussian Belief Propagation (GBP) as an algorithmic framework suited to the needs of a Spatial AI system, capable of delivering high performance despite resource constraints. The proposals in both~\cite{davison2018futuremapping} and \cite{davison2019futuremapping} are significantly motivated by rapid developments in processor hardware, and touch on the opportunities for closing the gap between intelligent perception and resource-constrained deployment devices. More recently, \cite{xia2020survey} surveyed the use of semantics for visual SLAM, particularly reviewing the integration of ``semantic extractors'' (object recognition and semantic segmentation) within modern visual SLAM pipelines.

Much of the growth in semantics-based approaches has coincided with the increase in capabilities brought about by the modern deep learning revolution. Schmidhuber~\cite{schmidhuber2015deep} presented a detailed review of deep learning in neural networks as applied through supervised, unsupervised and reinforcement learning regimes. They introduced the concept of ``Credit Assignment Paths'' (CAPs) representing chains of causal links between events, which help understand the level of depth required for a given Neural Network (NN) application. With a vision of creating general purpose learning algorithms, they highlighted the need for a brain-like learning system following the rules of fractional neural activation and sparse neural connectivity. \cite{liu2020deep} presented a more focused review of deep learning, surveying generic object detection. They highlighted the key elements involved in the task such as the accuracy-efficiency trade-off of detection frameworks, the choice and evolution of backbone networks, the robustness of object representation and reasoning based on additionally available context. 

A significant body of work has focused on extracting more meaningful abstractions of the raw data typically obtained in robotics such as 3D point clouds. Towards this end, a number of surveys have been conducted in recent years for point cloud filtering~\cite{han2017review} and description~\cite{hana2018comprehensive}, 3D shape/object classification~\cite{liu2019deep, guo2020deep}, 3D object detection~\cite{ioannidou2017deep, arnold2019survey, rahman2019recent, liu2019deep, guo2020deep}, 3D object tracking~\cite{guo2020deep} and 3D semantic segmentation~\cite{grilli2017review, ioannidou2017deep, zhang2019review, lateef2019survey, xie2020review, malinverni2019deep, liu2019deep, guo2020deep}. With only a couple of exceptions, all of these surveys have particularly reviewed the use of \textit{deep learning} on 3D point clouds for respective tasks. Segmentation has also long been a fundamental component of many robotic and autonomous vehicle systems, with \textit{semantic} segmentation focusing on labeling areas or pixels in an image by class type. In particular the overall goal is to label by class, not by instance. For example, in an autonomous vehicle context this goal constitutes labeling pixels as belonging to \textit{a} vehicle, rather than as a specific instance of a vehicle (although that is also an important capability). The topic has been the focus of a large quantity of research with resulting survey papers that focus primarily on semantic segmentation, such as \cite{lateef2019survey, garcia2018survey, guo2018review, zhao2017survey, siam2017deep, thoma2016survey, zhu2016beyond}.

Beyond these flagship domains, semantics have also been investigated in a range of other subdomains. \cite{Ramirez-Amaro2019} reviewed the use of semantics in the context of understanding human actions and activities, to enable a robot to execute a task. They classified semantics-based methods for \textit{recognition} into four categories: syntactic methods based on symbols and rules, affordance-based understanding of objects in the environment, graph-based encoding of complex variable relations, and knowledge-based methods. In conjunction with recognition, different methods to \textit{learn and execute} various tasks were also reviewed including learning by demonstration, learning by observation and execution based on structured plans. Likewise for the service robotics field, \cite{Liu2016} presented a survey of vision-based semantic mapping, particularly focusing on its need for an effective human-robot interface for service robots, beyond pure navigation capabilities. \cite{paulius2019survey} also surveyed knowledge representations in service robotics.

Many robots are likely to require image retrieval capabilities where semantics may play a key role, including in scenarios where humans are interacting with the robotic systems. \cite{enser2003towards} and \cite{Liu2007} surveyed the ``semantic gap'' in current content-based image retrieval systems, highlighting the discrepancy between the limited descriptive power of low-level image features and the typical richness of (human) user semantics. Bridging this gap is likely to be important for both improved robot capabilities \textit{and} better interfaces with humans. Some of the reviewed approaches to reducing the semantic gap, as discussed in~\cite{Liu2007}, include the use of object ontology, learning meaningful associations between image features and query concepts and learning the user's intention by relevance feedback. This semantic gap concept has gained significant attention and is reviewed in a range of other papers including~\cite{li2016socializing, sudha2015reducing, ramkumar2014ontology, zhang2012review, ehrig2006ontology, hare2006mind, wen2005review}. Acting upon the enriched understanding of the scene, robots are also likely to require sophisticated grasping capabilities, as reviewed in \cite{du2020vision}, covering vision-based robotic grasping in the context of object localization, pose estimation, grasp detection and motion planning. Enriched interaction with the environment based on an understanding of what can be done with an object - its ``affordances'' - is also important, as reviewed in \cite{ardon2020affordances}. Enriched interaction with humans is also likely to require an understanding of language, as reviewed recently by \cite{tellex2020}. This review covers some of the key elements of language usage by robots: collaboration via dialogue with a person, language as a means to drive learning and understanding natural language requests, and deployment, as shown in application examples.

\subsection{Summary and Rationale for this Survey}

The majority of semantics coverage in the literature to date has occurred with respect to a specific research topic, such as SLAM or segmentation, or targeted to specific application areas, such as autonomous vehicles. As can be seen in the previous section, there has been both extensive research across these fields as well as a number of key survey and review papers summarizing progress to date. These deep dives into specific sub-areas in robotics can provide readers with a deep understanding of technical considerations regarding semantics in that context. As the field continues to grow however there is increasing need for an overview that more broadly covers semantics across all of robotics, whilst still providing sufficient technical coverage to be of use to practitioners working in these fields. For example, while~\cite{cadena2016past} extensively considers the use of semantics primarily within SLAM research, there is a need to more broadly cover the role of semantics in various robotics tasks and competencies which are closely related to each other. The task, ``bring a cup of coffee'', likely requires semantic understanding borne out of both the underlying SLAM system \textit{and} the affordance-grasping pipeline. This survey therefore goes beyond specific application domains or methodologies to provide an overarching survey of semantics across all of robotics, as well as the semantics-enabling research that occurs in related fields like computer vision and machine learning. To encompass such a broad range of topics in this survey, we have divided our coverage of research relating to semantics into a) the fundamentals underlying the current and potential use of semantics in robotics, b) the widespread use of semantics in robotic mapping and navigation systems, and c) the use of semantics to enhance the range of interactions robots have with the world, with humans, and with other robots. 

This survey is also motivated by timeliness: the use of semantics is a rapidly evolving area, due to both significant current interest in this field, as well as technological advances in local and cloud compute, and the increasing availability of data that is critical to developing or training these semantic systems. Consequently, with many of the key papers now half a decade old or more, it is useful to capture a snapshot of the field as it stands now, and to update the treatment of various topic areas based on recently proposed paradigms. For example, this paper discusses recent semantic mapping paradigms that mostly post-date key papers by~\cite{cadena2016past, kostavelis2015semantic}, such as combining single- and multi-view point clouds with semantic segmentation to directly obtain a local semantic map~\cite{Qi2017a, lawin2017deep, huang2016point, Boulch2018, Dai2018a}. Whilst contributing a new overview of the use of semantics across robotics in general, we are also careful to adhere where possible to recent proposed taxonomies in specific research areas. For example, in the area of 3D point clouds and their usage for semantics, within Section~\ref{sec:semRep+SU+SLAM}, with the help of key representative papers, we \textit{briefly} describe the recent research evolution of using 3D point cloud representations for learning object- or pixel-level semantic labeling, in line with the taxonomy proposed by existing comprehensive surveys~\cite{guo2020deep, liu2019deep, lateef2019survey, xie2020review, ioannidou2017deep}. Finally, beyond covering new high level conceptual developments, there is also the need to simply update the paper-level coverage of what has been an incredibly large volume of research in these fields even over the past five years. The survey refers to well over a hundred research works from the past year alone, representative of a much larger total number of research works. This breadth of coverage would normally come at the cost of some depth of coverage: here we have attempted to cover the individual topics in as much detail as possible, with over 900 referenced works covered in total. Where appropriate we also make reference to prior survey and review papers where further detailed coverage may be of interest, such as for the topic of 3D point clouds and their usage for semantics.

Moving beyond single application domains, we also provide an overview of how the use of semantics is becoming an increasingly integral part of many trial (and in some cases full scale commercial) deployments including in autonomous vehicles, service robotics and drones. A richer understanding of the world will open up opportunities for robotic deployments in contexts traditionally \textit{too difficult} for safe robotic deployment: nowhere is this more apparent perhaps than for on-road autonomous vehicles, where a subtle, nuanced understanding of all aspects of the driving task is likely required before robot cars become comparable to, or ideally superior to, human drivers. Compute and data availability has also enabled many of the advancements in semantics-based robotics research; likewise these technological advances have also facilitated investigation of their deployment in robotic applications that previously would have been unthinkable - such as enabling sufficient on-board computation for deploying semantic techniques on power- and weight-limited drones. We cover the current and likely future advancements in computational technology relevant to semantics, both local and online versions, as well as the burgeoning availability of rich, informative datasets that can be used for training semantically-informed systems. 

In summary, this survey paper aims to provide a unifying overview of the development and use of semantics across the entire robotics field, covering as much detailed work as feasible whilst referencing the reader to further details where appropriate. Beyond its breadth, the paper represents a substantial update to the semantics topics covered in survey and review papers published even only a few years ago. By surveying the technical research, the application domains and the technology enablers in a single treatment of the field, we can provide a unified snapshot of what is possible now and what is likely to be possible in the near future.

\subsection{Taxonomy and Survey Structure}

\begin{figure}
    \centering
    \includegraphics[width=\textwidth]{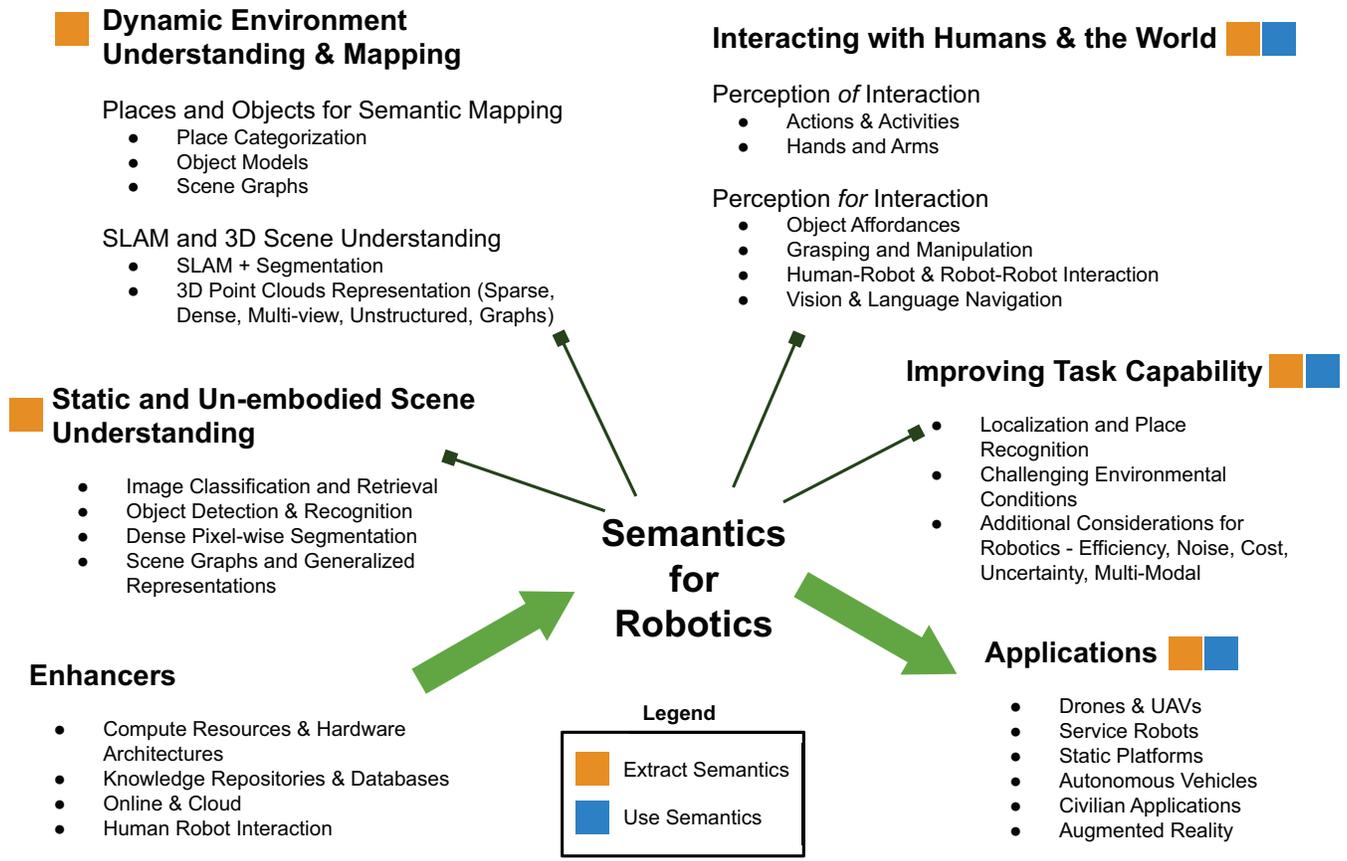}
    \caption{A taxonomy for semantics in robotics. Four broad categories of semantics research are complemented by technological, knowledge and training-related enhancers and lead to a range of robotic applications. Areas can be primarily involved in extracting semantics, using semantics or a combination of both.}
    \label{fig:taxonomy}
\end{figure}

Existing literature covering the role of \textit{semantics} in robotics is fragmented and is usually discussed in a variety of task- and application-specific contexts. In this survey, we consolidate the disconnected semantics research in robotics; draw links with the fundamental computer vision capabilities of extracting semantic information; cover a range of potential applications that typically require high-level decision making; and discuss critical upcoming enhancers for improving the scope and use of semantics. To aid in navigating this rapidly growing and already  sizable field, here we propose a taxonomy of semantics as it pertains to robotics (see Figure~\ref{fig:taxonomy}). We find the relevant literature can be divided into four broad categories: 

\begin{enumerate}
    \item \textit{Static and Un-embodied Scene Understanding}, where the focus of research is typically on developing intrinsic capability to extract semantic information from images, for example, object recognition and image classification. The majority of research in this direction uses single image-based 2D input to infer the underlying semantic or 3D content of that image. However, image acquisition and processing in this case is primarily static in nature (including videos shot by a static camera), separating it conceptually from a mobile embodied agent's dynamic perception of the environment due to motion of the agent. Because RGB cameras are widely used in robotics, and the tasks being performed, such as object recognition, are also performed by robots, advances in this area are relevant to robotics research. In Section \ref{secn:semanticsfundamental}, we introduce the fundamental components of semantics that relate to or enable robotics, focusing on topics that have been primarily or initially investigated in non-robotics but related research fields, such as computer vision. We cover the key components of semantics as regards object detection, segmentation, scene representations and image retrieval, all highly relevant capabilities for robotics, even if not all the work has yet been demonstrated on robotic platforms.
    
    \item \textit{Dynamic Environment Understanding and Mapping}, where the research is typically motivated by the mobile or dynamic nature of robots and their surroundings. The research literature in this category includes the task of semantic mapping, which could be topological, or a dense and precise 3D reconstruction. These mapping tasks can often leverage advances in static scene understanding research, for example, place categorization (image classification) forming the basis of semantic topological mapping, or pixel-wise semantic segmentation being used as part of a semantic 3D reconstruction pipeline. Semantic maps provide a representation of information and understanding at an environment or spatial level. With the increasing use of 3D sensing devices, along with the maturity of visual SLAM, research on semantic understanding of 3D point clouds is also growing, aimed at enabling a richer semantic representation of the 3D world. In Section \ref{secn:mappingSLAM}, we cover the use of semantics for developing representations and understanding at an environment level. This includes the use of places, objects and scene graphs for semantic mapping, and 3D scene understanding through Simultaneous Localization And Mapping (SLAM) and point clouds processing.
    
    \item \textit{Interacting with Humans and the World}, where the existing research ``connects the dots" between the ability to \textit{perceive} and the ability to \textit{act}. The literature in this space can be further divided into the ``perception \textit{of} interaction'' and ``perception \textit{for} interaction''. The former includes the basic abilities of understanding actions and activities of humans and other dynamic agents, and enabling robots to learn from demonstration. The latter encompasses research related to the use of the perceived information to act or perform a task, for example, developing a manipulation strategy for a detected object. In the context of robotics, detecting an object's affordances can be as important as recognizing that object, enabling semantic reasoning relevant to the task and affordances (e.g. `cut' and `contain') rather than to the specific object category (e.g. `knife' and `jar'). While object grasping and manipulation relate to a robot's interaction with the environment, research on interaction with other \textit{humans} and \textit{robots} includes the use of natural language to generate inverse semantics, or to follow navigation instructions. Section \ref{secn:interaction} addresses the use of semantics to facilitate robot interaction with the world, as well as with the humans and robots that inhabit that world. It looks at key issues around affordances, grasping, manipulation, higher-level goals and decision making, human-robot interaction and vision-and-language navigation.

    \item \textit{Improving Task Capability}, where researchers have focused on utilizing semantic representations to improve the capability of other tasks. This includes for example the use of semantics for high-level reasoning to improve localization and visual place recognition techniques. Furthermore, semantic information can be used to solve more challenging problems such as dealing with challenging environmental conditions. Robotics researchers have also focused on techniques that unlock the full potential of semantics in robotics, since existing research has not always been motivated by or had to deal with the challenges of real world robotic applications, by addressing challenges like noise, clutter, cost, uncertainty and efficiency. In Section~\ref{secn:improvingTask}, we discuss various ways in which researchers extract or employ semantic representations for localization and visual place recognition, dealing with challenging environmental conditions, and generally enabling semantics in a robotics context through addressing additional challenges.
\end{enumerate}

The four broad categories presented above encompass the relevant literature on how semantics are defined or used in various contexts in robotics and related fields. This is also reflected in Figure~\ref{fig:taxonomy} through `extract semantics' and `use semantics' labels associated with different sections of the taxonomy. Extracting semantics from images, videos, 3D point clouds, or by actively traversing an environment are all methods of  creating semantic representations. Such semantic representations can be input into high-level reasoning and decision-making processes, enabling execution of complex tasks such as path planning in a crowded environment, pedestrian intention prediction, and vehicle trajectory prediction. Moreover, the use of semantics is often fine-tuned to particular applications like agricultural robotics, autonomous driving, augmented reality and UAVs. Rather than simply being exploited, the semantic representations themselves can be jointly developed and defined in consideration of how they are then used. Hence, in Figure~\ref{fig:taxonomy}, the sections associated with `use semantics' are also associated with `extract semantics'. These high-level tasks can benefit from advances in fundamental and applied research related to semantics. But this research alone is not enough: advances in other areas are critical, such as better cloud infrastructure, advanced hardware architectures and compute capability, and the availability of large datasets and knowledge repositories. Section \ref{secn:discussion} reviews the influx of semantics-based approaches for robotic deployments across a wide range of domains, as well as the critical technology enablers underpinning much of this current and future progress. Finally, Section \ref{secn:conclusion} discusses some of the key remaining challenges in the field and opportunities for addressing them through future research, concluding coverage of what is likely to remain an exciting and highly active research area into the future.


\section{STATIC AND UN-EMBODIED SCENE UNDERSTANDING}
\label{secn:semanticsfundamental}
Many of the advances in the development and use of semantics have occurred in the computer vision domain, with a heavy focus on identifying and utilizing semantics in terms of images or scenes - for example in \textit{2D semantic segmentation}, \textit{object recognition}, \textit{scene representations}, \textit{depth estimation}, \textit{image classification} and \textit{image retrieval}. For the majority of these research tasks, image processing is primarily static in nature, conceptually distinct from a mobile embodied agent's dynamic perception of the environment. Nevertheless, these processes can all play a role in robotics, whether it be for a robot perceiving and understanding the scene in front of it, or searching through its map or database to recognize a place it has visited before. In this section we focus on the fundamental computer vision research involving semantics for static scene understanding which is directly relevant to robotics. The following subsections discuss semantic representations explored in the literature, firstly in the context of image classification and retrieval, followed by extracting and using semantic information in the form of objects and dense segmentation, and finally,  scene representation including scene graphs and semantic generalized representations.

\subsection{Image Classification and Retrieval}

Image classification is the process of determining the content shown within an image, typically at a \textit{whole} image level, although the process may involve detecting and recognizing specific objects or regions within the overall image. Complementing classification, image retrieval refers to the process of parsing a database of images and picking out images that meet a certain content criteria, which can be specified in a number of ways. For example, images can be retrieved based on a ``query'' image (find images with similar content), or based on some description of the desired content within the image (find images of trees). Image retrieval in particular has a strong overlap with visual mapping and place recognition techniques, and hence much of the mapping or spatially-specific retrieval research is covered in that section. Here we touch on classification and retrieval techniques with a focus on developments in the semantic space.

\subsubsection{Image-based Classification}
A typical goal of this task is to semantically tag images based on their visual content, for example, classifying an image as belonging to a category like ``peacock'' or ``train station''. The former represents the presence of an ``object'' within an image, whereas the latter represents a ``place''. In a broader sense however, image classification or recognition can be based on a hierarchy of concepts that might exist in the visual observation. \cite{Li-jiaLi2009} presented a hierarchical generative model that classifies the overall scene, recognizes and segments each object component, and annotates the image with a list of tags. It is claimed to be the first model that performs all three tasks in one coherent framework. The framework was able to learn robust scene models from noisy web data such as images and user tags from Flickr.com. \cite{Li2010} proposed a high-level image representation, \textit{Object Bank}, based on a scale-invariant response map of a large number of pre-trained generic object detectors, for the task of object recognition and scene classification. \cite{Yao2012} designed an approach to jointly reason about regions, location, class and spatial extent of objects, the presence of a class in the image, and the scene type. The novel reasoning scheme at the segment level enabled efficient learning and inference. \cite{Sanchez2013} proposed the \textit{Fisher Vector} - a state-of-the-art patch encoding technique - as an alternative to Bag of Visual Words (BoVW), where patches are described by their derivation from a universal generative Gaussian Mixture Model (GMM). \cite{Qi2019} developed a Bag of Semantic Words (BoSW) model based on automatic dense image segmentation and semantic annotation, achieved by graph cut and (Support Vector Machine) SVM respectively. While the SVM learns one-versus-all classification for different semantic categories, the reduced vocabulary size due to BoSW decreases computation time and increases accuracy. 
While effective, the methods above rely on hand-designed image features that are not optimally designed for image classification.  Consequently alternative approaches that can  automatically learn not only the classifier, but also the optimal features, have been investigated, with methods based on deep learning techniques generally being superior, as detailed below.

Like many other research areas, deep learning has played an increasing role in image classification over the past decade. With the availability of large-scale image dataset like ImageNet~\cite{deng2009imagenet} and the availability of Graphics Processing Units (GPUs), learning via \textit{deep} Convolutional Neural Network (CNN) such as AlexNet~\cite{krizhevsky2012imagenet} opened up enormous opportunities for enhanced scene understanding. \cite{zhou2014learning} introduced the \textit{Places} dataset, consisting of 7 million labeled scenes, and demonstrated state-of-the-art performance on scene-centric datasets. Through this dataset, complementing ImageNet, \cite{zhou2014learning} highlighted the differences in internal representations of object-centric and scene-centric networks. \cite{Zhou2018} extended this line of work, discussing the released Places database of 10 million images and its corresponding Places-CNNs baselines for scene recognition problems. \cite{zhou2015object} showed that object detectors naturally emerge from training CNNs for scene classification. They argue that since scenes are typically composed of objects, CNNs trained on them automatically discover meaningful object detectors, representative of the learned scene category. This work then demonstrated that the same network can perform both scene recognition and object localization in a single forward-pass, without an explicitly taught notion of objects.
The methods above are generally referred to as weakly supervised detectors, because they are trained with scene-level annotations but are able to localise objects within a scene. One of their main limitations was that their object detection accuracy was significantly inferior to fully supervised detection methods that relied on object localisation labels -- further research in the area has focused on reducing this performance gap. A CNN visualisation technique was presented in \cite{Zhou2016} where the Class Activation Map (CAM) was based on Global Average Pooling (GAP) in CNNs.
This work demonstrated that GAP enables accurate localization ability (i.e. activation of scene category-specific local image regions), despite being trained on image-level labels for scene classification. 

Meng \textit{et al}.~\cite{Meng2017} designed a first-of-its-kind semantic place categorization system based on text-image pairs extracted from social media. They concatenate and learn features from two CNNs -- one each for the two modalities -- to predict the place category. The proposed system uses a newly curated dataset with 8 semantic categories: home, school, work, restaurant, shopping, cinema, sports, and travel. \cite{Wei2019} explored aggregation of CNN features through two different modules: a ``Content Estimator'' and a ``Context Estimator'', where the latter is comprised of three sub-modules based on text, object and scene context. The authors claim it is the first work that leverages text information from a scene text detector for extracting context information.
The papers above were successful at introducing approaches that combined different modalities, but did not include geometrical information, limiting their ability to ``understand'' the scene.

When depth information is available, including when explicitly available from range-based sensors, RGB-D-based scene recognition and classification techniques classify scene images using aligned color and depth information. Wang \textit{et al.} \cite{wang2016modality} extracted and combined deep discriminative features from different modalities in a component-aware fusion manner. Gupta \textit{et al.} \cite{gupta2016cross} transferred the RGB model to depth net using unlabeled paired data according to their mid-level representations. More recently, Du \textit{et al.} \cite{du2019translate} presented a unified framework to integrate the tasks of cross-modal translation and modality specific recognition. While additional cues in the form of depth information or text descriptions improve image classification, the essence of the task is to enable semantic scene understanding by providing a ``compressed'' but meaningful representation of the visual observation, which a robot can then use to intelligently reason about the environment.

\subsubsection{Semantics for Image Retrieval and Classification}

Except in specific circumstances or in the case of instance-level retrieval (finding a specific image), broader content-based image retrieval generally requires some semantics-based knowledge of the content of images in order to be feasible. Much of the research in this area has focused on what form that knowledge should take, and how similarity should be calculated. \cite{Hong2016} explored concept-relationship based feature extraction and learning using textual and visual data for content based image retrieval. This was achieved by using five pre-defined specific concept relationships: \textit{complete similarity} (Beijing, Peking), \textit{type similarity} (Husky, Bulldog), \textit{Hypernym Hyponym} (Husky, Domestic Dog), \textit{Parallel relationship} (cat, dog), and \textit{unknown relationship}; these conceptual semantic relationships were shown to improve retrieval performance. \cite{huang2018sketch} proposed fusing low-level and high-level features from sketches and images, along with clustering-based re-ranking optimization for image retrieval. \cite{lu2017latent} developed a novel joint binary codes learning method that combines image features with latent semantic features (labels/tags of images); the intent was to encode samples sharing the same semantic concepts into similar codes, rather than only preserving geometric structure via hashing. \cite{Gordo2017} explored a method to retrieve images based on semantic information like ``a boy jumping'' or ``throw'', as opposed to traditional instance-level retrieval. They proposed a similarity function based on annotators' captions as a computable surrogate of true semantic similarity for learning a semantic visual representation. Learning a joint embedding for visual and textual representations improved accuracy, and enabled combined text and image based retrieval.

To reduce the semantic gap in existing visual BoW approaches, \cite{Bai2018} extracted SIFT~\cite{lowe2004distinctive}, Local Binary Patterns (LBP)~\cite{ojala2002multiresolution}, and color histogram features \textit{separately} for the foreground and background in an image, segmented using the SaliencyCut method~\cite{Cheng2015}. \cite{Quinn2018} presented \textit{Situate} - an architecture to learn models that capture visual features of objects and their spatial configuration to retrieve instances of visual situations like ``walking a dog'' or ``a game of ping-pong''. The method actively searches for each expected component of the situation in the image to calculate a matching score. \cite{Barz2019} explored hierarchy-based semantic embeddings for image retrieval by incorporating prior knowledge about semantic relationships between classes obtained from the class hierarchy of \textit{WordNet}~\cite{miller1998wordnet}. \cite{lee2019sfnet} used an end-to-end trainable CNN to learn ``semantic correspondences'' as dense flow between images depicting different instances of the same semantic object category, such as a bicycle. To train the CNN, they used binary segmented images with synthetic geometric transformations and a new differentiable \textit{argmax} function. 

\cite{li2019visual} proposed the Visual Semantic Reasoning Network (VSRN), based on connections \textit{between} image regions (ROIs from F-RCNN~\cite{ren2015faster}). This approach learns features with semantic relationships from the pairwise affinity between regions using a Graph Convolutional Network. The research included both image-to-text (caption) and text-to-image (image) retrieval capabilities; the learnt representation (also visualized in 2D) captures key semantic objects (bounding boxes) and semantic concepts (within a caption) of a scene as in the corresponding text caption. \cite{Engilberge2018} designed a new two-path neural network architecture that maps images and text while enabling spatial localization in the proposed semantic-visual embedding. Using this system, text queries like ``a burger'', ``some beans'' and ``a tomato'', can be localized in an image comprising all of these. 
\cite{Chang2017} presented a framework for event detection that uses ``semantic interaction'' based on pairwise affinity between semantic representations of multiple source videos. This approach enables the detection of events such as ``birthday party'', ``grooming an animal'', ``town hall meeting'' or ``marriage proposal''.

Semantic image retrieval and semantic image classification are closely related to each other, as improved methods for the latter inherently enable higher accuracy for the former. This is in part due to the high level of abstraction and reasoning capability that is attained through semantically representing images. Such an approach also bridges the gap between natural language semantics of humans and the user's intended retrieval outcome, thus reducing the \textit{semantic gap}. In a world where robots and humans co-exist, reducing the semantic gap will likely lead to more seamless human-robot interaction. 

\subsection{Object Detection and Recognition}
While semantic reasoning at the \textit{whole} image level can enable high-level decision making for a robot such as that required for path planning, semantically understanding the content \textit{within} an observed image is necessary for a robot to perform a task such as manipulating an object. A key task for many robots operating in real world environments therefore is object detection and recognition. While there is no absolute definition of what constitutes an object, general definitions revolve around the concept of objects being distinct things or entities in the environment, with the ability to be seen and touched. Before moving into more sophisticated uses of objects and object understanding in the context of robotic tasks like mapping and environmental interaction, we first cover some key background. 

\textit{Detection} is exactly that: detecting what objects are present in a particular environment, observed through the robot's sensors. \textit{Recognition} involves calculating what types of objects are present, and can be performed at both a class level - all the mugs in the room - and at instance level - a specific, single mug. Classical approaches to object recognition include \textit{template-based} approaches, which match potential objects in the scene to some form of 2D or 3D \textit{exemplar} of the object class; \textit{hand-crafted feature-based} methods, which build object structures on top of simple edge, SIFT~\cite{lowe1999object} or SURF~\cite{bay2006surf} features, and \textit{direct} approaches that perform matching based on pixel intensities or intensity gradients in an image. Because of the inherent difficulty of the problem in all but the simplest of circumstances, decades of research has developed enhancements to these processes including efficient tree-based search methods, various geometric matching tests and a range of other mechanisms. 

With the advent of modern deep learning techniques, a number of key advances were made in CNN-based object detection approaches, including Fast R-CNN~\cite{Girshick2015}, SPPNet~\cite{He2015} and Faster R-CNN~\cite{ren2015faster}. In particular, approaches have started to investigate the use of semantics as part of the detection pipeline. \cite{cadena2015icra} presented a modular scene understanding system that used semantic segmentation to improve object detection performance for out-of-the-box specific object detectors. In the 3D object detection area, \cite{simon2019complexer} explored a detection and tracking pipeline for autonomous vehicles using RGB, LiDAR, and dense semantic segmentation to form semantic point clouds (voxel grids), which are then fed into Complex-YOLO to obtain 3D detections. They use Labeled Multi-Bernoulli-Filter for multi-target tracking and claim to be the first to fuse visual semantics with 3D object detection. \cite{Yao2019} presented a generic object tracking framework that uses a semantic segmentation network for object proposals by ranking and selecting relevant feature maps from the final layer based on activations. Temporal information and template models were used to improve prediction. \cite{Zhang2018} designed a Detection with Enriched Semantics (DES) network for object detection that uses semantic segmentation as an input to the network (in a weakly supervised manner) to enrich the learnt feature maps. They also use a global activation module for extracting high-level information from higher-order layers of the network. \cite{Li2019a} proposed an end-to-end learning framework for pedestrian attribute recognition using a Graph Convolution Network comprising sub-networks: one to capture spatial relations between image regions and another to learn semantic relationships between attributes. 

The state-of-the-art 2D object detectors have also been shown to be useful in facilitating automated annotation and recovery of 3D objects when used in conjunction with sparse LiDAR data, as demonstrated recently in ~\cite{zakharov2020autolabeling}. Pursuing this capability is partly motivated by the current requirement for enhanced 3D scene understanding to enable practical applications like autonomous driving. Given this requirement, there has been significant activity in 3D object detection research, which is covered in detail in Section~\ref{sec:semRep+SU+SLAM}, especially in the context of using 3D point clouds. In robotics, object detection and recognition is fundamental to the core competencies of a robot, that is, perception and interaction. Spatially-intelligent robots would require understanding both the object of interest and everything else that surrounds it in order to successfully perform many tasks, which can only be achieved by accurate 2D and 3D object recognition and dense semantic segmentation.

\subsection{Dense Pixel-wise Segmentation}
Object-only semantic understanding is typically motivated by applications that define an object or region of interest; it is detected, recognized, often tracked, and in the case of robotics, grasped and manipulated. While dealing with objects, the ``background'' information is typically discarded. In an indoor environment where the task might be counting coffee mugs of different colors, this background information exists in the form of walls, floors, and ceilings. Similarly, for an outdoor environment, the task of pedestrian detection and tracking might consider roads and buildings as background. Beyond the notion of background, the lack of a single unique way of representing object shapes paves the way for dense image segmentation. 

Like object detection, segmentation is a research topic with great relevance to robotics (especially autonomous vehicles in recent years) where much of the pioneering work was performed in the computer vision domain. The core segmentation task is straightforward to describe: partitioning an image up into segments, but the ways in which this can be achieved, and its subsequent use cases for robotics vary hugely. Segmentation is tightly coupled to semantics because the most common segmentation approaches revolve around dividing up the image into semantically meaningful areas, with objects being a common example - in an autonomous vehicle context this could be pedestrians, vehicles, and traffic signs. Dense semantic segmentation is not only often more informative than \textit{specific} object detection, it also discloses various ways of leveraging additional context for performing a task, for example, utilizing spatial co-existence or physics based priors. Semantic segmentation is increasingly becoming a core part of systems in the robotics field and its development and use in mapping or interaction contexts is covered in the appropriate sections later in this survey. Here we overview the key foundational work which many of these robotic systems then build upon.

\subsubsection{2D Semantic Segmentation}

The world is a three-dimensional environment but much of the key work has occurred in the 2D image co-ordinate space. Already a significant research topic before modern deep learning techniques became dominant, earlier works involved elevating the capability of coherent region-based segmentation~\cite{lucchese2001colour}, including watershed~\cite{vincent1991watersheds} and mean-shift~\cite{comaniciu2002mean} like techniques, to a more meaningful parts-based semantic segmentation~\cite{schnitman2006inducing, athanasiadis2007semantic, lempitsky2011pylon}. \cite{schnitman2006inducing} induced semantic labeling using an example image from a common image domain based on a non-parametric model. \cite{athanasiadis2007semantic} presented semantic segmentation and object labeling based on a novel region-growing algorithm, combined with context obtained from ontological knowledge representation. \cite{lempitsky2011pylon} explored a hierarchical tree based, rather than flat, partitioning of an image~\cite{fulkerson2009class, shotton2006textonboost} within the scope of using graph cut~\cite{boykov2001fast} for image segmentation. Given that these approaches were mostly based on prior information about particular segmentation tasks, it was challenging for them to produce highly accurate segmentation results.

With the advent of modern deep learning came a proliferation of new semantic segmentation methods. Long \textit{et al}.~\cite{Long2015} presented the Fully Convolutional Network (FCN) that processes arbitrary size input images and produces correspondingly-sized output with efficient inference and learning. They also defined a skip architecture that combines semantic information from a deep, coarse layer with appearance information from a shallow fine layer to produce accurate and detailed segmentations. Yu \textit{et al}.~\cite{yu2015multi} designed a CNN specifically for dense prediction, with their proposed model based on ``dilated convolutions''  systematically aggregating multi-scale contextual information without losing resolution. Badrinarayanan \textit{et al}.~\cite{Badrinarayanan2017} presented SegNet, an ``encoder-decoder'' architecture for semantic segmentation. The main novelty of SegNet was in the up-sampling strategy, where the decoder used the pooling indices from the encoder's max pooling step to perform non-linear up-sampling, thereby eliminating the need for learning to up-sample. The system was relatively efficient in terms of memory and computation time. Another relevant model was the Unet~\cite{ronneberger2015u}, that consisted of an encoder-decoder architecture with skip connections from the contracting path (i.e., encoder) to the expanding path (i.e., decoder), with the goal of addressing the issue of vanishing gradients. These approaches represent some of the first attempts to use deep learning for semantic segmentation, and the impressive results at that time significantly motivated the development of new deep learning approaches.

The Efficient neural Network (ENet) was proposed by Paszke \textit{et al}.~\cite{Paszke2016} for tasks requiring low latency operations; their network was up to 18x faster, required 75x less FLOPs, and had 79x less parameters than comparable methods at the time, while achieving similar or better accuracy. Chen \textit{et al}.~\cite{Chen2018a} designed a semantic segmentation network based on: 1) atrous (dilated) convolutions that enlarged the field of view of filters using the same number of parameters, 2) Atrous Spatial Pyramid Pooling (ASPP) that aided in segmenting objects at multiple scales, and 3) fully connected CRFs that improved the localization boundaries. More recently, Takikawa \textit{et al}. presented the Gated-SCNN~\cite{takikawa2019gated}: a two-branch CNN model that analyses the RGB image and an image gradient (representing the shape information of the image). These branches are connected by a new type of gating mechanism that joins the higher-level activations of the RGB image to the lower-level activations of the gradient image. Gated-SCNN~\cite{takikawa2019gated} held the state-of-the-art segmentation results in several public computer vision datasets.

As the application of segmentation methods continued to expand into challenging scenarios such as autonomous vehicles, it became clear that improvements were needed to make them perform well in challenging or adverse environmental conditions. Sakaridis \textit{et al}.~\cite{Sakaridis2018} developed a pipeline to add synthetic fog to real clear-weather images using incomplete depth information, to address the problem of semantic scene understanding in foggy conditions. They used supervised and semi-supervised learning techniques to improve performance, introducing the Foggy Driving dataset and Foggy Cityscapes dataset. In order to deal with variations in weather conditions, \cite{erkent2020semantic} explored unsupervised domain adaptation to learn accurate semantic segmentation in a \textit{challenging target} weather condition using an \textit{ideal source} weather condition. 

Integrating information over time - in an autonomous vehicle context for example over multiple camera frames - can also improve performance in challenging circumstances. Kundu \textit{et al}.~\cite{Kundu2016} proposed a method for long-range spatio-temporal regularization in semantic video segmentation, in contrast to a naive regularization over the video volume, which does not take into account camera and object motion. They used optical flow and dense CRF over points optimized in Euclidean feature space to improve accuracy, and demonstrated the effectiveness of the method on outdoor Cityscapes data~\cite{Cordts2016}. \cite{guarino20temporal} presented a Bayesian filtering approach to semantic segmentation of a video sequence, where each pixel is considered a random variable with a discrete probability distribution function. He \textit{et al}.~\cite{He2017a} presented a superpixel-based multi-view CNN for semantic segmentation that leveraged information from additional views of the scene, by first computing region correspondences through optical flow and superpixels, and then using a novel spatio-temporal pooling layer to aggregate information over space and time. The training process in this case benefited from unlabeled frames that led to improved prediction performance. More recently, in order to address the problem of data labeling for supervised semantic segmentation, \cite{lottes2020gradient} employed gradient and log-based active learning for semantic segmentation to distinguish between crop and weed plants in an agricultural field.

To foster research in the context of temporally-coherent semantic segmentation, \cite{perazzi2016benchmark} presented a video segmentation benchmark using the proposed DAVIS (Densely Annotated VIdeo Segmentation) dataset. The benchmark helped increase the community's interest in developing novel techniques to address this task. \cite{caelles2017one} proposed a one-shot video object segmentation method that used only a single labeled training example, that is, the first frame. In order to reduce the reliance on large-scale data, \cite{wang2017video} leveraged existing annotated image datasets to augment video training data, enabling them to learn diverse saliency information while also preventing overfitting due to limited original data. In an extended work~\cite{wang2017saliency}, the authors considered the task of unsupervised video segmentation, leveraging object saliency cues to achieve temporally-consistent pixel labeling. \cite{perazzi2017learning} presented a learning strategy based on single image segmentation and incremental frame-by-frame processing to refine video object segmentation. While spatio-temporal cues have been demonstrated to be useful for segmentation in \textit{videos}, \cite{pathak2017learning} presented a learning framework to correctly group pixels based on motion, to improve object segmentation in \textit{static images}.

Beyond semantic segmentation based on classes, instance-level segmentation - recognizing specific singular areas of interest - is also highly relevant for many robotic applications. Held \textit{et al}.~\cite{Held2016} developed a real-time probabilistic 3D segmentation method that combined spatial, temporal, and semantic information to help in the decision between splitting and merging of the initial coarse segmentation, significantly reducing under- and over-segmentation results on the KITTI dataset. The Faster R-CNN~\cite{ren2015faster} approach was also extended by He \textit{et al}.~\cite{He2017} for instance segmentation, by adding a branch for predicting object masks in parallel with existing bounding box recognition. The proposed system demonstrated state-of-the-art results on the following COCO challenges: instance segmentation, bounding box object detection, and person keypoint detection. Bai \textit{et al}.~\cite{Bai2017} presented a CNN combining classical watershed transform and modern deep learning to produce an energy map of an image, where object instances are unambiguously represented as energy basins, enabling direct extraction of high quality object instances.
Methods based on metric learning~\cite{harley2017segmentation,newell2017associative} have been explored for instance-level segmentation, with the aim of learning a pixel-based embedding where pixels of the same instance have a similar embedding. One of the current state-of-the-art methods is the masking score RCNN (MS-RCNN)~\cite{huang2019mask}, based on a network block which learns the quality of the predicted instance masks. Unifying the tasks of semantic segmentation and instance segmentation, \cite{kirillov2019panoptic} proposed a joint scene understanding task, termed ``panoptic segmentation'', while also defining a novel Panoptic Quality (PQ) metric to evaluate performance on this task. Panoptic segmentation has become the focus of research in many applications, in particular autonomous driving, with a number of novel techniques developed only recently~\cite{kirillov2019panopticFPN, xiong2019upsnet, li2020fusing, hou2020real, cheng2020panoptic, wang2020axial}.
It is likely that future 2D segmentation approaches will target several of the goals above in a joint manner.

\subsubsection{Semantic Segmentation using Depth and Other Sensing Modalities}

While visual segmentation using RGB imagery or video has been a significant focus of the research community, other sensing modalities can aid in segmentation, which in turn can then help the sensing pipeline. These other modalities are not just depth-related (in the case of commonly used LiDAR sensors on autonomous vehicles for example), but can include other sensors including infrared and multispectral cameras.

Silberman \textit{et al}.~\cite{Silberman2012} proposed a method to parse ``messy'' indoor scenes into floors, walls, supporting surfaces, and object regions, while also recovering support relationships. They classified objects into ``structural classes'' based on their role: a) ground, b) permanent structures like walls and ceilings, c) large furniture, and d) props that are movable. For reasoning about support, they used physical constraints and statistical priors. Gupta \textit{et al}.~\cite{Gupta2015} designed a set of algorithmic tools for perceptual organization and recognition in indoor scenes from RGB-D data. Their system produced contour detection, hierarchical segmentation, grouping by amodal completion, object detection and semantic labeling of objects and scene surfaces. \cite{cadena2016rss} explored the use of a Multi-modal stacked Auto-Encoder (MAE)~\cite{ngiam2011multimodal} to jointly estimate per-pixel depth and semantic labels. The authors demonstrated that learning a shared latent representation aids in semantic scene understanding even when input data has imperfect or missing information. \cite{Valada2017} presented UpNet, a fusion architecture that can learn from RGB, Near-InfraRed, and depth data to perform semantic segmentation. They also introduced a first-of-its-kind multispectral segmentation benchmark of unstructured forest environments. \cite{chen2019learning} explored the use of easy-to-obtain synthetic data for semantic segmentation, particularly making use of geometry in the form of synthetic depth maps. They used geometric information to first adapt the input to the segmentation network via image translation, and then adapt the output of the network via \textit{simultaneous} depth and semantic label prediction. More recently, \cite{wong2020identifying} addressed the problem of open-set instance segmentation by projecting point clouds into a category-agnostic embedding space, where clustering is used to perform 3D segmentation independent of semantics.

Semantic segmentation can also be used to improve the results of processing some of the sensing modalities. \cite{Bai2016} proposed semantically selecting dynamic object instances (and removing the background) from stereo pairs of images, in order to estimate optical flow for each instance individually. The estimated flow for the background and foreground could then be merged to obtain the final, improved result. \cite{Yang2018a} explored the use of semantics to improve per-pixel disparity estimation by simultaneously predicting per-pixel semantic labels and using intermediate CNN feature embeddings from the segmentation network. \cite{Dovesi2019} presented RTS2Net - a single, compact, and lightweight architecture for real-time ``semantic stereo matching'' (the term coined to refer to the combination of two tasks: depth estimation and semantic segmentation). The two tasks are accomplished by corresponding sub-networks, along with a disparity refinement network that uses the output from the predicted semantics. \cite{Wu2019} explored the use of \textit{pyramid} cost volumes, defined through multiple scales of feature maps, to better capture the disparity details in stereo matching, while also incorporating semantic segmentation to rectify disparity values along object boundaries. The latter task is achieved through the use of a single semantic cost volume (cost volumes are the set of correspondence costs per pixel used to infer the optimal disparity map). More recently, \cite{guizilini2020semantically} proposed a novel architecture that leverages a pre-trained semantic segmentation network to self-supervise monocular depth estimation via pixel-adaptive convolutions. 

\subsubsection{Jointly Learning Semantic Segmentation with Other Tasks}

The semantic segmentation process closely relates to other tasks required for robots and autonomous vehicles like depth estimation, and researchers have investigated jointly learning it with these other tasks. \cite{neven2017fast} developed a real-time implementation of ENet~\cite{Paszke2016} to simultaneously address three autonomous driving related tasks: semantic segmentation, instance segmentation, and depth estimation. They proposed a shared encoder but with different decoder branches for the three tasks. \cite{Kokkinos2017} presented a multi-task CNN to jointly handle a large number of functional tasks: boundary detection, normal estimation, saliency estimation, semantic segmentation, human part segmentation, semantic boundary detection, region proposal generation and object detection. They demonstrated effective scaling up of diverse tasks with limited memory and incoherently annotated datasets. \cite{Nekrasov2019} explored joint learning of two tasks, depth estimation and semantic segmentation, using a single model with an uneven number of annotations per modality. The latter was achieved using hard knowledge distillation through a teacher network. Through computational efficiency-related modifications of an existing semantic segmentation network, they achieved real time operation. They demonstrated their system's utility through 3D reconstruction based on SemanticFusion~\cite{McCormac2017} for both indoor and outdoor data. \cite{Yang2019} proposed a CNN to predict polarization information (per-pixel polarization difference) along with semantic segmentation using monocular RGB data, as opposed to using a micro-grid array polarization camera or polarized stereo camera. The work was motivated by the challenges of specular scene semantics like water hazards, transparent glass, and metallic surfaces, where polarization imaging often complements RGB semantic segmentation. Finally, as an extension of the Stixel representation~\cite{Badino2009}, \cite{Schneider2016} combined both depth and semantics per pixel for compactly representing an environment, benchmarking pixel-wise semantic segmentation and depth estimation tasks. For robots carrying a suite of sensors, joint learning based on input from these different sensing modalities can be leveraged to address increasingly complex and challenging tasks.

\subsection{Scene Representations}

Semantic representation of an observed scene can be defined at various levels of detail and is typically motivated by the application scenario. While image classification produces a single semantic label for an image, object recognition and semantic segmentation provide labeling at region- and pixel- level respectively. However, a scene can be represented in distinct ways that may or may not explicitly take into account the pixel- or region-level semantics, but still have a high-level representation which is semantically meaningful. We explore these additional scene representation approaches in this section, including scene graphs and zero-shot learning of generalized semantic scene representations.

The representation of a scene involves the extraction of visual information that can summarise the contents of the image captured from a particular scene~\cite{smeulders2000content}. For typical computer vision and robotics problems, imagery constitutes a broad domain where the variability is unlimited and unpredictable, even for images with the same semantic meaning~\cite{smeulders2000content}. Historically, the main goal of an effective image representation is the reduction of the semantic gap~\cite{santini1999user}, defined as the difference between the visual information extracted from an image and its high-level semantic interpretation.
Traditional image representations transformed the original image pixels into features spaces of color, shape or texture, with the goal of amplifying particular characteristics of the image that are relevant for representing the semantic information of the scene, while at the same time suppressing irrelevant information.
Such a goal can be partially achieved by promoting feature invariance properties that makes the representation robust to (generally) non-relevant distortions to the image, such as geometric transformation, illumination changes, and weather variations. One important trade-off encountered in doing so was between discrimination power and robustness to distortions. So much of the classical work on the design of feature representations was significantly influenced by the goal of finding an effective operating point that balanced discriminative and invariance properties.
\textit{Color} invariance is typically achieved by transforming the original RGB (Red Green Blue) space into a more robust HSV (Hue Saturation Value) space (other similar spaces have also been proposed), where the hue channel is invariant to the illumination and camera direction with respect to an object's orientation~\cite{smeulders2000content}. The invariance to \textit{shape information} is commonly obtained by computing image derivatives using scale- and rotationally-invariant operators~\cite{freeman1990steerable}. \textit{Texture} invariance is generally obtained by representing an image in the frequency domain~\cite{smeulders2000content}.

The classical representation of images involves the partitioning of the image into regions, and the computation of features from each region. Image partitioning can be done in a dense manner~\cite{smeulders2000content}, where the whole image is broken into regions of fixed~\cite{carneiro2007supervised} or variable~\cite{achanta2012slic} size, or in a sparse manner~\cite{lowe2004distinctive}, with the selection of salient image patches. Features (including color, shape and/or texture) are then extracted from the region and summarised into a fixed dimension representation using, for example, a particular spatial distribution~\cite{schmid1997local} or a histogram~\cite{swain1992indexing}. Images can then be represented by a collection of these features~\cite{smeulders2000content}. 

Some of the most prominent image representations in computer vision and robotics before deep learning representations largely replaced them~\cite{krizhevsky2012imagenet}, were based on Bag of Visual Words (BoVW)~\cite{sivic2003video}. In brief, these methods were based on the sparse image partitioning of the image, where each region is labeled as a particular visual word, with visual words being learned from a large collection of image regions. The image is then represented by a histogram of the visual words. 
For example in the place recognition domain (again, prior to the use of deep learning representation approaches), Williams \textit{et al.}~\cite{williams2009comparison} presented a comparison of different image representations, concluding that methods based on BoVW were generally superior. In the following subsections, we discuss two distinct ways of representing a scene: Scene Graphs and Generalized Semantic Representations, both of which are critical to semantic scene understanding for robotics.

\subsubsection{Scene Graphs}
Understanding a visual scene fully requires knowledge beyond just what is present in the scene. Much of the meaningful information can be extracted by examining the \textit{relationships} between objects and other components of the scene, which is one of the main motivations of the scene graph research area.

Lin \textit{et al}.~\cite{Lin2013} developed a CRF-based method to address indoor scene understanding from RGB-D data. The method integrates information from 2D segmentation, 3D geometry, and contextual relations between scenes and objects to classify the volume cuboids extracted from the scene. With this formulation, scene classification and 3D object recognition tasks are coupled and can be jointly solved through probabilistic inference. Johnson \textit{et al}.~\cite{Johnson2015} proposed scene graphs that represent objects, attributes of objects, and the relationships between them in order to retrieve semantically related images. Their method used CRFs to ground the scene graphs to local regions in the images. As an extension of~\cite{Johnson2015}, Schuster \textit{et al}.~\cite{schuster2015generating} explored the scope of scene graphs created automatically from a natural language description of a scene, using rule-based and classifier-based parsing to improve image retrieval. Following the work by~\cite{Johnson2015}, Xu \textit{et al}.~\cite{Xu2017} developed a method to generate scene graphs by end-to-end learning, using an RNN and iterative message passing. The main focus on this work was on improving reasoning about spatial relationships based on surrounding contextual cues within an image. 

Li \textit{et al}.~\cite{Li2017} presented a Multi-level Scene Description Network (MSDN) that jointly learns to leverage mutual connections across three semantic levels of scene understanding: object detection, scene graph generation, and region captioning. Zellers \textit{et al}.~\cite{Zellers2018} explored a motifs-based scene parsing - looking for repeated patterns or regularly appearing substructures, in this case within graph representations of the scene. In particular, they proposed Stacked Motif Networks to capture higher order motifs in scene graphs, focusing on global contextual information to inform local predictors of objects and relations. Herzig \textit{et al}.~\cite{Herzig2018} developed a method that constrains the neural network architecture to be invariant to structurally identical inputs, based on the conditions of the Graph-Permutation Invariance (GPI). Johnson \textit{et al}.~\cite{Johnson2018} presented a method to generate images from complex sentences using scene graphs, that enables explicit reasoning about objects and their relationships. Their method uses graph convolutions to compute scene layout, that is then converted into an image using a cascaded refinement network trained adversarially against a pair of discriminators.

More recently, Ashual and Wolf~\cite{ashual2019specifying}
introduced a method that generated multiple diverse output images per scene graph using scene layout and appearance embedding information. Chen \textit{et al}.~\cite{chen2019scene} explored a semi-supervised method that automatically estimated probabilistic relationship labels for unlabeled images using a small number of labeled examples. The goal of this work was to address a recurring issue that affects much of the research in this field; that the models have been trained with small visual relationships sets, with each relationship having thousands of samples. Addressing the issue of scene graph generation dataset biases, Gu \textit{et al}.~\cite{gu2019scene} introduced a scene graph generation method that uses external knowledge and image reconstruction loss to reduce this bias. To further address these biases, zero-shot learning techniques have been proposed that aim to achieve a generalized scene representation, which are discussed in the next subsection. 

\subsubsection{Semantic Generalized Representations via Zero-Shot Learning}
A common practical issue in robotics is that the system can only be exposed to a subset of the classes it is expected to encounter during actual deployment. Zero-shot Learning (ZSL) addresses this problem: defined as a learning problem, it focuses on the scenario where the training process has access to the visual and semantic representations of only a subset of the classes. This observed subset is labeled as the \textit{seen} classes; the challenge during deployment and testing involves the classification of \textit{unseen} visual classes that do not belong to the set of \textit{seen} classes~\cite{xian2018zero}. Zero-shot learning generally depends on the learning of a mapping from the visual to the semantic space using the \textit{seen} classes~\cite{lampert2013attribute,Palatucci2009,xian2018zero,Zhang2015}, where the hope is that this learnt mapping can be reliably used to classify the \textit{unseen} classes. 
\cite{Chen2018} proposed SP-AEN for zero shot learning to tackle the semantic loss problem in ZSL. Acknowledging that \textit{classification} and \textit{reconstruction} are contradictory objectives, an independent visual-to-semantic embedding was introduced for both the tasks. Both these semantic embeddings were used for adversarial learning, and to transfer information between the two tasks.

An alternative to mapping from the visual to the semantic space is to assume the availability of the semantic representations of the unseen classes during training. This assumption allows the implementation of a conditional generative model that is trained to generate visual representations from their semantic representations using the seen classes. Once the generator is stable, it is then possible to generate unseen visual representations which can be used to directly train a visual classifier with real \textit{seen} visual representations and generated \textit{unseen} visual representations~\cite{kumar2018generalized}. This method can work better than others for the problem of \textit{Generalised} Zero-Shot Learning (GZSL), where the testing process involves the classification of seen and unseen classes. GZSL is more challenging than vanilla ZSL because the classifier tends to be severely biased towards the classification of seen classes. Much of the current research in the field is therefore focused on achieving a balance between the classification of seen and unseen classes. The general theme of this research is to combine the semantic and visual representations, with modulation of the classification of seen and unseen classes based on the input test visual sample. For example, Felix \textit{et al}.~\cite{felix2018multi} proposed a cycle consistency loss to generate visual representations from semantic ones, and then in reverse generate the semantic representation. The modulation of seen and unseen classes has also been addressed by Atzmon and Chechik~\cite{atzmon2019adaptive}. One can also learn a joint semantic and visual space that mitigates the need for learning a mapping between these two spaces~\cite{Felix2019}. The classifier is learnt in conjunction with a domain classifier to differentiate between seen and unseen domains. Niu \textit{et al}.~\cite{Niu2019} proposed to reduce the \textit{projection domain shift}, defined as the lack of generalization of visual-semantic mapping based on seen categories to unseen categories. Their approach involved learning an adaptive mapping for each unseen category, followed by progressive label refinement using unlabeled test instances. More recently, \cite{vyas2020leveraging} proposed LsrGAN, which explicitly transfers the knowledge of semantic relationships between seen and unseen classes by generating corresponding mirrored visual features.

The overarching goal of object detection, semantic segmentation and scene representation is to enable effective scene understanding. It can be commonly observed in all these fundamental areas of research that the use of additional sources of information in the form of sensor data, knowledge repositories and textual descriptions can significantly enhance the task performance. This is highly relevant to robotics, as a robot typically carries multiple sensors, can \textit{actively} explore its environment and has access to additional resources from the ``cloud'' online or on a local network. These tasks also have high functional overlap with those required of robots in many applications. In the following sections, we discuss how  semantic understanding approaches like those discussed so far can be used in robotics, and the associated further research advances required to do so.


\section{DYNAMIC ENVIRONMENT UNDERSTANDING AND MAPPING}
\label{secn:mappingSLAM}
Mapping, localization, and navigation are among the longest running research areas in mobile robotics, as key capabilities for many autonomous systems operating in the sky, in water, or on land. In neuroscience and biology, there has been a long running debate about what animals explicitly construct maps for navigation and to what extent they do so, versus achieving navigation through behavioural and reactive techniques~\cite{epstein2017cognitive}. Likewise in robotics, there have been multiple research streams in the robot navigation space; some involving reactive type techniques like behaviour-based robotics~\cite{brooks1989robot, braitenberg1986vehicles, maes1990learning} but many involving the explicit construction of a map and localization of a robot within that map, in order to plan and execute navigation tasks. Although the field has been an active one for many decades, much of the pivotal work in \textit{modern} mapping systems occurred in the late 90s and early 2000s with the advent of Simultaneous Localization And Mapping (SLAM) -- which is interchangeably referred to as both a problem field and desired capability.

Synonymous with the development of modern SLAM techniques was the widespread utilization of sonar and later laser range sensors~\cite{durrant2006simultaneous, bailey2006simultaneous,thrun2002probabilistic}, which facilitated the production of highly accurate, geometric maps of the environment such as occupancy grid maps. SLAM and geometric maps enabled a significant range of navigational capabilities for robots actually deployed in the real world, especially in domains like mining and logistics~\cite{zlot2014efficient, wolcott2014visual, guivant2004navigation, schreiber2013laneloc, jacobson2020localizes}. But as demand grew for more sophisticated autonomous systems that could understand and interact in richer ways with their environments~\cite{jain2019discrete, frossard2019deepsignals, homayounfar2019dagmapper}, and the robots and people occupying those environments, researchers have been focusing on enriching mapping representations -- which is where semantics has played a key role.

In recent years semantic SLAM~\cite{rosinol2020kimera, Salas-Moreno2013}, semantic mapping~\cite{McCormac2017, Sunderhauf2017} and semantically-informed localization~\cite{Gawel2018, garg2018lost} have all emerged as major new areas of focus in this research field. Initial forays into semantic approaches were largely based on adding semantic labeling or segmentation on top of existing traditional map representations~\cite{Pronobis2012, Salas-Moreno2013, Sunderhauf2017}, but as the field has progressed semantics has become an increasingly integral component of techniques~\cite{rosinol2020kimera, rosinol20203d}. Before we review these systems, it is appropriate to first briefly revisit some of the key background in robotic mapping and navigation research.

\subsection{A Brief History of Maps and Representations}

A practical SLAM system typically employs a combination of interoceptive sensors (e.g., rotary encoders, accelerometers, inertial measurement devices) and exteroceptive sensors (e.g., LiDAR, sonar, radar, cameras). The mapping functionality in SLAM derives largely from the capability of the exteroceptive sensors to \textit{measure} (directly or indirectly) structural or visual elements in the scene. While LiDAR and sonar were dominant sensing modalities in the early days of SLAM~\cite{durrant2006simultaneous, bailey2006simultaneous,thrun2002probabilistic}, the use of cameras (e.g., standard RGB cameras, depth cameras, and event cameras) as the primary sensors for SLAM (thus leading to visual SLAM) is currently a major area of interest in robotics and computer vision. A key reason behind the popularity of visual SLAM is the flexibility and relatively low cost of optical sensing devices (e.g. consumer RGB and depth cameras). Moreover, compared to other exteroceptive sensors, cameras provide a richer source of information and thus the prospect of extracting higher-level understanding (e.g. semantics) of the scene.

\subsubsection{Classical 3D Maps}
Classical visual SLAM methods have largely focused on extracting the geometric 3D structure of the environment. Feature-based methods~\cite{davison03,davison07,klein2009parallel,mei09,strasdat12,mur-atal15} (both monocular and stereo) construct maps in the form of sparse 3D point clouds, where the 3D points are typically reconstructions of the salient local features detected in the images. To facilitate the generation of 2D-3D correspondences, descriptors of the local features associated with the reconstructed 3D points are also often embedded in the map. In contrast to feature-based methods, direct methods~\cite{newcombe2011dtam,engel13,forster14,Engel-et-al-pami2018} utilize a photometric error formulation to estimate structure and motion. Maps produced by direct SLAM methods are typically semi-dense or dense 3D point clouds, where each reconstructed 3D point is associated with corresponding pixels (that are not necessarily locally salient) observed across multiple frames in the input sequence. For visualisation, the point clouds are often texture-mapped with pixel RGB values. In contrast to conventional cameras, depth cameras~\cite{geng11} such as the Microsoft Kinect are able to directly acquire depth information at frame rate (such cameras are often called RGB-D cameras since they also record in the RGB channels, in addition to depth). Accordingly, RGB-D SLAM algorithms~\cite{HenryKHRF10,newcombe11kinectfusion,EndresHESCB12,KahlerPRSTM15,InfiniTAM_ECCV_2016} are able to construct dense volumetric 3D maps of the environment. Since depth cameras can acquire metrically consistent depths, the 3D maps generated via RGB-D SLAM algorithms do not suffer from the global scale ambiguity problem which affects monocular visual SLAM systems.

Classical visual SLAM methods (as outlined above) have reached a level of maturity where accurate 3D maps (sparse or dense point clouds) and localisation of the observer within the map can be efficiently computed (though some require hardware accelerators such as GPUs). However, most practical robotics applications require more than just 3D maps; semantic SLAM is a cogent and timely extension to classical visual SLAM.

\subsubsection{Topological Maps} 
Topological maps represent an environment as an abstract graph~\cite{brooks1985visual}, where nodes represent distinct places a robot has visited and edges between the nodes represent topological relations like proximity and order~\cite{kortenkamp1994topological}. One of the limitations of classic metric maps had been the accumulation of error (drift) in global coordinates, despite the use of multiple sensors, beacons and elaborate error-tracking systems~\cite{giralt1983integrated}. Topological maps do not suffer as directly from the accumulation of movement errors, as the robot has only to navigate locally between adjacent place nodes, although drift is still an issue for loop closure. Furthermore, topological maps are highly scalable when compared to classical metric maps that use detailed a priori models of the world~\cite{kosaka1992fast, fennema1990experiments}. Topological SLAM systems have been widely explored in the past where the use of multi-sensory approaches~\cite{choset2001topological} has gradually been replaced with visual similarity-based metric error correction via loop closures~\cite{levin2004visual, newman2005slam, se2005vision}. One of the key enablers for relatively modern topological mapping systems has been the development of Bag-of-Visual-Words (BoVW)~\cite{sivic2003video} like methods, leading to robust large scale appearance-based topological SLAM systems like FAB-MAP~\cite{cummins2008fab}. Existing surveys on topological maps~\cite{garcia2015vision} and visual place recognition~\cite{Lowry2016} provide further details about developments in this field in recent years. 

\subsubsection{Hybrid Approaches}
Hybrid map representations have been demonstrated to achieve an ideal balance between classic metric maps and topological maps~\cite{kuipers1991robot}, where a map is represented as a hierarchy. In such representations, metric information from the local geometry of the scene is incrementally fused into a global ``topometric" map, which is defined at a large scale using topological relations between distinct places (the hierarchy is almost always globally topological and locally metric rather than the reverse). Thrun~\cite{thrun1998learning} proposed an integration of grid-based and topological maps, where the latter partitions the former into coherent regions. Simhon and Dudek~\cite{simhon1998global} explored a hybrid map representation where local metrically accurate maps, dubbed ``islands of reliability'', form nodes of a topological model of the world, thus avoiding the need to perform large-scale error integration. The \textit{Atlas} framework developed by \cite{bosse2003atlas} comprised an interconnected set of local coordinate ``frames''. Each frame is a local metric map of the environment, connected to other frames via transformations represented as edges in the global graph of coordinate frames. \cite{tomatis2003hybrid} proposed a hybrid map representation, based on extracting ``corners'' and ``openings'' that represented topology, and ``lines'' that represented local geometric structure using a $360^{\circ}$ laser scanner. \cite{mei2010closing} explored a landmark-based co-visibility graph representation of the environment, where co-visibility corresponds to the connectivity of a topological map and inter-frame motion is used to encode metric transforms between landmarks.

More recently, hybrid maps have been explored beyond the bounds of two-layer hierarchies, and have started to introduce semantic concepts. \cite{Pronobis2012} developed a probabilistic framework based on chain graphs to create a hierarchical hybrid map comprising four layers: sensory (accurate metric map), place (places and paths information), categorical (geometry and appearance of objects and landmarks), and conceptual (instances of spatial concepts relating a cereal box with kitchen). The proposed system used laser and camera sensors in indoor environments and attempted to relate conceptual knowledge with object and place semantics. \cite{Riazuelo2015} presented a cloud service-based semantic mapping system comprising an ontology to code concepts/relations in maps and objects (CAD models), built on top of an RGB-D metric map using both keyframe-based and 3D occupancy grid-based map representations. The proposed system enabled semantic mapping of novel environments and searching for novel objects within a semantic map. In \cite{Kochanov2016}, authors used stereo frames to obtain depth, scene flow, visual odometry and semantic segmentation, all of which formed the input for semantic mapping based on 3D occupancy. This enabled reasoning on objects, and led to object instance discovery based on temporally consistent shape, appearance, motion, and semantic cues in the map, while also being able to handle dynamic objects. More recently, \cite{yue2020hierarchical} presented a hierarchical framework for probabilistic semantic mapping using multiple cooperative robots in a distributed setting. Although hybrid maps, particularly those involving a semantic layer, are more amenable to human-robot interaction, they are still in a phase where certain implementation choices might limit their universal applicability. It's still unknown whether, given the complexity of robotic tasks and applications, a generalized solution is viable, or whether specialized solutions will be required for each class of applications.

\subsection{Places and Objects for Semantic Mapping}
Semantic mapping has been explored both in the context of place-level and object-level representation of the environment, where a more detailed semantic representation typically combines place and object labels in a hierarchical manner to create a hybrid map, as also discussed in the previous subsection. In this subsection, we first cover the use of scene classification and place categorization, where researchers have focused on abstracting the changing appearance of the scene into meaningful place labels as the robot explores an environment. We then discuss one of the clearest opportunities to enrich robot mapping systems with the use of \textit{objects}. Since objects can be categorized and are meaningful by themselves, all this information can be used in a variety of ways by semantic mapping and localization approaches. In our discussion below, we further split the objects-based approaches into two categories: those that use prior knowledge of the expected objects, often including 3D models, and those that have limited or no knowledge available beforehand, instead learning how to use objects at deployment time. 

\subsubsection{Scene Classification and Place Categorization}

Semantic maps of an environment can be constructed by categorizing places with semantic labels which are typically pre-defined. Such semantic categories for different places can be defined by only considering the functionality of that place, for example, ``kitchen'', ``printer area'', and ``seminar room''~\cite{Rottmann2005}. Alternatively, a more general hierarchical framework approach can also consider the structural properties, for example, having a broad classification based on ``room'' and ``corridor'' be followed by specialization labels of rooms like \textit{office} and \textit{classroom}~\cite{Luperto2016}.
~\cite{Rottmann2005} combined vision and range information to extract objects (e.g. monitor and coffee machine) and geometric features, respectively, which were fed to AdaBoost~\cite{freund1995desicion} to classify places and perform efficient global localization.
~\cite{Stachniss2006} used semantics as ``background knowledge'' to explicitly represent environments with corridors and other indoor structures. Such a semantic distinction led to efficient multi-robot exploration of an environment, through learning of a behaviour which rewarded robots to preference exploring corridors, which led to unexplored branches of connecting rooms.

\cite{Goeddel2016} developed a CNN-based place classification system trained using black and white images of occupancy grid maps (black uncarved space vs white carved free space) obtained through 2D laser scan data. The semantic classes used for the task included corridor, doorway, and room. \cite{Sunderhauf2016} presented a CNN-based semantic mapping system that overcame the closed-set limitations of supervised classification by complementing the system with one-vs-all classifiers, in order to recognize new semantic classes online. The proposed system used Bayesian filtering to incorporate prior knowledge and ensure temporal coherence. Furthermore, they also demonstrated the effective use of semantics for improving object recognition and modulating a robot's behavior during navigation tasks. \cite{Liao2016} presented a scene classification CNN that incorporates object-level information by regularization of semantic segmentation, demonstrated on indoor RGB-D data. \cite{Premebida2017} proposed a Dynamic Bayesian Mixture Model (DBMM) -- a mixture of heterogeneous base classifiers -- that incorporates time-based inferences from previous class-conditional probabilities and priors. Their system used 2D laser scans and indoor data for experiments, and extended their prior work~\cite{Premebida2015}. \cite{Mancini2017} explored the use of fully-convolutional CNN for learning better feature representations for the task of semantic place categorization. This system used a Naive Bayes Nearest Neighbor (NBNN) method within the learning framework for end-to-end training. \cite{Mancini2018} developed a novel method for domain generalization for place classification to deal with the unknown/unseen deployment scenarios, when the test data might not be similar to the data used for training. In this case, domain generalization was achieved by automatically computing a model for the unknown domain through combining models of the known domains. \cite{Zheng2018} presented \textit{TopoNets} for semantic mapping based on the topological structure of the environment, using a Sum-Product Network (SPN) as the backbone for learning and inference. The proposed system was demonstrated on various tasks: place classification, inferring semantics of unexplored space, and novelty detection using the COLD dataset, which includes semantic classes like doorway, kitchen, office, bathroom and laboratory.

Most of the aforementioned place categorization systems are based on supervised learning, consider only a finite set of pre-defined place labels, and therefore are not suitable for recognizing newly encountered places. \cite{Ranganathan2012} developed an online Bayesian change-point detection framework: PLISS (Place Labeling through Image Sequence Segmentation), enabling discovery of novel place categories, along with uncertainty estimation through consideration of the spatio-temporal characteristics. The semantic place labels obtained in this manner were further combined with high-level information like adjacency and place boundaries using Conditional Random Field (CRF), in order to obtain a semantic map of the environment~\cite{Ranganathan2011}. \cite{Fazl-Ersi2012} proposed an alternative approach to addressing open-set place categorization, HOUP (Histogram of Oriented Uniform Patterns), as an image descriptor. Although the framework does not create new semantic classes, the proposed representation is demonstrated to exhibit a balance between strong discriminative power for (specific) place recognition, and generalization capability for place categorization.

\subsubsection{Using Pre-defined 3D Object Models}

The ability of a robot to track objects in its working environment is critical for performing many tasks. Approaches based on scan registration were among the earliest to emerge in this domain, such as the work presented in~\cite{Stuckler2012}. They proposed to learn and track 3D object models from RGB-D indoor data by aligning multiple views of these objects within multi-resolution ``surfel'' maps. Real-time alignment on a CPU was achieved using a probabilistic optimization framework and an efficient variant of ICP. Other approaches have built a point-cloud based 3D map, and then identified objects within it as in~\cite{Wei2012}. The authors developed a shape-based method to build a 3D semantic map using point cloud data from an RGB-D camera. Data was segmented based on planes (horizontal/vertical), and an a-priori model library was used to identify objects and extract object features; this information was then used for wheelchair navigation. Another example can be found in~\cite{Gunther2013}, where the authors developed a mapping-only framework based on indoor RGB-D data that created a triangle mesh for extracting and classifying planar regions as different furniture objects. The recognized objects were then replaced by their corresponding 3D CAD models following ICP alignment. Their system used OWL-DL (Web Ontology Language - Description Logic) and SWRL (Semantic Web Rule Language) as the ontology for defining object-property relations. \cite{Vasudevan2008} presented a hierarchical mapping approach based on detecting and clustering different objects while also considering their spatial relationships using a Bayesian classifier. For example, an \textit{office} is considered to be constituted by ``work-space'', ``meeting-space'', and ``storage-space'', each of which are further composed of several objects.

Many researchers have focused on the ability to build a semantic map while also being able to track objects in the map. In~\cite{dame2013cvpr}, the authors demonstrated the use of object-specific knowledge to obtain accurate maps within a dense SLAM system. They also highlighted how 3D object tracking and 3D reconstruction could benefit each other, thus improving reconstruction of unseen parts and enabling accurate estimation of the scale of the map. \cite{Salas-Moreno2013} presented SLAM++, performing 3D object recognition and tracking to produce an explicit graph of objects (with 6-DoF poses), which is then used in a pose-graph optimization framework for instance-level object-oriented 3D SLAM. The authors used a database of 3D object models and performed relocalization and loop closures in large cluttered environments, while also enabling interaction with objects.

Researchers have also explored the possibility of performing 3D object detection and recognition using a pre-built 3D map. 
\cite{Tateno2016} proposed incremental real-time segmentation of a 3D scene, reconstructed by SLAM, in order to perform 3D object recognition and pose estimation. They also highlighted the advantages of using multiple views of the objects, as opposed to single-view based (2.5D) object recognition. Their system used 3D models and was demonstrated through an AR application. \cite{Dong2017} developed semantic representation of an environment based on sparse point clouds, provided by a SLAM process, and semantic object detections, for example, cars detected through YOLO~\cite{redmon2016you}. They used vision and inertial sensors (accelerometer and gyrometer - now ubiquitous in phones and drones) to obtain semantic and syntactic attributes respectively. These representations were fed to a localization-and-mapping Bayesian filter to enable persistent object representation and re-detection of temporarily occluded objects.

\subsubsection{Without Pre-defined Object Models}

Without explicit knowledge about some or all of the objects that are encountered in an environment, robots must be equipped with a means by which to deal with novel objects. 

\cite{Rogers2012} presented a CRF model to jointly classify objects and room labels on a mobile robot, by incorporating information from both recognition of trained objects and classification of novel objects. They created a map with topologically connected rooms and metrically connected object poses using SURF features for object recognition. \cite{Pham2016} proposed an unsupervised geometry-based approach for segmentation of 3D point clouds into objects and meaningful scene structures, which form a high-level representation of a 3D geometric map. They also developed a novel global plane extraction algorithm that enforced planes to be mutually-orthogonal or parallel, conforming with man-made indoor environments. \cite{McCormac2018} presented Fusion++, an online object-level SLAM system with a 3D graph map of arbitrary reconstructed objects, where objects are incrementally refined via depth fusion, and are used for tracking, relocalization and loop closure detection, without intra-object warping. The proposed pipeline uses Mask-RCNN for instance segmentation that initializes per-object TSDF reconstruction, and was demonstrated on RGB-D indoor sequences. 

The open set recognition research field is also relevant here~\cite{scheirer2012toward, li2005open, phillips2011evaluation, scheirer2014probability, bendale2015towards, bendale2016towards}, and some existing mapping work has focused on dealing with open set conditions. \cite{Sunderhauf2017} built a semantic map with both object-level and low-level (point and mesh based) geometric representations that functions under open-set conditions and handles unseen instances. Their pipeline involves feature-based RGB-D SLAM, deep-learnt object detection, and 3D unsupervised segmentation~\cite{Pham2016}. \cite{Furrer2018} incrementally builds a database of object models from a traverse of a mobile agent requiring no prior knowledge of shapes or objects present in the scene. The presented pipeline includes: Global Segmentation Map (GSM) built from RGB-D images, object-like segment extraction, intra-segment matching and merging with previous instances in the database, and reconstruction of unobserved parts of the scene from merged models. \cite{Grinvald2019} incrementally builds volumetric object-centric maps using a RGB-D camera, while also reasoning jointly over geometric and semantic cues using a frame-wise segmentation approach. Their system infers high-level category information about detected and recognized elements, and discovers novel objects in the scene without requiring prior information about the objects. The proposed method also enables a distinction between unobserved and free space for enhanced human-robot interaction.

Semantic mapping pipelines are not typically computationally cheap. While continual improvements in compute hardware help with this issue, as covered in Section \ref{secn:discussion}, efficiency is always beneficial with respect to cost, power consumption and deployment versatility. Research has therefore focused on improving absolute efficiency and scalability to larger environments. \cite{Tateno2015} developed a real-time incremental segmentation method for 3D point clouds obtained through SLAM, yielding segmentation in real-time, with complexity independent of the size of the global model. The proposed method is generally applicable to any frame-wise segmentation and any SLAM algorithm and was demonstrated in indoor environments. \cite{Vineet2015} presented dense, large-scale, outdoor semantic reconstruction of a scene in (near) real time that was also capable of handling dynamic objects through semantic fusion. They used hash-based techniques for large-scale fusion and efficient mean-field inference with dense CRFs, claiming it to be the first of its kind. \cite{Nakajima2019} presented highly accurate object-oriented scene reconstruction in real-time by using fast and scalable object detection for semantics and geometric incremental segmentation. They reduced computational cost and memory footprint by only labeling segmented \textit{regions} and not individual elements in the 3D map. \cite{Pham2019} performed on-the-fly dense reconstruction and semantic segmentation of 3D indoor scenes using an efficient super-voxel clustering, and a CRF based on higher-order constraints derived from structural and object cues.

A reliance on objects brings with it new challenges, with one of the largest being the inconvenient property of them being movable. \cite{Runz2017} presented a dense RGB-D SLAM system that segments the scene into different objects using either motion or semantic cues while tracking and reconstructing their 3D shape in real time. It allows objects to move freely by fusing its shape over time using only the pixels associated to that object label. Consequently it is able to deal with dynamic scenes without treating moving objects as outliers, as was the approach in much prior research. \cite{Runz2019} presented MaskFusion, a real-time, object-aware RGB-D SLAM system that recognizes, segments, and assigns semantic labels to different objects, even if multiple objects are moving. The proposed system uses image-based, instance-level semantic segmentation to create an object-level semantic map, unlike the voxel-level representations used in prior work. More recently, \cite{di2020unified} developed a novel framework for dense piece-wise semantic reconstruction of dynamic scenes using motion and spatial relations, where moving objects are handled by imposing constraints based on the spatial locations of neighboring superpixels.

Shape is an important property when dealing with objects, resulting in research focusing on using parameterized geometric primitives. \cite{Hashemifar2017} represented indoor environment objects as cuboids for semantic mapping; the cuboid detection method is based on image segmentation and plane fitting, and the cuboid matching is based on features like emptiness, orientation, surface coverage, and distance from edges. \cite{Nicholson2019}~presented QuadricSLAM, a factor graph based SLAM system that uses dual quadrics to represent 3D landmarks, derived from 2D object detections obtained over multiple views. They proposed a new geometric error formulation while also addressing the challenges of object occlusions. Building on this work,~\cite{hosseinzadeh2018structure} integrated additional planar and point constraints that help stabilise the SLAM estimate. Later,~\cite{hosseinzadeh2019real} explored how single view point cloud reconstructions of objects (via a CNN) can effectively constrain the shape of dual quadric landmarks. \cite{Drews2010} developed a framework to detect and compactly represent changes in the environment. This is achieved though multi-scale sampling of point cloud data, change detection using Gaussian Mixture Models, a superquadrics-based representation of objects that caused the change, and final refinement and optimization. Other related research includes research on cuboids from~\cite{Lin2013} as well as basic primitives from~\cite{Kaiser2015}.

\subsubsection{Scene Graphs at Environment Level}
A dense grid representation of an image or an environment can be thought of as a specific case of a graph structure. We discussed the use of scene graphs previously at image level based on spatial relationships between various objects or regions observed in a single image. In a robotics context, the concept of a scene graph typically comprises spatial elements at an environment level which the robot explores over time. In this vein, a number of researchers have explored the use of scene graphs for better representation of the environment~\cite{Blumenthal2013, Aksoy2010, Grotz2017, Kim2019, Zeng2018, Gawel2018, rosinol20203d}, leading to improved spatial reasoning.

Using the concept of a Directed Acyclic Graph (DAG)~\cite{Shuey1986}, \cite{Blumenthal2013} proposed a 3D scene representation, Robot Scene Graph (RSG), which defines the organization of topological and spatial relations between objects, semantics of such relations, time-based handling, computational assets, and resource sharing. \cite{Blumenthal2014} extended this work with a Domain Specific Language (DSL) with four levels of abstraction for RSG~\cite{Blumenthal2013}, used for model-driven engineering tool chains in robotics. RSG-DSL is capable of expressing (a) application specific scene configurations, (b) semantic scene structures and (c) inputs and outputs for the computational entities that are loaded into an instance of a world model. \cite{Kim2019} presented a 3D scene graph constructed using a RGB-D data processing pipeline: keyframe extraction, spurious detection rejection (object detection based), local 3D scene graph construction, and finally, graph merging and updating for a global 3D scene graph. The proposed method was demonstrated on two tasks: Task Planning and Visual Question and Answering. \cite{Bozcan2019} developed a Boltzmann Machines-based generative scene model for representing objects and their spatial relations and affordances while also considering co-occurrences. In order to solve the cross-view localization problem, \cite{Gawel2018} used a graph of semantic blobs defined over a sequence of images, with a random walks strategy used to match a query sequence. \cite{pan2020spatio} demonstrated a video captioning system based on a spatio-temporal scene graph that explicitly captures object interactions using directed temporal edges and undirected spatial edges. 

Recently, \cite{rosinol20203d} proposed 3D Dynamic Scene Graphs (DSGs) which define an environment with multiple layers of abstraction, starting from a metric-semantic mesh, to objects, places and rooms and eventually to the whole building. While the aforementioned entities form the nodes in the directed graph, the edges encode pairwise spatio-temporal relations and explicitly model dynamic entities in the scene like humans and robots. DSGs represent the current state-of-the-art in terms of a high-level actionable representation of the environment, where robots can semantically reason about the space they are operating in and interact with humans.

\subsection{Semantic Representations for SLAM and 3D Scene Understanding}
\label{sec:semRep+SU+SLAM}
\subsubsection{SLAM with Semantic Segmentation}

\cite{flint2010cvpr} proposed a semantically meaningful indoor mapping system based on the Manhattan World assumption, where photometric cues were combined with pose information and sparse point cloud data obtained from an underlying metric SLAM system. In a subsequent work~\cite{flint2010eccv}, the authors developed a more efficient approach based on dynamic programming to label the indoor environment as floor, wall or ceiling. Further extending the use of Manhattan world, \cite{flint2011iccv} presented a joint inference procedure based on a Bayesian framework that combines photometric, stereo and 3D data to reason about floor and ceiling planes, and thus enable effective semantic scene understanding.

Beyond reasoning at the surface level, researchers have also explored 3D structure-based semantic representations. \cite{Kundu2014} proposed a 3D occupancy map based on joint inference of 3D scene structure and semantic labels for outdoor monocular video data. They used a CRF model defined in 3D space and class-specific semantic cues to constrain the 3D structure in areas where multi-view constraints are weak. \cite{Hermans2014} described a process for transferring labels from 2D to 3D based on Bayesian updates and dense pairwise 3D CRFs for indoor RGB-D data, combined with a fast 2D semantic segmentation approach based on Randomized Decision Forests. \cite{Stuckler2015} integrated multi-view image segmentations within an octree-based 3D map by modelling geometry, appearance, and semantic labeling of surfaces for indoor RGB-D video. 

For robotics applications, real-time operations are typically necessary. Hence, some of the research work has particularly been focused on efficient processing. The pipeline proposed in~\cite{Stuckler2015} is based on random decision forests and probabilistic labeling using a Bayesian framework, and performs in real time or better on CPU and GPUs. \cite{Li2016} developed a scale-drift-aware, monocular semi-dense mapping system that can seamlessly switch between indoor and outdoor scenes where previous methods struggled. The system also saves computation time by enabling frame skipping for 2D segmentation, and through only considering keyframe connectivity and spatial consistency. \cite{McCormac2017} presented \textit{SemanticFusion}, a dense semantic 3D mapping system for indoor RGB-D data where semantic predictions from a CNN are probabilistically fused from multiple viewpoints using ElasticFusion (Dense SLAM)~\cite{Whelan2015}. The fusion technique also improves 2D semantic labeling performance over single frame predictions. The proposed system works in real time at 25 Hz. \cite{Xiang2017} combined CNN-based 2D semantic labeling with RNN-based data association (found using ElasticFusion~\cite{Whelan2015}) for dense semantic mapping on RGB-D indoor data. \cite{Yang2017} presented a 3D scrolling occupancy grid map, achieving near real time performance, with relatively low memory and computational requirements and an upper bound as environment size scales. To achieve this, they used a novel hierarchical CRF model with CNN-based 2D segmentation to optimize 3D grid labels on top of a stereo ORB-SLAM based grid map, and used superpixels to enforce smoothness. 

\cite{Hazirbas2017} proposed semantic labeling of indoor RGB-D data based on an encoder-decoder type network with two branches, one each for RGB and Depth, which improved performance when combined together. They found that the ``HHA'' representation~\cite{gupta2014learning} of depth images -- encoding \textit{H}orizontal disparity, \textit{H}eight above ground, and the \textit{A}ngle the pixel’s local surface normal makes with the inferred gravity direction -- improved performance. Both sparse and dense fusion also improved performance. \cite{Ma2017} used a CNN to predict semantic segmentation for RGB-D sequences. They trained the CNN to predict multi-view consistent semantics in a self-supervised way, enabling improved fusion at test time -- enhancing \cite{Hazirbas2017} with multi-scale loss minimization. Their system performed better than single-view baselines.
Kimera~\cite{rosinol2020kimera} is a real-time metric-semantic SLAM system that works in real-time and integrates 2D semantic segmentation, IMU measurements, and optional depth measurements into a dense semantically annotated mesh of the environment.

\subsubsection{Semantic Scene Understanding using 3D Point Clouds}
\label{sec:sem+3DPointCloud}

While 3D scene information can (somewhat laboriously) be derived from scale-unaware monocular visual SLAM algorithms for online mapping~\cite{newcombe2011dtam, engel2014lsd, mur-atal15, weerasekera2017icra}, a common avenue for estimating pixel depth in robotics research is to use range sensors, for example, Laser Scanners~\cite{nuchter2008towards,Xiong2010}, stereo camera pairs~\cite{Morreale2019}, LiDAR, radar, sonar or RGB-D sensors (like Microsoft Kinect~\cite{zhang2012microsoft}, Intel RealSense, Apple PrimeSense and Google Tango)~\cite{Wolf2015,Boulch2018}. The point cloud data from these sensors can be used to obtain a semantic map of the environment either directly~\cite{nuchter2008towards, Xiong2010, maturana2015voxnet} or in conjunction with color cameras~\cite{Riemenschneider2014,Wolf2015}. In this section, we discuss various semantic mapping techniques based on 3D information obtained either directly from range sensors~\cite{Xiong2010,Wolf2015} or indirectly inferred using SfM (Structure from Motion) offline~\cite{Riemenschneider2014} or other commercial solutions~\cite{Armeni2016}, like Matterport~\cite{bell2014capturing}.

\cite{Xiong2010} proposed a method for creating a 3D semantic model from cluttered point cloud data using a CRF model to discover and exploit contextual information for classifying planar patches, without relying on pre-defined rules~\cite{nuchter2008towards}. The results suggested that using co-planar context improved semantic classification results, while other tested context types didn't help.

Researchers have also focused on efficiently processing point cloud data by either avoiding redundant processing~\cite{Riemenschneider2014} or using approximation techniques~\cite{Wolf2015}. \cite{Riemenschneider2014} proposed exploiting the geometry of a 3D mesh model obtained from multi-view SfM reconstruction to avoid the redundant labeling of visually-overlapping individual 2D images. Instead of clustering similar views, their method searched for the ideal view that best supported the correct semantic labeling of each face of the underlying 3D mesh. Their proposed single-image approach performed better than label fusing of multiple images, while also being more efficient. \cite{Wolf2015} presented an efficient semantic segmentation framework for indoor RGB-D point clouds combining a Random Forest classifier and dense CRF to learn common spatial relations via pairwise potentials. The use of parallelization and mean-field approximation for CRF inference enabled a halving of computation time.

Many conventional semantic segmentation based techniques consider small-scale point cloud data. Focusing on larger scalability, \cite{Armeni2016} presented a detection-based semantic parsing method for large-scale indoor point clouds. The proposed pipeline was based on a hierarchical approach that first created semantically meaningful spaces (e.g. rooms) and then parsed them into their structural and building elements (e.g. walls and columns) -- all of it in a global 3D space, where the first step injected strong 3D priors into the second. Demonstrated at scale in an area covering over 6000 square meters and 215 million points, the study highlighted a unique set of challenges and opportunities associated with parsing such large point clouds, including the richness of recurrent geometric information and the introduction of additional semantic classes.

Recent trends include the advent of modern deep learning-based approaches, the increased availability of 3D point clouds~\cite{Silberman2012, Armeni2016} for robots, and a growing number of shape datasets~\cite{de2013unsupervised, wu20153d}. Connected with these trends has been significant growth in research focusing on 3D object detection and semantic segmentation along with shape and scene completion, areas highly relevant to dense semantic mapping. 3D ShapeNets~\cite{wu20153d} and VoxNet~\cite{maturana2015voxnet} pioneered the use of 3D CNNs for object recognition using 3D point clouds. Previously existing ``2.5D'' approaches based on RGB-D data mainly considered depth as an additional 2D channel to the RGB input and have been extensively leveraged for tasks like object recognition~\cite{socher2012convolutional, quadros2012occlusion, de2013unsupervised, song2014sliding, gupta2014learning, alexandre20163d}, grasp detection~\cite{lenz2015deep}, vehicle detection~\cite{li2016vehicle} and semantic segmentation~\cite{hoft2014fast, gupta2014learning}. The use of voxel grids~\cite{wu20153d, maturana2015voxnet} to learn 3D representations for such tasks is both conceptually different and significantly robust, as discussed in subsequent sections.

\subsubsection{\textit{Dense} Volumetric Representation and 3D CNNs}
Wu \textit{et al}.~\cite{wu20153d} proposed representing a 3D geometric shape on a 3D voxel grid and learnt a joint probabilistic distribution of binary variables on the grid using a Deep Belief Network (DBN)~\cite{hinton2006fast}. Distinct from their generative model aimed at shape reconstruction, Maturana and Scherer~\cite{maturana2015voxnet} presented VoxNet: a basic 3D CNN architecture designed for voxel grid representation of point clouds aimed at real-time 3D object detection, with an order of magnitude fewer trainable parameters than 3D ShapeNets~\cite{wu20153d}. \cite{cciccek20163d} extended U-Net~\cite{ronneberger2015u} to 3D U-Net to learn from sparsely annotated volumetric data, applied to biomedical data segmentation. Unlike the use of pre-segmented objects in~\cite{wu20153d, maturana2015voxnet,cciccek20163d}, Huang and You~\cite{huang2016point} extended the use of 3D CNNs using a LeNet~\cite{lecun1998gradient} based architecture for semantic segmentation of raw point cloud data in outdoor scenes comprising a variety of semantic classes. 3D voxel representations used with 3D CNNs often lead to coarse semantic segmentation~\cite{huang2016point, qi2016volumetric}, by assuming all the points within a voxel belong to the same object class. Addressing this problem, \cite{tchapmi2017segcloud} proposed SEGCloud as an end-to-end framework, where coarse voxel predictions from a 3D Fully Convolutional CNN are transferred back to the raw 3D points via trilinear interpolation. Fine-grained semantics are then obtained from a global consistency constraint enforced by the Fully Connected Conditional Random Field.

\subsubsection{Scene Completion and Semantic Segmentation}
``Semantic Scene Completion''~\cite{Song2016} refers to the joint learning task of scene completion and semantic segmentation, leading to enhanced understanding of the environment, particularly relevant to robotic exploration. Song \textit{et al}.~\cite{Song2016} used single-view depth map observations to produce a complete 3D voxel representation with semantic labels. They proposed \textit{SSCNet}: an end-to-end 3D CNN that uses a dilation-based 3D context module to efficiently expand the receptive field and enable 3D context learning. Dai \textit{et al}.~\cite{Dai2017a} presented a 3D CNN for shape completion that encodes the global context using semantic class predictions from a 3D shape classifier. They used a patch-based 3D shape synthesis method to refine the predictions, by imposing 3D geometry from shapes retrieved from a prior database. In follow-up work, Dai \textit{et al}.~\cite{Dai2018} designed a 3D CNN to process an incomplete 3D scan and predict a complete 3D model with per-voxel semantic labels. The proposed CNN is a fully-convolutional generative network, with filter kernels that are invariant to the overall scene size, enabling processing of large scenes with varying spatial extent. They also developed a coarse-to-fine inference strategy to produce high resolution output.

\subsubsection{\textit{Sparse} Volumetric Representation and 3D CNNs}
One of the common challenges in dealing with 3D point cloud data is its non-uniform density, where certain regions of the scene may not have any information at all, or where existing information may not always be semantically informative. A uniform voxel grid therefore may not be the best way to represent the data. In this vein, \cite{Riegler2017} proposed OctNet, a 3D CNN suited to sparse 3D data, enabling efficient representation of high resolution input. Their method hierarchically partitioned the 3D voxel grid using a set of unbalanced octrees~\cite{meagher1982geometric} where each leaf node stored a pooled feature representation, thus enabling effective allocation of memory and computation to the relevant dense regions. Sparse 3D convolutions have also been explored to avoid redundant computations~\cite{graham2015sparse, engelcke2017vote3deep, graham20183d}. \cite{graham2015sparse} extended sparse 2D CNNs used for hand-writing recognition~\cite{graham2014spatially} to sparse 3D CNNs, based on the concept of performing convolutions only on the ``active'' spatial locations that do not differ from their ``ground state''.

\cite{engelcke2017vote3deep} presented Vote3Deep: a feature-centric voting mechanism~\cite{wang2015voting} to perform sparse 3D convolutions by only considering non-zero feature locations. The use of ReLU~\cite{glorot2011deep} after a sparse convolutional layer also prevented the dilation of non-empty cells, thus achieving greater sparsity than~\cite{graham2015sparse}. \cite{graham20183d} proposed Submanifold Sparse Convolutional Networks (SSCN) that fix the spatial locations of active sites to prevent the ``submanifold dilation problem''~\cite{graham2015sparse}, keeping the sparsity unchanged for subsequent layers of the network. \cite{jampani2016learning} proposed learning high dimensional linear filters for sparse feature spaces using permutohedral lattices, and demonstrated learning the kernel parameters for a general bilateral convolution in semantic segmentation. \cite{li2016fpnn} presented the Field Probing Neural Network (FPNN) that employs field probing filters to efficiently extract features from volumetric distance fields, learning the weights and locations of the probing points that compose the filter. In order to obtain a semantically informative map, \cite{Morreale2019} proposed using semantic segmentation to remove redundancy and noise from dense 3D point clouds attributed to planar surfaces like ground and walls. They developed a variety of point cloud simplification techniques that employed global and local region statistics to selectively decimate 3D points, while mostly preserving those near intra-class edges and discontinuities. 

\subsubsection{2D Multi-View Representation of 3D}
The spatial sparsity within 3D point clouds can lead to decreased spatial resolution and increased memory consumption when using regular voxel grids~\cite{qi2016volumetric}. To mitigate these issues, an alternative approach processes multiple 2D views generated from the 3D point cloud~\cite{su2015multi} and then projects them from 2D back to 3D~\cite{lawin2017deep, Boulch2018}. \cite{su2015multi} presented a Multi-View CNN (MVCNN) to learn shape recognition from 2D renderings of a 3D point cloud, demonstrating that even a single 2D view based prediction could outperform 3D CNNs based on volumetric representation~\cite{wu20153d}. \cite{lawin2017deep} projected the point cloud onto a set of synthetic 2D images which are fed to a 2D CNN for semantic segmentation, and then re-projected this to the point cloud. Benefiting from the abundance of 2D labeled data and a multi-stream fusion of color, depth and surface normals, their method achieved superior performance to comparable baselines. \cite{Boulch2018} proposed to sample multiple 2D image views of a point cloud using random and multi-scale sampling strategies. They employed both RGB and depth-based views for 2D pixel-wise semantic labeling, with labels efficiently back-projected via buffering. \cite{tatarchenko2018tangent} explored ``tangent convolutions'' with an emphasis on leveraging 2D surface information from 3D point clouds, by projecting local surface geometry on a tangent plane around every point to obtain a set of tangent images. 

Although the methods based on multiple 2D views of 3D data are in general more efficient and scalable than voxel grid-based methods, they may not always be applicable for challenging 3D shape recognition tasks due to loss of information during projection operations~\cite{klokov2017escape}. Alternatives in the form of hybrid representations~\cite{guo2020deep} of 3D point cloud have also been considered. Dai \textit{et al}.~\cite{Dai2018a} proposed an end-to-end trainable 3D CNN to predict per-voxel semantic labels. Their method combines two streams of features: one extracted from per-voxel max pooling of multiple 2D RGB views and the other from 3D geometry, demonstrating a significant performance improvement attributed to this joint 2D-3D learning.

\subsubsection{Unstructured Points Based Representations}
While volumetric representations present the environment in a structured manner, handling point cloud data as a set of unordered and unstructured points has proven to also be a viable approach. \cite{Qi2017a} presented \textit{PointNet}, a novel deep network that processes points sets without voxelization or rendering and learns both global and local features. Demonstrated to be efficient on tasks like object classification, part segmentation, and scene semantic parsing (dense semantic mapping), PointNet works on an unordered set of points exhibiting permutation-invariance, unlike ordered 2D image data or a 3D volumetric grid, while also being robust to input perturbation and corruption. \cite{Qi2017c} proposed \textit{PointNet++}, a hierarchical NN that applies PointNet recursively on a nested partitioning of the input point set, enabling learning of local features with increasing contextual scales. With the ability to capture local structures induced by the metric space of points, PointNet++ demonstrated improvement on semantic segmentation of single-view 3D point clouds. \cite{Engelmann2018} explored direct semantic segmentation of unordered point clouds by extending PointNet~\cite{Qi2017a}, particularly enlarging its receptive field over a large 3D scene to incorporate larger-scale spatial context at both input and output level. Building on point-level representations, significant research has focused on developing learning algorithms based on characteristics like unordered structure, intra-point interaction and transformation-invariance~\cite{liu2019deep}, for example, by using Point-wise convolutions~\cite{hua2018pointwise, wu2019pointconv}, Recurrent Neural Networks~\cite{ye20183d, huang2018recurrent}, Graph Neural Networks~\cite{te2018rgcnn, wang2019dynamic} and Autoencoders~\cite{yang2018foldingnet,zamorski2018adversarial}. 

\subsubsection{Graph- and Tree-Based Representations at Scene Level}
In order to better capture the structure of 3D point clouds without requiring voxel grids, researchers have explored the use of graph-~\cite{simonovsky2017dynamic, Qi2017} and tree-~\cite{klokov2017escape, zeng20183dcontextnet} based networks. \cite{simonovsky2017dynamic} proposed an Edge-Conditioned Convolution (ECC) to learn from the local neighborhood of the point cloud represented as a graph, where filter weights were dynamically generated for each specific input sample. \cite{Qi2017} designed a 3D Graph neural network that builds a k-nearest neighbor graph on top of 3D point clouds, where each node in the graph corresponds to a set of points represented by appearance features extracted by a unary CNN from 2D images. This approach consequently was able to leverage both 2D appearance and 3D geometric relationships. The method is capable of extracting long range dependencies within images -- typically difficult to model in traditional techniques. Inspired from superpixels~\cite{achanta2012slic}, \cite{landrieu2018large} introduced SuperPoint Graphs (SPG) where nodes represent simple shapes, and edges encode contextual relationships between object parts, enabling semantic segmentation of large-scale point clouds.

\cite{klokov2017escape} proposed Kd-Net for parsing 3D point clouds, using a kd-tree structure~\cite{bentley1975multidimensional} to form the computational graph, and hierarchical representations where the root node is recursively computed from representations of its children. Addressing the lack of overlapping receptive fields in Kd-Net, \cite{li2018so} presented SO-Net based on the Self-Organizing Map (SOM)~\cite{kohonen1990self}, where the receptive field overlap is controlled by performing point-to-node k-nearest neighbor (kNN) search on the SOM, enabling a better representation of the underlying spatial distribution of point clouds.

The increasing ubiquity of 3D sensing~\cite{bell2014capturing, Wolf2015, Boulch2018}, maturity of visual SLAM~\cite{mur2017orb, rosinol2020kimera} and SfM techniques~\cite{schonberger2016structure, kong2019deep, vijayanarasimhan2017sfm}, and the evolution of deep learning have together facilitated extensive research on 3D point clouds. This has led to numerous novel approaches, for example, ShapeNet~\cite{wu20153d}, VoxNet~\cite{maturana2015voxnet}, OctNet~\cite{Riegler2017}, PointNet~\cite{Qi2017a}, tangent convolutions~\cite{tatarchenko2018tangent}, SSCN~\cite{graham20183d} and Kd-Net~\cite{klokov2017escape}, which have pushed the boundaries of 3D scene understanding. Although  capable of processing information at an environment level -- beyond what can be achieved through a single image, these methods do not typically consider \textit{incremental} scene understanding. In the near future, bridging the advances in 3D point cloud processing and 3D reconstruction capabilities of modern SLAM techniques will significantly benefit robots in semantic scene understanding.

\section{INTERACTING WITH HUMANS AND THE WORLD}
\label{secn:interaction}
The absence of semantics in much of the pivotal early robotic mapping and navigation research was due in part to the fact that significant progress in mapping and navigation could be made without the robot needing a richer understanding of the world around it. Most of the mapping systems in current commercial deployments like mines and ports have little higher-level understanding of the environment they are operating in. This same story also played out to some extent in robotics research focused on enriching the interactions of robots with the world around them, and with the humans who occupy that world. Early work in areas like manipulation and human-robot interaction necessarily focused more on the technical mechanics of the interaction, not least in part due to the computational and hardware limitations of the time.

It is now becoming increasingly evident that further progress will rely on robots moving beyond a primitive understanding of the world around them. Robots will need to understand the ``things'' in the environment surrounding them and what the ``affordances'' of those things are, in order to understand what the robot can achieve with those things. In mixing freely with humans in the environment, robots will also need to understand humans; perhaps not initially at a deep cognitive level, but at least in terms of recognizing what humans are doing - their actions - in order to inform the robot's potential interactions with humans. The following sections survey the state of robotics research into semantics for enhancing the capability of robots to understand and interact with the world around them. We first discuss the literature in the context of ``perception \textit{of} interaction'' which covers the works related to understanding of different actions and activities performed by humans, and the use of hands and arms to interact with objects. Then, we discuss ``perception \textit{for} interaction'' covering the \textit{use} of perception to interact with the objects or humans around a robot, including research related to interaction with objects (affordances, manipulation and grasping), performing higher-level tasks, interacting with humans and other robots and finally, navigation based on vision and language.

\subsection{Perception \textit{of} Interaction}
\subsubsection{Actions and Activities}
When interacting with the world, especially with the humans that occupy it, an understanding of actions and activities is essential. In its most basic form, activity recognition can be simply treated as a supervised machine learning problem---present the machine with a short video clip and classify the entire clip into one (or more) of $k$ different categories learned from a large corpus of labeled examples~\cite{Simonyan:NIPS2014, Laptev:CVPR2008, Rodriguez:CVPR2008,Soomro:arxiv2012,Carreira:CVPR2017}. Moving beyond just classification, localization of specific activities within a long video sequence involves identifying both the temporal and spatial location of each activity, often in so-called ``action tubes''~\cite{Gkioxari:CVPR2015}. This approach to activity recognition and detection is typified by the various tasks in the annual ActivityNet Challenge and its associated dataset of human activities~\cite{Caba:CVPR2015}.

Many of the works in this area consider activity recognition from the perspective of a third-person observer (a notable exception being the EPIC-Kitchens egocentric activity classification dataset~\cite{Damen:ECCV2018}) and learn black-box models that segment and label activities. In some cases the temporal aspect of an action is modeled through dense trajectory features~\cite{Wang:CVPR2011}, which track the movement of keypoints through time. A sequence of atomic action units or ``Actoms''~\cite{gaidon2011actom} can also be used to represent a semantically-consistent action sequence as opposed to learning its discriminative parts. \cite{gaidon2014activity} show that a complex activity composed of several short actions is better represented and compared using a hierarchy of motion decomposition. The two-streams approach~\cite{Simonyan:NIPS2014} separates appearance and motion (optical flow). A rank-pooling method developed by Fernando and colleagues~\cite{Fernando:PAMI2016, Fernando:ICML2016, Fernando:CVPR2016} encodes frame order as a function of the frame's appearance. An efficient first-order approximation of rank-pooling, known as Dynamic Images~\cite{Bilen:CVPR2016}, has been applied successfully in many deep learning approaches for activity classification. However, even Carreira and Zisserman~\cite{Carreira:CVPR2017}, who ask the question ``\emph{Quo Vadis}, Action Recognition?''\footnote{Where are you going action recognition?}, do not fully consider a more fine-grained understanding of human actions (e.g. different types of swimming) likely necessary for robotic applications. Robots are likely to require not just classification of the activity being performed, but also localization of both the person performing the activity and the object being interacted with (if any). 

One of the most important pieces of semantic information for understanding activities is the pose of the human performing the activity and how that pose changes over time. Ramanan and Forsyth~\cite{Ramanan:NIPS2003}'s work is an early example of research proposing to use human pose and motion to classify activities. In their model body parts are tracked over time, with sequences annotated using a hidden Markov model. In more recent work, Wang \textit{et al}.~\cite{Wang:action:CVPR2013} improve the estimation of human pose by first building dictionaries of spatial and temporal part sets and then using a kernel SVM to classify actions. Luvizon \textit{et al}.~\cite{Luvizon:CVPR2018} solve human pose estimation and activity recognition jointly and train a single deep learning model to perform both tasks, with pose and appearance features combining to predict actions.

In the context of robotic vision, Ramirez-Amaro \textit{et al}.~\cite{Ramirez-Amaro2019} present a recent survey of the most representative approaches using semantic descriptions for recognition of human activities, with the intention of subsequent execution by robots. In such a context it is advantageous to characterize human movement through multiple levels of abstraction (in this particular case four) from high-level processes and tasks to low-level activities and primitive actions. Given this hierarchy and a ``semantic'' representation of a human demonstrated task, the ultimate goal is for a robot to replicate the task, or assist the human in achieving the intended outcome.

Similar to the work of Ramanan and Forsyth~\cite{Ramanan:NIPS2003}, Park and Aggarwal~\cite{Park2004} propose to recognize human actions and interactions in video by modeling the evolution of human pose over time. Their framework uses a three-level abstraction where the pose of individual body parts (e.g. head and torso) are identified and linked at the lowest level using a Bayesian network; actions of a single person are modeled at the mid-level using a dynamic Bayesian network; and alignment of multiple dynamic Bayesian networks in time allows inference of interactions at the highest level. The result is a representation that can be translated into a meaningful semantic description in terms of subject, verb and object.

Reducing the reliance on large quantities of annotated training data, Cheng \textit{et al}.~\cite{Cheng2013a} developed a zero-shot learning framework that also considers a three-level hierarchy that takes into account low level features, mid-level semantic attributes and high-level activities mapped through a temporal sequence. The result is a model that can recognize previously unseen human activities. Cheng and his colleagues extended this work by proposing a two-layer active learning algorithm for activity recognition for unseen activities, using a semantic attributes-based representation. In this case an attribute is a human-readable term describing an inherent characteristic of an activity~\cite{Cheng2013}.

\subsubsection{Hands and Arms}

Within the broader topic of general activity recognition, hands and arms play a key role in activities where humans are involved. \cite{Ramirez-Amaro2014} proposed forming a semantic representation of human activities by extracting semantic rules based on three hand motions: move, not move, and tool use, and two object properties: ObjectActedOn and ObjectInHand. Their system is adaptable to new activities which can be learned on-demand. \cite{Ramirez-Amaro2015b} extended the work in~\cite{Ramirez-Amaro2014} and explored the use of virtual reality-based additional 3D information to understand and execute the demonstrated activities. They integrated their method on an iCub humanoid robot. \cite{Ramirez-Amaro2015a} also developed a method to infer human-coordinated activities (like use of two hands), that was invariant to observation from different execution styles of the same activity (e.g. left-handed vs right-handed). They used a three-level approach: extract \textit{relevant} information from observations, infer the observed activity, and trigger the motion primitives to execute the task.

\cite{Ramirez-Amaro2015} explored an unsupervised learning method based on Independent Subspace Analysis (ISA) to extract invariant spatio-temporal features directly from unlabeled video data. A second stage automatically generated semantic rules, enabling high-level reasoning about human activities that resulted in improved performance. \cite{Ramirez-Amaro2017} proposed semantic representations of human activity, where semantic rules based on relationships between human motions and object properties are first learned for basic and complex activities, and then used to infer the activity. They used Web Ontology Language (OWL) and KnowRob for incorporating knowledge/ontology. \cite{Ramirez-Amaro2017a} extended the semantic reasoning method from~\cite{Ramirez-Amaro2014} by including gaze data (in addition to the third person view) to segment and infer human behaviors. They demonstrated the complementary nature of first- and third-person view (using egocentric and external cameras respectively), enabling the system to deal with occlusions. Finally, \cite{Ramirez-Amaro2019a} presented a novel learning-by-demonstration method that enables non-expert operators to program new tasks on industrial robots. The proposed semantic representations were found to be invariant to different demonstration styles of the same activity.

\cite{Yoon2018} presented a new representation for 3D reconstructed trajectories of human interactions, referred to as ``3D Semantic Map'' - a probability distribution over semantic labels observed from multiple views. They use semantics of body parts, for example the head, torso, legs and arms, and other objects in the vicinity, such as a dog or a ball. Spatio-temporal semantic labels are inferred by graph-cut formulation based on multiple 2D labels. The authors claim ``This paper takes the first bold step towards establishing a computational basis for understanding 3D semantics (of human interaction) at fine scale.'' Chang \textit{et al}.~\cite{Chang2016,Chang2018} took a complementary approach by finding structural correspondences of objects that have similar topologies or motions. For example, their method is able to match corresponding body parts (including arms and hands) of a humanoid robot with that of a human from a video sequence without any prior knowledge of the body structures. \cite{Wang2018} discussed the availability of large amounts of visual data from wearable devices and existing methods used to characterize everyday activities for visual life-logging, for example, semantic annotations of visual concepts and recognition and visualization of activities. Motivations in this area of research include its use for behavioral analysis and in assistive living scenarios. 

\subsection{Perception \textit{for} Interaction}

\subsubsection{Object Affordances}

The term ``affordances'' was coined by the psychologist Gibson~\cite{gibson2014ecological} to explain how inherent values of things in the environment can be perceived and how this information can be linked to the potential actions of an agent. In one of the early applications of the concept in robotics, \cite{fitzpatrick2003learning} presents a robot that learns the roll-ability of four different objects by ``playing'' with them and observing the changes in the environment; however the generalization of this affordance to novel objects and affordances is not studied.
In order to formalize the concept of affordances in the context of autonomous robotics, \cite{csahin2007afford} reviews the usage of affordances in other fields, and proposes a framework based on relation instances of the form \texttt{(effect, (entity, behavior))} which are acquired through the interaction of the agent with its environment.  \cite{dogar2007primitive} implements this framework for a mobile robot, enabling goal-oriented navigation by learning the affordances of its environment through the effects of primitive behaviours such as ``traverse'' or ``approach''.

Performing complex manipulation actions requires the encoding of temporal dependencies in addition to an understanding of affordances.
Closely related to the formalization of affordances above, the Object-Action Complexes (OACs) concept \cite{kruger2011object} attaches the performed actions to the objects as attributes and provides grounded abstractions for sensory–motor processes.  By defining affordances as state transition functions, OACs allow for forward prediction and planning. Several researchers have studied deriving such attributes from visual information \cite{Aksoy2011,Aein2013,Aksoy2015a,yang2015robot}. The Semantic Event Chain (SEC) framework developed by \cite{Aksoy2011} analyzes the sequence of changes of the spatial relations between the objects that are being manipulated by a human or a robot. This approach generates a transition matrix which encodes topological variations in a sequence graph of image segments during the manipulation event, allowing transfer of manipulation actions from human demonstration to a robotic arm. \cite{Aein2013} designed an architecture for generic definition of a robot's human-like manipulation actions. Manipulations are defined using 3 levels: a top level, which abstractly defines objects and their relations and actions; a mid level, which defines chaining of action primitives via SEC; and a bottom level that defines sensory data collection and communication with the robot control system. 

Using these representations, a number of approaches have been presented for autonomously learning affordances in the context of manipulation.
\cite{Aksoy2015a} proposes online incremental learning of an archetypal Semantic Event Chain model for each manipulation action by observation, without requiring any prior knowledge about actions or objects. Similarly, \cite{yang2015robot} leverages information from the cloud, by learning manipulation action plans from cooking, fetched from the web. Lower-level perception modules, grasping type recognition and object recognition is combined with a probabilistic manipulation action model at the higher level. \cite{kappler2012templates} studies pre-grasp sliding actions to facilitate grasping, for instance sliding a book to the table edge so that it could be grasped. In this work, objects are divided into categories, such as box or cylinder, based on sub-symbolic object features including shape and weight. A sliding action is simulated before actual execution, conditioned on the object class. Afrob~\cite{varadarajan2012afrob} provides a database for affordances, targeted at domestic robot scenarios, consisting of 200 object categories and structural, material and grasp affordances.

The following subsections discuss more concrete applications of affordance and semantics to robotic tasks, ranging from low-level, e.g. item ``graspability'', to high level concepts such as understanding language and social behaviour.

\subsubsection{Grasping and Manipulation}

Robotic grasping is concerned with the problem of planning stable grasps for an object using a robotic gripper. Early approaches to this problem were largely analytical, focusing on providing a theoretical framework based on mathematical and physical models of contact points between the manipulator and object, used to quantify the stability and robustness of grasps~\cite{roa2015grasp, bicchi2000robotic}. However, the reliance of such methods on precise physical and geometrical models makes them brittle in the presence of noise or uncertainty found in robotic systems. As such, many modern approaches to robotic grasping are data-driven, using generalizable models and experience-based approaches to improve calculation times, robustness to errors and enable generalisation to novel objects \cite{bohg2013data}.

A common approach in data-driven grasp synthesis is to use semantic information to transfer grasping knowledge to ``known'' or ``familiar'' objects. A number of works rely on precise object recognition and pose estimation to apply pre-computed grasps from a known object model to an instance of that model in reality \cite{detry2011learning, detry2010refining, zeng2017ew, morales2006integrated, tremblay2018dope, papazov2012rigid, collet2011moped}. More general methods work with familiar objects; those which are semantically or geometrically similar to objects previously seen, and are able to generalise grasps across instances of an object's class (for example mugs) \cite{Dang2012, hillenbrand2012transferring} or object parts \cite{detry2012generalizing, detry2013learning} (for example handles)~\cite{ten2016localizing}. Song \textit{et al.} \cite{song2015learning} present a hybrid grasp detection pipeline for familiar objects that fuses local grasp estimates based on depth data with global information from a continuous category-level pose estimation process, which is able to overcome the limitations of both approaches individually.
Rather than rely on object models, Manuelli \textit{et al.}~\cite{manuelli2019kpam} define grasping and manipulation actions by learning semantic 3D keypoints from visual inputs that generalise across novel instances of an object class, making them more robust to intra-category shape variation.

More recently, data-driven approaches backed by deep learning have proven very effective at predicting robotic grasps on arbitrary objects from visual information~\cite{morrison2018closing, morrison2019learning, mahler2017dex, mahler2019learning, ten2017grasp}, seemingly removing the immediate need for semantic information in the grasping pipeline, at least in terms of grasping alone. However, in many robotics applications, the ability to grasp and transport arbitrary objects in a stable manner is not enough in itself, but rather a prerequisite for higher-level manipulation tasks where the choice of grasp is dependent on the goal~\cite{billard2019trends}. 
The representation of semantics in robotic grasping may take many different forms and is largely informed by the task to be completed. 

In the simplest case, the semantics of a manipulation task may be defined at an object recognition level, where a grasp detection algorithm is combined with an object detection or semantic segmentation system (Section~\ref{secn:semanticsfundamental}).
This is a common choice in pick-and-place style tasks, such as warehouse fulfilment, where the requirement for semantics is limited to object identification. A common successful approach in the Amazon Robotics Challenge was to use an object-agnostic grasp detection algorithm in parallel with a standalone semantic segmentation system~\cite{morrison2018cartman,zeng2018robotic,schwarz2018fast}. 
Rather than use two separate streams, other approaches use joint representations of object-specific grasp detection. Guo \textit{et al.}~\cite{guo2016object} train a CNN to recognise the most exposed object in a cluttered scene while simultaneously regressing the best grasp pose for the detected object. This joint model outperformed a system comprising two models trained individually on object and grasp detection. \cite{Jang2017} improve this idea by using a two-stream attention-based network that learns object detection, classification, and grasp planning end-to-end. The two streams, ventral and dorsal, perform object recognition and geometric-relationship interpretation respectively, and outperform a single-stream model on the same task.

To grasp an object, humans rely not only on spatial information, but also knowledge about the objects, such as its expected mass and rigidity, and the constraints of the task to be performed~\cite{verhagen2012cortical}, for example, grasping a knife by the handle correctly to perform a cut, or not blocking the opening of a mug to pour water. In the same way, to perform higher level tasks robots must also understand the possible function (``affordances'') of objects and how they relate to the constraints of tasks to be performed. To this end, a number of research works focus on encoding semantic constraints within a robotic grasping pipeline. 

Affordance prediction can also be performed as a standalone task as a precursor to a separate grasp planning pipeline, relying largely on visual cues. Myers \textit{et al.}~\cite{myers2015affordance} present an approach inferring the affordances of objects based on geometric information and visual cues. By combining hand-coded features with traditional machine learning methods they were able to predict object part affordances such as \textit{cut}, \textit{contain}, \textit{pound} and \textit{grasp} in RGB-D images of common kitchen and workshop tools. Nguyen \textit{et al.} \cite{nguyen2016detecting} extended this approach by using a CNN to predict pixel-wise affordances, removing the need for hand-designed features. Kokic \textit{et al.}~\cite{kokic2017affordance} use a CNN to detect affordances on arbitrary 3D objects based on geometric rules, for example, that thin, flat sections of objects afford (provide) \textit{support}. The affordance detection is combined with a separate object classification and orientation stream to plan task-specific grasps.

Song \textit{et al.}~\cite{song2011multivariate, song2015task} present a probabilistic framework for encoding the relationships between features for task-specific grasping using Bayesian networks. Their model jointly encodes features of the target object, such as category and shape, task constraints and grasp parameters, hence modelling the dependencies between relevant features. However, the approach also relies on hand-crafted features for representing objects and task constraints, making it difficult to scale to extra tasks and objects. Hjelm \textit{et al.}~\cite{hjelm2015learning} use metric learning to automatically learn the relevant visual features for a given task from human demonstration. Along the same lines, Dang and Allen~\cite{Dang2012} present ``semantic affordance maps'' that link local object features (a depth image) along with tactile and kinematic data to semantic task constraints for an object. Given an object of a known class and a task, the optimal grasp approach vector is computed from the semantic affordance map, followed by a grasp planning step along the computed approach vector. 

The above methods all employ some level of category-specific encoding, meaning that they won't generalise to new object classes. In contrast, Ard\'on \textit{et al.}~\cite{Ard2018} proposed employing semantic information from the surrounding environment of a target object to improve inference of affordances. They used a weighted graph-based knowledge base to model attributes based on shape, texture and environment, such as locations or scenarios in which the object is likely to appear. Their method does not require a-priori knowledge of the shape or the grasping points. Detry \textit{et al.}~\cite{Detry2017} present another approach to grasping previously-unseen objects in a way that is compatible with a given task. They use CNN-based semantics to classify image regions that are compatible and incompatible with a task in combination with a geometric grasp model. For training on tasks like ``pouring'', task constraints like ``grasp away from opening'' are used to define the \textit{suitable} and \textit{unsuitable} labels for vertices of object mesh models. The ``semantic success'' of a task is assessed based on whether the grasp is compatible with the task and ``mechanical success'', by having the robot attempt to lift the object off the table. 

Combining semantic information with robotic grasping is likely a crucial step towards equipping robots with the ability to autonomously carry out tasks in unstructured and dynamic environments. While the above approaches all achieve impressive results, the largest outstanding questions still revolve around accurately and efficiently transferring semantic knowledge to new objects and tasks.

The following section looks at approaches that go beyond grasping and make use of semantic information for higher-level task completion. 

\subsubsection{Higher-level Goals and Decision Making}

Semantic maps and information can be exploited to facilitate higher-level goal definitions and planning, for humanoids \cite{Kaiser2015,Zeng2018} and domestic robots \cite{blodow2011autonomous, cosgun2018context, Zeng2018, galindo2013inferring}. These approaches often use RGB-D data to extract semantic information. For humanoids, \cite{Kaiser2015} used RGB-D data to build representations of the environment from geometric primitives like planes, cylinders and spheres. This information is used to extract the affordances of objects, such as their ``pushability'' and ``liftability'', as well as full-body capabilities such as support, lean, grasp and hold. \cite{Grotz2017} extends the work in \cite{Kaiser2015} with semantic object information, such as chair or window, and uses spatio-temporal fusion of multiple geometric primitives (using a spatial scene graph representation) to identify higher level semantic structures in the scene.

For object manipulation, \cite{Zeng2018} parses a given goal scene as an axiomatic scene graph composed of object poses and inter-object relations. A novel method is proposed for scene perception to infer the initial and goal states of the world. \cite{blodow2011autonomous} utilizes various segmentation methods for generating initial hypotheses for furniture drawers and doors, which are later validated with physical interaction. Given a task, this approach is able to answer queries such 
as ``Which horizontal surfaces are likely to be counters?'' by processing knowledge built during mapping. \cite{cosgun2018context} present an interactive labeling system, where human users can manually label semantic features, such as walls and tables, and static objects, such as door signs. These labels can then be used to define navigation goals such as ``go to the kitchen'', or for specialized behaviour such as for passing doors. \cite{galindo2013inferring} presents a goal-inferring approach that detects deviations from operational norms using semantic knowledge. For instance, if the robot ``knows'' that perishable items must be kept in a refrigerator, and observes a bottle of milk on a table, the robot will generate the goal to bring that bottle into a refrigerator.

Robotic tool use for manipulation brings the topics of object affordances, grasping and task-level decision making together. \cite{mar2017can} studies tool affordances through self-exploration based on their geometry - how the tool is grasped. In particular, they study the dragging action, parameterized by the angle of the drag, with the effect measured as the object displacement. In more recent work, Fang \textit{et al.}~\cite{fang2020learning} address task-oriented grasping from the point of view of tool usage, where a robot must grasp an object in a way which allows it to complete a task. For example, the best way to grasp a pair of scissors differs depending on the task: transporting the scissors may require a different, safer grasp to using them to cut a piece of paper. Their approach uses self-supervised learning in simulation to jointly learn grasping and manipulation policies that are then transferred to novel tools in the real world.

\subsubsection{Semantics for Human-Robot and Robot-Robot Interaction}
Semantics play a crucial rule in human-robot interaction, in particular for collaborative task execution which goes beyond the human-command-robot-execute model. This is achieved using inverse semantics, introduced by Tellex \textit{et al.}~\cite{Tellex2015}, to enable the robot to generate help requests (for an assembly task) for robots in natural language, by emulating the ability of humans to interpret requests. Inverse semantics involved using a Generalized Grounding Graph ($G^3$) framework~\cite{tellex2011understanding} for the task and using models based on both the environment and the listener. Gong and Zhang~\cite{Gong2018} proposed temporal spatial inverse semantics to enable a robot to communicate with humans by extending the natural language sentence structure that can refer to previous states of the environment, for example, ``Please pick up the cup beside the oven that was on the dining table''. They used the G3 framework for mapping between natural language and groundings in an environment, and evaluated their method through randomly generated scenarios in simulation.

Semantics have also been used to program a robot to perform a task using natural language sentences. Pomarlan \textit{et al.}~\cite{Pomarlan2018} developed a system that converts a linguistic semantic specification (like an instruction to a robot: ``set a table (re-arranging four cups)'') into an executable robot program, using interpretation rules. The resultant robot program or plan can be run in simulation and enables further inference about the plan. Savage \textit{et al.}~\cite{Savage2019} developed a semantic reasoning module for a service robot that interprets natural language by extracting semantic role structures from the input sentence and matching them with known interpretation patterns.

Another venue for semantics in Human-Robot interaction is interactive robot learning from demonstration. Niekum \textit{et al.}~\cite{niekum2013incremental} proposed an approach to using demonstrations for learning different
behaviors and discovering semantically grounded primitives and incrementally building a finite-state representation of
a task. Recently, semantics, powered by deep learning, was used to enable a robot to understand salient events in a human-provided demonstration~\cite{sermanet2016unsupervised} or perform unsupervised learning of the robot's user type from joint-action demonstrations, in order to predict human actions and execute anticipatory actions~\cite{8520627}.

Mapping with Human-in-the-loop is another example where semantics and semantic mapping find application. Hemachandra~\textit{et al.}~\cite{Hemachandra2014} presented a semantic mapping algorithm, Semantic Graph (a factor graph approach), that fuses information from natural language descriptions with low-level metric and appearance data, forming a hybrid map (metric, topological, and semantic). \cite{Yang2018} proposed a mapless navigation system based on semantic priors, using reinforcement learning on joint embeddings created from image features, word embeddings, and a spatial and knowledge graph (using a graph convolutional network) to predict the actions. 

In the context of autonomous navigation and motion planning, one of the desired capabilities of a robot is to actively understand its environment and act accordingly. Beyond the use of active SLAM techniques~\cite{stachniss2004exploration, carlone2010application, carlone2014active} to accurately explore a partially-known environment, a robot also needs to adhere to human-like ``social behavior'' for an enhanced human-robot interaction experience. Semantic reasoning potentially enables robot social awareness when planning motion around humans~\cite{lemaignan2017artificial,sisbot2007human}, expressing socially competent navigation in shared public places. This is typically achieved through crowd flow modeling in different social scenarios and contexts~\cite{kidokoro2015simulation}.

Robot interaction can occur not only with humans but also with other robots. Schulz \textit{et al.}~\cite{schulz2011lingodroids} developed a system that enables a pair of robots to interact by autonomously developing a sophisticated language to negotiate spatial tasks. This is achieved through language games that enable the robots to develop a shared lexicon to refer to places, distances, and directions based on their cognitive maps, and was demonstrated on real robots. Another example of robot-robot interaction is the computational model of perspective-taking introduced by Fischer and Demiris~\cite{fischer2019computational}. There, one robot assumes the viewpoint of another robot to understand what that other robot can perceive from this point of view. Such a perspective-taking ability has previously been shown to be of advantage in human-robot interactions~\cite{Fischer2016,breazeal2006using}.

\subsubsection{Vision-and-Language Navigation} Vision-and-language navigation (VLN) is a task where an agent attempts to navigate to a goal location in a 3D simulator when given detailed natural language instructions. Anderson \textit{et al.} presented the first VLN task and dataset in \cite{r2r}. Numerous methods have been proposed to address the VLN problem. Most of them employ the CNN-LSTM architecture, with an attention mechanism to first encode instructions and then decode the embedding to a series of actions. Together with the proposing of the VLN task, Anderson \textit{et al.}~\cite{r2r} developed the teacher-forcing and student-forcing training strategy. The former equips the network with a basic ability for navigating using ground truth action at each step, while the latter mimics the test scenarios, where the agent may predict wrong actions, by sampling actions instead of using ground truth actions. 

The trend of deep learning based methods benefiting from more training data is also true in vision-and-language navigation. To generate more training data, Fried \textit{et al.} \cite{sf} developed a speaker model to synthesize new instructions for randomly selected robot trajectories. Other methods of increasing the training data available have also been proposed: Tan \textit{et al.}~\cite{envdrop} augment the training data by removing objects from the original training data to generate new \textit{unseen} environments. To further enhance the generalization ability, they also train the model using both imitation learning and reinforcement learning, so as to take advantage of both off-policy and on-policy optimization.
Predicting future actions is a key component of many approaches, but not all chunks of an instruction are useful for predicting this next action. Ma \textit{et al.} \cite{selfMonitor} use a progress monitor to locate the completed sub-instructions and to focus on those which have the most utility in predicting the next action.
Another way to improve navigation success is to equip the agent with backtracking.
In \cite{regret}, Ma \textit{et al.} propose treating the navigation process as a graph search problem and predict whether to move forward or roll back to a previous viewpoint. In \cite{fast}, each passed viewpoint is viewed as a goal candidate, and the final decision is the viewpoint with the highest local and global matching score. Most recently, Qi \textit{et al.} \cite{qi2019rerere} presented a dataset of varied and complex robot tasks, described in natural language, in terms of objects visible in a large set of real images. Given an instruction, success requires navigating through a previously-unseen environment to identify an object. This represents a practical challenge, but one that closely reflects one of the core visual problems in robotics.


\section{IMPROVING TASK CAPABILITY}
\label{secn:improvingTask}
Semantic representations extracted through static or dynamic scene understanding processes can be leveraged by robots to improve their performance at tasks, whether a basic task of localization or higher-level functions like intention prediction of other drivers on the road. Beyond the foundational semantic understanding research, task- or application-specific semantic representations are often required to achieve specific goals. In this section, we discuss various ways in which researchers extract or employ semantic representations for localization and visual place recognition, dealing with challenging environmental conditions, and additional considerations required for enabling semantics in a robotics context.

\subsection{Semantics for Localization and Place Recognition}
A major sub-component of the navigation problem is localization - assuming a map exists, how can a robot calculate its location within the map. This process occurs in a number of different ways and has differing, overlapping terminology. Localization often refers to calculating the robot's location in some form of metric map space, while place recognition often refers to recognizing a particular sensory snapshot or encoding of a distinct location, without necessarily any use of or calculation of metric pose information. In a SLAM system performing mapping, the act of recognizing a familiar location is called ``closing the loop'' or ``loop closure''. Depending on the sensory modality used, the terminology is often pre-fixed with the appropriate term - for example \textit{visual} place recognition. In this section we survey the increasing use of semantics for localization and place recognition systems, with a focus on vision-based techniques given their natural affinity for semantics-based approaches.

Visual Place Recognition (VPR) is integral to vision-based mobile robot localization. Given a reference map of the environment comprised of unique ``places'', the task is to recognize the currently observed place (query) and decide whether or not the robot has seen this before. VPR is a widely researched topic in robotics, as reviewed in a recent survey~\cite{Lowry2016}. Distinct from place \textit{classification} or \textit{categorization}, processes that produce place-level semantic labels representing the \textit{general} appearance of the scene, here we discuss place \textit{recognition} that produces a localization cue by identifying a \textit{specific} place. In particular, we review the VPR methods that use semantic information at either scene, object, patch, edge or pixel level for effective localization.

\subsubsection{Object-level Semantics for VPR}
Semantic information in the form of object detection is often used for place recognition and localization, as well as in full SLAM systems (see  Subsection~\ref{sec:semRep+SU+SLAM}). \cite{kim2009simultaneous} developed a hierarchical graphical model that enabled simultaneous object and place recognition using bidirectional interaction between objects and places. Similarly, \cite{Luo2011} presented a hierarchical random field model based on SIFT~\cite{lowe2004distinctive} and GIST~\cite{oliva2006building} using relative pose context among point features, objects and places. \cite{Atanasov2016} explored the use of a semantically-labeled prior map of landmarks based on object recognition~\cite{felzenszwalb2009object} for localizing a robot. They developed a sensor model to encode semantic observations with a unified treatment of missed detections, false alarms and data association, and demonstrated it functioning on simulated and real-world indoor and outdoor data. 
\cite{weng2018semantic} proposed ``semantic signatures'' as descriptors of images comprising \textit{type} and \textit{angle} sequences of objects visible from a given spatial location. They used trees, street lights and bus stops as objects for this task. 

With further advances in the field of object recognition, methods like Faster-RCNN~\cite{ren2015faster} have been adopted for visual localization. Recently, \cite{zhang2019coarse} developed a hierarchical localization pipeline based on a semantic database of objects detected using \cite{ren2015faster} and SURF~\cite{bay2006surf} descriptors. In particular, the coarse localization estimates based on object matching were refined using keypoint-level SURF matching to obtain the final result. \cite{han2018learning} presented a novel place representation based on a graph where nodes were formed by holistic image descriptors and semantic landmarks derived from \cite{ren2015faster}; the edges in the graph indicated whether or not the nodes represented the same landmark/place; finally, the learnt graphical embeddings were used for visual localization.

Another source of semantic information in urban environments is Geographical Information System (GIS) databases, which provide semantic and pose information for various objects like traffic signs, traffic signals, street lights, bus stops and fire hydrants. \cite{ardeshir2014gis} extracted and combined GIS data with object semantics to improve both object detection (bounding box) and geospatial localization. They tackled the inherent noise within GIS information by formulating a higher-order graph matching problem, solved robustly using RANSAC. Coarse localization is enabled by searching over a dense grid of locations for high similarity between GIS objects and objects detected within the query image. Similar to~\cite{ardeshir2014gis}, \cite{castaldo2015semantic} explored the use of Deformable Part Models (DPM)~\cite{felzenszwalb2009object} along with GIS information for localization but based on ``cross-view'' matching, where a query image from street-level is matched against an aerial top-view GIS semantic map. They proposed a Semantic Segment Layout (SSL) descriptor to validate the spatial layout of semantic segments detected in a query image in accordance with the GIS map.

In order to localize precisely, certain semantic categories in the scene can be more useful than the others depending on the underlying application scenario. \cite{pink2008visual} presented a lane markings-based localization system, where a prior map is derived from aerial images and map matching is performed using ICP. Aimed at assisted and autonomous driving, \cite{schreiber2013laneloc} explored the use of highly accurate maps comprising visible lane markings and curbs for precise and robust localization. They employed a Kalman Filter~\cite{kalman1960new} for map matching using stereo cameras and an IMU to achieve centimeter-level precision. \cite{armagan2017learning} proposed using segment building facades to estimate camera pose, given a 2.5D map of the environment comprising building outlines and their heights. \cite{mousavian2016semantic} developed an approach for geo-locating a novel view of a scene and determining camera orientation using a 2D map of buildings along with a sparse set of geo-tagged reference views. They detected and identified building facades, along with their geometric layout, in order to localize a query image.

\subsubsection{Beyond Object-Level Semantics}
Semantic segmentation of visual information can not only be leveraged at object or pixel level, but at edge level as well. \cite{meguro2007development} and \cite{bazin2009dynamic} segmented the sky using Infrared images to use it as a unique signature for localization. The ultraviolet spectrum has also been explored for the task of sky segmentation based navigation~\cite{stone2014sky} and tilt-invariant place recognition~\cite{stone2016skyline}. \cite{ramalingam2010skyline2gps} explored an automatic skyline segmentation approach for an upward-facing omnidirectional color camera. These omni-skylines were observed to be unique for a specific location in a city, particularly in GPS-challenged urban canyons. \cite{saurer2016image} developed a skyline extraction and representation method for localizing in mountainous terrains. \cite{pepperell2016routed} proposed a ``sky blackening'' method to remove the sky region from images in order to improve place recognition performance across day and night cycles. Beyond sky-based representations, variations in weather conditions can also be used to localize a static camera, by analyzing the temporal variations in the appearance of the viewed scene~\cite{jacobs2007geolocating}. Recently, \cite{Yu2018} presented an image descriptor based on the VLAD aggregation~\cite{jegou2011aggregating} of semantic edge features~\cite{yu2017casenet} defined between distinct semantic categories like building-sky and road-sidewalk. \cite{singh2016semantically} developed an image descriptor based on histograms of semantically-labeled ``superpixels''~\cite{ren2003learning, gould2009decomposing} within an image grid, and demonstrated its utility for geo-localization, semantic concept discovery, and road intersection recognition.

\subsubsection{Life-long Operation of VPR Systems}

Appearance variability of places can occur at vastly different time scales. At one extreme, time of day, weather fluctuations and even momentary lighting changes contribute towards intra-day appearance changes. Longer-term factors such as seasons, vegetation growth, and climate change also lead to appearance variability. A significant additional contributor to appearance variations are human activities, such as construction work, general wear and tear, updating of signage/billboards/fa\c{c}ades including increased digital signage, and abrupt changes to traffic flow. Compared to natural causes of appearance variability, the changes caused by human activities can be far less ``cyclical'' or predictable.

The fact that the appearance of places can change indefinitely and unpredictably imposes a requirement that localization and VPR systems remain accurate and robust under those continuous changes, throughout the lifetime of the system. Compared to the majority of research in VPR that has so far considered static datasets (though with significant appearance variations within the datasets), lifelong VPR is an under-investigated area. To handle continuous appearance changes, some approaches attempt to extract the ``semantic essence'' of a place that is independent of appearance, and to transfer the appearance of the place seen in a particular condition to unseen conditions~\cite{latif2018addressing,porav2018adversarial,anoosheh2019night}. To date, such methods have primarily been demonstrated on \textit{natural} variations such as time of day and seasons. Another paradigm continuously accumulates data to refine the system~\cite{churchill2013experience,doan2019scalable,dymczyk2015gist}. The practicality of such approaches depend on the ability to continuously collect data from the target environment at high frequency (e.g. weekly), which can be achieved using deliberate recording schemes (e.g. mapping vehicles~\cite{anguelov2010google}, taxi fleets~\cite{geyer2020a2d2, kesten2019lyft}) or opportunistic and crowd-sourced schemes such as user-uploaded videos~\cite{warburg2020mapillary} or webcams~\cite{jacobs2009global}. The fundamental challenge arises from the perpetually growing database, which demands a VPR algorithm that is scalable; the computational effort for inference and continuous refinement of the VPR system must grow slowly or not at all with the database size.

\subsection{Semantics to Deal with Challenging Environmental Conditions}

\subsubsection{Addressing Extreme Variations in Appearance and Viewpoint}
Semantic cues have been demonstrated to be of high utility for place recognition under challenging scenarios where appearance- and geometry-only features alone fall short, for example, dealing with extreme appearance variations across repeated traverses of the environment~\cite{garg2019robust}. These non-uniform appearance variations are often caused by weather, time of day and season cycles. \cite{toft2017long} presented an appearance-invariant localization system based on a 3D model of semantically-labeled points and curves, which were projected onto a single query image to minimize error for pose estimation. \cite{Garg17IrosImproving} proposed a temporal place segmentation approach based on semantic place categories~\cite{zhou2014learning} to improve VPR~\cite{milford2012seqslam} under significant appearance variations occurring both within and across the traverses. \cite{naseer2017semantics} developed a method to learn salient image descriptions based on regions of the image which are geometrically stable over large time gaps (across seasons and weather changes). They combined this representation with an off-the-shelf holistic representation to obtain a robust descriptor. 

Cross-view matching, typically referring to the problem of matching images across highly-varied viewpoints like aerial top-view versus ground front-view~\cite{lin2013cross, castaldo2015semantic, Gawel2018}, is a challenging task due to extreme variations in visual appearance and geometry between the matching pair, often requiring additional information to enable effective localization. \cite{lin2013cross} explored a cross-view geo-localization approach that used SVMs to learn relationships among image triplets: ground level image, its corresponding aerial image and a land cover attribute map. More recently, \cite{Gawel2018} presented a localization system for cross-view matching where aerial images were captured using UAVs. They developed a semantic blobs-based hybrid representation derived from pixel-wise semantic segmentation, and combined topological, metric, and semantic information for wide-baseline image matching, i.e., forward vs reverse and aerial vs ground.

In order to deal with \textit{simultaneous} extreme variations in scene appearance (day-night) and camera viewpoint (forward-reverse), \cite{garg2018lost} proposed Local Semantic Tensor (LoST) as a global image descriptor based on pixel-wise semantic labels and spatial aggregation of local descriptors derived from a CNN. They extracted keypoints from maximally-activated locations within the CNN feature maps, and performed semantic keypoint filtering and weighted local descriptor matching to re-rank the place matches for high-precision place recognition. \cite{garg19Semantic} extended this approach with a concatenated descriptor that encoded semantics both explicitly and implicitly, in the form of selective semantic feature aggregation and a deep-learnt VLAD descriptor~\cite{arandjelovic2016netvlad}, respectively.

\subsubsection{Linking Place Categorization with Place Recognition}
Bridging the gap between semantic place categorization and the place recognition problem, \cite{wang2018visual} proposed a 3-layer perception framework using a topological directed graph as a map of the environment. Each node represented a set of semantic regions, and each edge represented a set of rotation-recognition regions. This structural representation enabled the recovery of the optimal path for indoor semantic navigation based on the number of semantic regions encountered. \cite{Martinez-Gomez2016} developed a framework for semantic localization that used 3D global descriptors based on a Bag of Visual Words (BoVW) approach~\cite{sivic2003video} to train a classifier for categorizing rooms labeled as unique places.

\subsubsection{Semantics within High-Level Representations}
In this subsection, we discuss the use of image representations that capture shapes and patterns at a higher abstraction level, which may not be \textit{strictly} semantically meaningful to humans. To some extent, this can be attributed to feature extraction from higher-order layers of CNNs and methods like \textit{Edge Boxes}~\cite{zitnick2014edge} that provide a ``semantically-aware'' representation of whole images or image regions. \cite{sunderhauf2015place} explored the use of Edge Boxes to detect and represent ``ConvNet Landmarks'' for visual place recognition. \cite{cascianelli2016robust} represented the environment using a co-visibility graph of semantic image patches~\cite{zitnick2014edge}, where an edge was established only if patches were observed in the same image, unlike the common notation of edges as spatial relationships. \cite{seymour2018semantically} presented Semantically-Aware Attentive Neural Embeddings (SAANE) that fused appearance features and higher-order layers of the CNN to learn embeddings for matching places under a wide variety of environmental conditions. \cite{sunderhauf2015performance} demonstrated that the higher-order layers of CNNs encoded semantic information about a place, which can be utilized for partitioning the descriptor search space. \cite{garg2018don} showed that the fully-connected layers of a place categorization CNN, being semantics-aware, exhibited viewpoint-invariance and hence enabled place matching from opposing viewpoints, even with the additional challenge of significant appearance variations. The use of additional modalities like pixel-wise depth for visual place recognition has also been explored in the past~\cite{fiolka2013distinctive, cupec2015place, cieslewski2016point, garg2019look, maffra2019real, vankadari2020unsupervised, taubner2020lcd}, where meaningful representations in the form of surfaces~\cite{fiolka2013distinctive}, planes~\cite{torii201524}, and lines~\cite{cupec2015place, taubner2020lcd} have been shown to improve VPR performance.

\cite{Larsson2019} proposed FGSN (Fine-grained Segmentation Networks) based on k-means clustering of CNN embeddings for self-supervised pixel labeling and 2D-2D point correspondence. They showed that fine-grained clustering, even though non-semantic, is more suitable for visual localization than methods based on a limited number of semantic classes, for example, the 19 in Cityscapes~\cite{Cordts2016} and 66 in the Vistas~\cite{neuhold2017mapillary} dataset. However, it was also observed that pre-training on semantic segmentation networks improved performance significantly, suggesting semantics would still play a key, albeit different, role.

\subsection{Enabling Semantics for Robotics: Additional Considerations}

Methods developed primarily in other research fields like computer vision often do not readily translate to robotic systems~\cite{sunderhauf2018limits}. Many critical issues for successful robotics deployment are not addressed or not prioritized in many of the research fields from which robotics draws much of its inspiration. Robotic deployment has a range of additional constraints and opportunities compared to much dataset-based research, including the availability of multiple sensing modalities, limited computational resources, a focus on real-time and on-line deployment, and a range of problems including obscuration, clutter and uncertainty that are not adequately encountered in dataset-based research. Hence, further innovation is typically required to bridge the gap between a laboratory or dataset-demonstrated system and a deployable solution for robots ``in the wild''. In this subsection, we cover some of the relevant research bridging this divide that has not already been covered in previous sections.

\subsubsection{Efficiency}
In order to classify 3D objects in real-time, Maturana and Scherer~\cite{maturana2015voxnet} developed a simplified version of a 3D CNN~\cite{maturana20153d} for real-time object recognition. \cite{zeng2018rt3d} proposed \textit{RT3D}: a real-time 3D vehicle detection method based on pre-ROI-pooling convolutions~\cite{dai2016r} that accelerate the classification stage, thus completing detection in real-time with detection accuracy comparable to state-of-the-art. More recently, \cite{hou2020real} demonstrated real-time panoptic segmentation using a single-shot network based on a parameter-free mask construction operation that reuses dense object predictions via a global self-attention mechanism. 

\subsubsection{Noise, Occlusions and Clutter}
In order to deal with noise, occlusion and clutter in robotic sensing and point-based representation, \cite{bobkov2018noise} designed 4D CNNs for object classification that process robust point-pair based shape descriptors~\cite{wahl2003surflet, rusu2009fast, drost2010model, wohlkinger2011ensemble, birdal2015point} represented as 4D histograms. Occlusions also pose a challenge for object-based representation of the environment and is an active area of research in the context of object-based SLAM~\cite{tan2013robust, mu2016slam, gaudilliere2020perspective, feldmantowards, ok2019robust, liao2020rgb}. Moreover, occluded objects affect the evaluation of object detection systems~\cite{hoiem2012diagnosing} and require novel error measures to deal with partially-visible objects~\cite{Nicholson2019}. To deal with background clutter, Wang \textit{et al}.~\cite{wang2012could} proposed first removing the background, posing it as a binary classification task, before segmenting the point cloud data into semantic classes of interest for autonomous driving like cars, pedestrians and cyclists.

\subsubsection{Cost}
As a cost-effective alternative to a LiDAR-based 3D point cloud, Pseudo-LiDAR~\cite{wang2019pseudo} has been proposed to represent pixel depth estimated using stereo cameras as a 3D point cloud, leading to a significant performance improvement for 3D object detection. In a subsequent work, the authors presented Pseudo-LiDAR++~\cite{you2019pseudo} to improve depth estimation of faraway objects while also incorporating a depth signal from cheaper LiDAR sensors that typically have sparse 3D coverage. \cite{vianney2019refinedmpl} extended the concept of Pseudo-LiDAR to monocular systems, while also addressing the numerical and computational bottleneck of the dense Pseudo-LiDAR point cloud. Recently, Qian \textit{et al}.~\cite{qian2020end} developed an end-to-end Pseudo-LiDAR framework based on differentiable Change of Representation (CoR) modules to further improve detection accuracy.

\subsubsection{Uncertainty Estimation}
Uncertainty estimation is essential for semantic understanding and decision making during mapping and interaction. Current approaches for uncertainty estimation include approximations of Bayesian deep learning~\cite{mackay1992practical} such as dropout sampling~\cite{gal2016dropout}, deep ensemble methods~\cite{lakshminarayanan2017simple} and the recently proposed Stochastic Weight Averaging-Gaussian (SWAG)~\cite{maddox2019simple}. 
For object detection, Monte Carlo Dropout Sampling was used in~\cite{miller2018dropout} to measure label uncertainty. Other works proposed estimating the uncertainty in detecting the location of the object in the image, introducing the probabilistic bounding box~\cite{hall2020probabilistic}. In~\cite{ wang2020inferring}, an alternative representation of probabilistic bounding boxes was introduced through the spatial distribution of a generative bounding box model. In~\cite{harakeh2020bayesod}, probabilistic bounding boxes were generated by replacing the standard non-maximum suppression (NMS) step in the object detector by a Bayesian inference step, allowing the detector to retain all predicted information for both the bounding box and the category of a detected object instance. 

Dropout sampling for semantic segmentation can be slow for robotics applications, prompting ~\cite{huang2018efficient} to propose a Region-based Temporal Aggregation (RTA) method which leverages temporal information using a sequence of frames. Other works expressed uncertainty as a measure of quality in the predictions, such as predicting the quality of the IoU for semantic segmentation~\cite{rottmann2019uncertainty} or per-frame mAP (mean average precision) in the case of object detection~\cite{rahman2020performance}. Recently, \cite{blum2019fishyscapes} showcased that unseen object categories can be mislabeled by state-of-the-art semantic segmentation networks with high certainty. To mitigate this the authors presented the Fishyscapes dataset, for pixel-wise uncertainty estimation for autonomous driving to enable detection of such anomalous objects.

\subsubsection{Multi-modal and Non-Vision Based Approaches} Given that robots typically carry a suite of sensors, effective integration of information from multiple modalities is also an active area of research. Towards this goal, \cite{Pronobis2010} developed a multi-modal semantic space labeling system leveraging information from camera, laser scanner, and wheel odometry sensors. This work was further extended in~\cite{Pronobis2011} to incorporate high-level information as ``properties'' of a place, defined in terms of shape, size, appearance, and doorway (binary). While using multiple modalities, authors in \cite{Mozos2012} concluded that although combining features from depth and color drastically reduce uncertainty, depth information contributes more to the performance due to inherent illumination-invariance. For synchronized multiple modalities, \cite{Jung2016} proposed a Local N-ary Patterns (LNP) descriptor to describe relationships among neighboring pixels of reflectance and depth images, as an extension of the Local Binary Pattern descriptor~\cite{Ojala2000}. As a general framework for semantic place categorization using any sensor modality, \cite{Premebida2015} demonstrated the superiority of DBMM (Dynamic Bayesian Mixture Models) based temporal inference over a commonly used SVM approach.

With the success of deep learning based methods, amenable to early, intermediate and late fusion techniques, use of multiple modalities has now become much more accessible~\cite{ramachandram2017deep}, including in semantic-based approaches. The complementary properties of the modalities can be harnessed via multi-modal fusion. Recently, Feng \textit{et al}.~\cite{feng2020deep} reviewed the state-of-the-art deep learning based approaches for object detection and semantic segmentation that employ multi-modal fusion, particularly exploring the answers to the ``\textit{what}, \textit{when} and \textit{how} to fuse'' questions. Chen \textit{et al}.~\cite{chen20153d} proposed a 3D object proposal method using stereo cameras, posing it as an energy minimization problem exploiting object size priors and depth-informed features like height above the ground plane, free space and point cloud density. In subsequent work~\cite{chen2017multi}, the authors presented the \textit{MV3D} (Multi-View 3D) object detection network, which combines RGB images and a LiDAR's front and Bird's-Eye Views (BEV) to enable effective multi-modal fusion based on interactions between different layers of the network. A similar multi-modal fusion is proposed in~\cite{ku2018joint}, dubbed \textit{AVOD} (Aggregate View Object Detection), with a novel use of high-resolution feature maps, $1\times1$ convolutions and a look-up table for 3D anchor projections, achieving real-time and memory-efficient high detection performance. Vora \textit{et al}.~\cite{vora2019pointpainting} presented \textit{PointPainting} as a way to ``decorate'' the LiDAR point cloud with the semantic segmentation output of color images for improved 3D object detection, addressing the limitations of previous fusion concepts that led to ``feature blurring''~\cite{liang2018deep, wang2018fusing} and limited maximum recall~\cite{qi2018frustum, yang2018ipod, wang2019frustum}.

A subset of research in place categorization semantically categorizes places using sensors other than those used for visual perception. With the increasing focus on Internet of Things and the ubiquity of edge-based sensors, upcoming solutions for semantic place categorization might be able to leverage these additional sources of information and existing approaches to effectively use multi-stream data. \cite{Krumm2013} proposed place classification of geographical locations based on individual demographics, timing of visits, and nearby businesses using government diary data. The demographic and temporal features of individual visits included age, gender, arrival and departure details and season. The place categories included home, work, restaurant, library, place of worship and recreation. \cite{Krumm2015} extended this work by also considering additional cues like sequential information in the form of the periodicity of the individual visits, cross labels extracted from multiple visits to the same place, travel distance, and place co-occurrences. \cite{Lv2016} presented a semantic place classification system using GPS trajectories of users. Their method is based on visit-level features including day of week, time of day, duration, and response rate, which were then used to infer place-level features. \cite{Roor2017} developed a semantic place labeling system based on multiple sensor sources including Bluetooth, smartphone, GPS, motion activity, WLAN and time, where place categories included home, work, friend and family, nightlife and education. 

Depending on the application type, the use of semantics can vary significantly. For example, \cite{Weiss2010} proposed a particle filter-based semantic state estimation system for semantic mapping and navigation of agricultural robots. Here, the ``semantic states'' (classes) were defined according to the topology of crop rows, considering the ``side'', ``start'', ``end'', and ``gap'' as the semantic elements of the agricultural field.

\section{PRACTICAL ASPECTS: APPLICATIONS AND ENHANCERS}
\label{secn:discussion}
Semantics are clearly a promising avenue for enhancing the capabilities and utility of robotic systems. At this point in time however, the uptake of systems that make significant use of semantics varies significantly across different application domains, from autonomous vehicles to service robots to augmented reality applications. There are also a range of technology advances that will likely facilitate further advances in semantics-based robotics, ranging from advances in compute capability (and an associated reduction in cost, bulk, and power consumption, all relevant benefits where robotic systems are concerned) as well as in online ``in the cloud'' computational and data resources. In this section we cover these \textit{practical considerations}, discussing current and future application scenarios for semantics in robotics, and cover the technology advances that will support these technologies into the future.

\subsection{Applications}

A number of robotics and autonomous platforms employing some degree of semantics can be seen in use today, with a much larger number in use at the research or proof of concept stage. Robotics and autonomous platforms span a wide range of application scenarios including domestic services: house cleaning (iRobot Roomba~\cite{roomba}), lawn mowing (Robomow RS~\cite{robomaw}), pool maintenance (Zodiac VX65 iQ~\cite{poolCleaner}); scientific exploration: underwater in deep sea (AUV Sentry~\cite{auvSentry, kaiser2016design}), planetary rovers (Curiosity~\cite{curiosity, manning2014mars}); hospitals: intelligent transportation (Flexbed~\cite{wang2014intelligent}), socializing with patients~\cite{alves2015social}, surgical procedures (Da Vinci~\cite{davinci, byrn2007three}), and providing care~\cite{hanheide2017and,fattal2019sam}; social interaction (Nao~\cite{Nao}, Pepper~\cite{pepper}, Vector~\cite{vector}); telepresence (Beam~\cite{beam}); hospitality \cite{pinillos2016long}; shopping centers and retail~\cite{kejriwal2015product, kanda2010communication, gross2009toomas}; office environment~\cite{kumar2014tea}; logistics and fulfillment~\cite{huang2015robotics}; last mile delivery (Starship~\cite{starship}); and many others~\cite{roboList}.

In the following sections, we discuss the major application areas from the perspective of different autonomous platform types (UAVs, Service Robots, Static Platforms) as well as covering particular application domains (Autonomous Driving, Augmented Reality and Civilian Applications).

\subsubsection{Drones and Unmanned Aerial Vehicles (UAVs)}
Drones and UAVs represent one of the most resource-constrained robotic platforms, where balancing factors like flight time, payload capacity, cost, power, real-time operation and sensor choice requires careful consideration. As the application and operating scenarios can vary significantly, but are generally more constrained compared to say autonomous vehicles, research with regards to the use of visual semantics on drones is still relatively preliminary.

Numerous benchmark datasets have focused on UGVs, leading to significant advancements in research for enabling semantic scene understanding~\cite{geiger2012we, Cordts2016, geyer2020a2d2}. However, the same is not true for UAVs. \cite{du2018unmanned} recently presented a UAV object detection and tracking benchmark that particularly highlights the contrast between the challenges of vision research for UAVs versus UGVs. The primary differences occur due to variations in object density per frame, relative sizes of observed objects, viewpoint, motion and real-time requirements. To further the research efforts in this domain, \cite{zhu2018vision} released a benchmark challenge \textit{VisDrone2018} for object detection and tracking with 2.5 million annotated instances of objects within 180K video frames. \cite{vosselmanuavid} presented UAVid, an urban street scene segmentation dataset, particularly aimed at high-resolution slanted views from low-altitude flights, in contrast to the more common top view-based datasets~\cite{rottensteiner2014results, campos2016processing, debes2014hyperspectral, demir2018deepglobe}. In a similar vein, \cite{nigam2018ensemble} released \textit{AeroScapes}, an aerial-view dataset of 3269 images with dense semantic annotation, captured using a fleet of drones. The scarcity of labeled data has also motivated researchers to explore alternative ways to utilize the existing data and approaches to semantic scene understanding. The authors in~\cite{nigam2018ensemble} used an ensemble approach to semantic segmentation of aerial drone imagery, based on knowledge transfer via progressive fine-tuning through different source domains. Similarly, \cite{benjdira2019unsupervised} explored the use of GANs~\cite{goodfellow2014generative} for unsupervised domain adaptation to improve cross-domain semantic segmentation in aerial imagery.

With the availability of better benchmark datasets and tools, there's the potential for advances in semantic scene understanding for UAV applications mirroring those made for road-based vehicles. Currently, UAVs are used in a diverse range of application scenarios that differ in operating environments as well as the end task. \cite{leira2015automatic} demonstrated a Search And Rescue (SAR) system to detect and track objects on ocean surfaces where accessibility by other means is typically impractical. Similarly, natural environments that are prone to disasters like volcanic eruptions, forest fires or landslides can be better accessed by UAVs, which can aid in detecting and assessing such situations~\cite{zheng2017bi}. Nature conservation and wildlife management are also key areas where UAVs can offer a significant practical advantage over other platforms. \cite{van2014nature} presented a nature conservation drone that detects, tracks and counts various animals in the wild. Extending this idea to the use of multiple devices, \cite{rivas2018detection} developed a software platform that enables the use of different multirotors for cattle detection and management, mitigating the need to attach GPS devices to animals. \cite{bondi2018spot} developed Systematic POacher deTector (SPOT) based on drones to spot poachers at night time. Beyond natural environments, UAVs have also been employed in the construction industry for site management and control~\cite{huang2018construction}. Another distinct application area for UAVs is last mile delivery, which is already well on its way from research to commercial deployment~\cite{wing}. As both the end-task and operating environment vary widely, the use of multiple sensors beyond just RGB cameras to achieve the objective is common, for example, the use of thermal cameras~\cite{leira2015automatic, bondi2018spot} and LiDAR~\cite{kellner2019new}.

Autonomous UAV navigation will also benefit from advances in semantic understanding of the world. Driven in significant part by advances in the field of deep learning, autonomous navigation is a topic of active exploration in the context of drones and UAVs, especially enabling end-to-end autonomous navigation. \cite{sadeghi2016cad2rl} proposed Collision Avoidance via Deep Reinforcement Learning (CAD$^2$RL) where training is done only using simulated 3D CAD models, demonstrating generalization to real indoor flights. \cite{loquercio2018dronet} developed DroNet that takes as an input a monocular image and outputs the steering angle along with a collision probability. Distinct from~\cite{sadeghi2016cad2rl} and other methods that use synthetic data for training, \cite{loquercio2018dronet} explored the use of ground vehicle driving data from city streets and demonstrated generalization to variations in viewpoint (from high altitude flights) and types of environment (parking lots and corridors). This study in particular shows that there are underlying meaningful semantic concepts that are not specific to a robotic platform or operational domain.

Besides using data from simulation or a different application domain for learning, a somewhat uncommon paradigm for learning has been explored in~\cite{gandhi2017learning}, based on real crash flights. The authors sample naive trajectories and crash into random objects (11500 times) to form a crash dataset, which is then used to learn a UAV navigation policy, demonstrating a drone's ability to navigate in an extremely cluttered and dynamic environment. Apart from a completely end-to-end pipeline, modular approaches have also been explored where perception and control are handled separately~\cite{devin2017learning, clavera2017policy, muller2018driving}. Similarly, \cite{loquercio2019deep} demonstrated autonomous navigation for a racing drone via sim-to-real transfer of its CNN training for predicting goal direction using data generated from domain randomization. More recent work in the field of drone navigation includes learning to fly from a single demonstration~\cite{kaufmann2019beauty}, deploying navigation capability on mW-scale nano-UAVs~\cite{palossi201964}, and motion planning based on an opponent's reactions~\cite{spica2020real}. 

Autonomous UAV navigation requires a high-level understanding of the environment, which can either be learnt via end-to-end training techniques or via modular methods that help to semantically reason about the surroundings. Towards the goal of semantic navigation, \cite{mandel2020towards} employed the Observe-Orient-Decide-Act-loop (OODA-loop) decision-making theory~\cite{boyd1987discourse, boyd2018discourse}, and incorporated semantic information from the environment into the decision making process of a UAV to dynamically adjust its trajectory. A semantic reasoning capability can potentially enhance the performance in the end task of a practical application by adding situational awareness. In this vein, \cite{cavaliere2016towards} proposed to improve object detection and tracking by integrating ontological statements-based semantics in their tracking method for improved awareness of drones in critical situations~\cite{cavaliere2017semantically}. With a similar goal, \cite{jeon2019relationship} leveraged the relationships between objects and defined an ontology to describe the objects surrounding a UAV to better detect a threat situation.

Further advances in navigation and scene understanding for UAVs, particularly those based on semantic reasoning, will play a critical role in improving UAV autonomy and broadening the range of applications where deploying drones is feasible.

\subsubsection{Service Robots}
Much robotics research has been driven by a motivation to develop robots that can provide a range of services to humans. Within each of the application areas, there are several aspects of service robotics that need to be taken into account. Recent review and survey articles have discussed the uses, scope, opportunities, challenges, limitations and future of service robots in various contexts: human-centric approaches~\cite{he2017survey}, navigating alongside humans~\cite{cheng2018autonomous}, cultural influences~\cite{cheng2018autonomous}, needs in healthcare~\cite{vanni2017need}, acceptance in a shopping mall~\cite{niemela2017monitoring}, consumer responses to hotel service robots~\cite{tussyadiah2018consumer}, requirements of the elderly with cognitive impairment~\cite{korchut2017challenges}, uncertainty in natural language instructions~\cite{muthugala2018review}, roles in the hospitality industry~\cite{rosete2020service}, mechanical design~\cite{bakri2019review}, hospitals~\cite{mettler2017service}, social acceptance in different occupational fields~\cite{savela2018social}, welfare services~\cite{aaen2018robots}, restaurant services~\cite{tuomi2019service}, service perception and responsibility attribution~\cite{gracia2020robots}, and planning and reasoning in general-purpose service robots~\cite{walker2019desiderata}. These studies highlight the aspects of service robotics which are not necessarily technical but highly relevant to their effective deployment.

A commercially-viable service robot typically requires a hardware design that improves its utility in a range of application scenarios. For example, the modular design of Care-o-bot~\cite{careobot} enables application-specific modifications like replacing one of the arms by a serving tray, or using just its mobile platform for serving, beyond its prescribed use cases in grocery stores, museums or as a butler. Similarly, PR2~\cite{pr2} is capable of performing house chores like cleaning a table and folding a towel, as well as fetching objects. Both these robots are equipped with a sensor suite that enables them to perform autonomous navigation, grasping and manipulation. These competencies are an active area of research which drives continuous improvement in the robotic solutions available. The underlying software platform for PR2 and many other robots is supported by the open-sourced Robot Operating System (ROS)~\cite{ros}, making research studies easier and systematic~\cite{do2018accurate, berenson2012robot}. Beyond ROS, specialized software development kits are typically available for robots to control or access their sensor data as is the case with Spot~\cite{spot}, a new commercially available service robot that can move through trotting in both indoor and outdoor environments, while being capable of climbing stairs, opening doors and fetching a drink.

Apart from the hardware designs and underlying operating kernels, in order to complete a task and provide a service, a robot requires advanced perception and control abilities. This is typically achieved by using multiple sensors and performing tasks in a modular fashion. \cite{bellotto2008multisensor} proposed a multi-sensor based human detection and tracking system and demonstrated its portability on different robotic platforms using a Pioneer and Scitos robot. \cite{kumar2014tea} presented a tea-serving application using P3DX robot in an office environment, where an overall environment awareness was achieved by developing individual competencies in isolation like line following, obstacle avoidance, empty tray detection, approaching-hand detection and person following. Recently, aiming for precision agriculture, \cite{kim2020an} presented an intelligent pesticide spraying robot that semantically segments the fruit trees that require spraying. Using semantic reasoning, perception, control and interaction abilities can be further improved. \cite{Wei2012} demonstrated smart wheelchair navigation using semantic maps, emphasizing enhanced safety, comfort, and obedience due to the use of semantics. \cite{niekum2013incremental} presented a semantically-grounded learning-from-demonstration approach for a furniture assembly task using a PR2. \cite{Savage2019} developed a semantic-reasoning based method for interpretation of voice commands for improved interaction between a service robot and humans. \cite{Ramirez-Amaro2015b, Ramirez-Amaro2017, Ramirez-Amaro2019a} demonstrated the use of semantic representations for human/robot activity understanding using humanoid and industrial robots. More recently, \cite{zeng2020semantic} proposed Semantic Linking Maps (SLiM) that exploit common spatial relations between objects to actively search for an object based on probabilistic semantic reasoning. 

Many service robots that work for or alongside humans require the ability to extract, represent and share semantic knowledge with their users or co-workers to perform these services intelligently and sometimes jointly. Semantic knowledge representation in the form of semantic maps enriches the metric and/or topological spatial representation that these robots traditionally carry. Similarly, semantically reasoning about the service task will close the gap between the way humans and robots understand their environment, which then facilitates more natural robot-user communication.

\subsubsection{Static Platforms}
\label{sec:disc-staticPlatforms}
Other than robotic platforms, a number of practical applications also involve static platforms based on perception sensors. We briefly cover these platforms here because it's likely that in future many robotic systems will operate in a ``shared perception'' environment where their onboard sensing is extended by static sensing systems. Many of the capabilities initially developed in a surveillance context, especially sophisticated scene understanding driven by semantics, are also likely to be relevant as these capabilities are ported into robotic platforms.

Visual surveillance is one such example. Video surveillance via Closed-Circuit TeleVision (CCTV) has long been used around the world, with millions of cameras installed. Recent surveys have highlighted the diverse use-cases of visual surveillance and its research advances~\cite{sharma2019visual, jones2017visual, bouwmans2019natural, bouchrika2018survey, kumaran2019anomaly, olatunji2019video, singh2018applications, tripathi2018suspicious}. The application areas for visual surveillance are vast and pose unique challenges depending on the task at hand, many of which are relevant to robotics.

\cite{bouwmans2019natural} highlighted the use of visual surveillance in natural environments, for example, studying social behavior of insects in a group~\cite{knauer12005application}, analyzing climate impact based on interactions between natural beings~\cite{dickinson2010automated}, behavior adaptability under tough environmental conditions~\cite{spampinato2014texton} and inspiring robots to mimic locomotion of certain animal species~\cite{iwatani2016position}. The authors also discussed the methods and challenges of background subtraction, often the first step in the process, particularly when tracking a moving object under challenging conditions like those posed by a marine environment. \cite{bouwmans2019human} presented the applications and challenges of visual surveillance of human activities. The authors covered a wide range of uncontrolled environment scenarios: traffic flow monitoring~\cite{sarkar2020microscopic}, detecting traffic incidents~\cite{monteiro2008robust} and anomalies~\cite{kumaran2019anomaly}, vehicle parking management~\cite{cho2016automatic}, safety operations at public places like train stations~\cite{tarrit2018vanishing} and airports~\cite{bouchrika2016towards}, maritime administration~\cite{hu2019single} and retail store monitoring~\cite{rashmi2018rule}. As with the natural environments, monitoring human activities in man-made environments also poses similar challenges including background subtraction when tracking a moving object. Furthermore, as human activities can be unpredictable and have more serious implications for false negatives (not being able to detect an incident), visual surveillance in this particular context demands particularly robust and high performance techniques.

Irrespective of the operating environment, the key enablers of a visual surveillance system are the fundamental abilities to holistically and continuously understand the visual observations. These basic abilities are major areas of research and directly impact the performance of the surveillance system. Research includes gait recognition~\cite{bouchrika2018survey, khan2017gait}, person re-identification~\cite{zheng2016mars, xu2018attention}, face recognition~\cite{rao2017attention, masi2018deep, sharif2017face}, single-object tracking~\cite{kristan2018sixth}, multi-object tracking~\cite{milan2016mot16, gaidon2016virtual, garg2015hierarchical} especially in dense crowded scenes~\cite{dendorfer2020mot20}, human activity recognition~\cite{tripathi2018suspicious}, tracking across non-intersecting cameras~\cite{bouchrika2016towards}, object detection especially in the context of abandonment and removal~\cite{tripathi2019abandoned} and incorporating contextual knowledge~\cite{rashmi2018rule}. Many of these capabilities are also critical to robotic platforms, especially as they operate in human-rich environments.

The fundamental abilities required for visual surveillance and the challenges associated with them together dictate the need for better approaches that are based on semantic reasoning. The current solutions tend to address challenges that are often specific to the operating environment and the type of object being tracked. A more generalizable solution would require semantic scene understanding which can either be inferred from the observed scene or through the use of knowledge databases. With semantic reasoning, the fundamental task of detecting and tracking an entity of interest can likely be performed in a more robust and generalized manner. 

\subsubsection{Autonomous Vehicles}

Industry has been largely responsible for increased research activity in the autonomous vehicle space, as evidenced by an increasing number of surveys and reviews dedicated to autonomous driving. Autonomous driving is a great test case for semantics, because it has rapidly become clear that classical robotics techniques, like SLAM, are clearly insufficient \textit{by themselves} for enabling autonomous driving, and that a richer understanding of the environment is almost certainly required.

Recently, \cite{feng2020deep} reviewed \textit{multi-modal} object detection and semantic segmentation for autonomous driving, describing various methods, datasets and on-board sensor suites while highlighting associated challenges. \cite{arnold2019survey} surveyed 3D object detection methods for autonomous driving applications. \cite{grigorescu2019survey} presented a survey of deep learning techniques for autonomous driving. \cite{yin2017use} reviewed the available datasets for self-driving cars. \cite{janai2017computer} reviewed problems, datasets and state-of-the-art computer vision methods for autonomous vehicles and \cite{bengler2014three} reviewed driver-assistance systems. The tremendous push for autonomous cars from different companies has led to the availability of enormous labeled datasets captured from fleets of vehicles mounted with many sensors~\cite{caesar2020nuscenes, geyer2020a2d2, sun2020scalability}. Both industrial and academic researchers have invested significant effort in understanding and solving the challenges posed in developing high levels of autonomous driving. 

Scene understanding based on benchmark datasets has its limitations as it does not account for a number of scenarios which an autonomous robot might encounter in real-world. For example, detecting a ``stop sign'' may not be trivial due to occlusions, weather and lighting conditions, infrastructural variations, special sub-categories and temporary road blocks where humans carry the stop signs~\cite{ytAndrej2020TeslaAI, p1007michael2020}. One of the practical solutions to such object recognition problems is \textit{active learning} where approximate detectors can be used to automatically sample more data for less confident predictions, especially under a map-vision disagreement, to train better classifiers or handle special cases, as adopted by Tesla, Inc.~\cite{ytAndrej2020TeslaAI}. The Tesla talk~\cite{ytAndrej2020TeslaAI} also highlights how general evaluation methods are supplemented by curated unit tests for measuring the scene understanding capability of an autonomous car before deploying an updated solution. Furthermore, semantic scene understanding for autonomous driving is at times limited by differences in road infrastructure and traffic rules across different geographies, for example, left- versus right-lane driving and differences in positioning of traffic lights near intersections. Therefore, learning general trends via deep learning often needs to be complemented by incorporating prior knowledge in the algorithm implementation and sensor specification~\cite{articleBoschAD}.

Given the range of challenges in enabling autonomous driving, some focus has shifted to task-specific and highly-engineered solutions. Such approaches include multi-modal object detection (Uber)~\cite{liang2018deep} and semantic segmentation (Bosch)~\cite{feng2020deep}, Pseudo-LiDAR~\cite{wang2019pseudo} based 3D object detection (Huawei)~\cite{vianney2019refinedmpl}, multi-view 3D object detection (Baidu)~\cite{chen2017multi}, sequential fusion (nuTonomy)~\cite{vora2019pointpainting}, monocular 3D object detection~\cite{chen2016monocular}, and learning near-accident driving policies (Toyota)~\cite{cao2020reinforcement} . In particular, application-specific biases or assumptions have proven to be of benefit. \cite{chen2016monocular} proposed 3D object detection from a single monocular image in the context of autonomous driving, leveraging various priors suited to the application, for example, size, shape and location of objects, ground plane context and semantic class selection. \cite{yi2020segvoxelnet} presented SegVoxelNet, which utilizes semantic and depth context for 3D object detection using LiDAR point clouds to better identify ambiguous vehicles on road. Researchers are making good progress towards solving the more specific and narrow problems within the autonomous driving field~\cite{chen2016monocular, chen20153d, ku2018joint, zeng2018rt3d, li20173d, wang2012could}.

Autonomous driving and driving-assistance systems need to solve a variety of fundamental and high-level tasks. In order to make sense of raw sensor data, continuous effort is being made to improve semantic scene understanding using various techniques including multi-task fusion~\cite{liang2019multi}, multi-sensor fusion~\cite{liang2018deep, liang2019multi}, panoptic segmentation~\cite{kirillov2019panoptic, xiong2019upsnet} and open-set instance segmentation~\cite{wong2020identifying}. With enhanced semantic reasoning, higher-level tasks for autonomous driving can be better addressed, for example, motion planning~\cite{zeng2019end, sadat2019jointly}, 3D object tracking~\cite{frossard2018end}, multi-object tracking~\cite{dendorfer2020mot20}, pedestrian behaviour prediction~\cite{jain2019discrete}, long-range human trajectory prediction~\cite{mangalam2020not}, turn signal-based driver intention prediction~\cite{frossard2019deepsignals}, predicting a pedestrian's intention to cross~\cite{liu2020spatiotemporal} and intention prediction in the form of trajectory regression and high-level actions~\cite{casas2018intentnet}. Furthermore, for real-time operations, efficient methods are being developed for semantic segmentation~\cite{oh2020segmenting, zhang2018efficient}, 3D object detection~\cite{yang2018pixor, luo2018fast}, online multi-sensor calibration~\cite{zhuo2020nline}, semantic stereo matching~\cite{Dovesi2019} and object tracking~\cite{luo2018fast}. As the current systems rely on high-definition maps for localization and detecting road infrastructure and fixed objects, compressing such maps for efficient operations is also being explored~\cite{wei2019learning, barsan2018learning}. Semantic mapping, particularly suited to autonomous driving on roads, is being explored at deeper levels involving lane detection in complex scenarios~\cite{bai2018deep}, road centerline detection~\cite{mattyus2018matching, wang2016torontocity}, lane boundary estimation~\cite{homayounfar2018hierarchical}, complex lane topology estimation~\cite{homayounfar2019dagmapper}, drivable road boundary extraction~\cite{liang2019convolutional}, crosswalk drawing~\cite{liang2018end}, and aerial view-based road network estimation~\cite{mattyus2017deeproadmapper}. Recent research has also demonstrated that using semantic maps can improve performance at tasks like object detection~\cite{yang2018hdnet} and localization~\cite{ma2019exploiting}.

Autonomous driving is a challenging task and one that is likely to rely heavily on vehicles having a deep and accurate understanding of how the world around them functions. As research in this field continues it will be interesting to see how far progress is made adopting a road-driving-specific focus, versus a broader understanding of the world, and how many of the insights gained are then relevant to other robotic domains like service robotics and drones. Will semantic advances for autonomous vehicles provide the final piece of the puzzle to safely and widely deployable highly autonomous on-road vehicles? And will these advances be of high relevance to other robotic domains, or will there be domain-specific aspects of how semantics are learnt and used?

\subsubsection{Civilian applications} 
A number of vision and robotics based solutions have been developed with primary applications in civilian society. This includes visual surveillance, robotic search and rescue operations, social interactions, and disaster management. Beyond static surveillance, discussed in the Subsection~\ref{sec:disc-staticPlatforms}, visual surveillance based on moving cameras is also becoming more common~\cite{yazdi2018new}. This not only increases the scope of surveillance applications but is also now more feasible due to the availability of better hardware and software technologies. This includes body-worn cameras of police or defence personnel~\cite{stoughton2017police} and crowd monitoring using drones or ground robots. The advanced sensing capability and maneuverability along with low cost and size makes UAVs a great choice for a standalone surveillance system~\cite{saeed2017argus}, although their effect on privacy and civil liberty remains an active topic of discussion~\cite{bracken2016domestic}. Visual surveillance using UAVs has been explored in various contexts including change detection~\cite{mesquita2019fully}, scene classification for disaster detection~\cite{zheng2017bi}, construction site management~\cite{huang2018construction}, and border patrol~\cite{minaeian2016effective}. Similarly, ground robots (UGVs) have also been used for indoor~\cite{saad2016room} and outdoor surveillance~\cite{meghana2017design}. \cite{yazdi2018new} highlighted some of the key challenges of visual surveillance based on moving cameras, including dealing with abrupt motion variations, occlusions, power and compute time constraints and increased background complexity.

Apart from surveillance, another important class of civilian applications is in disaster and crisis management. \cite{kostavelis2017robots} reviewed the use of robots in this context. A recent example is the last mile delivery operations run by Starship Technologies~\cite{starship} during the COVID-19 pandemic. Similarly, airborne deliveries by Wing~\cite{wing} and other such platforms can help in providing critical supplies during floods, landslides or bushfires, where ground accessibility is poor or non-existent. For operation in disaster situations, remote teleoperation is still a primary control mechanism - thus mitigating its corresponding challenges is also important. In this vein, \cite{fuchida2018arbitrary} explored the use of arbitrary viewpoints to mitigate blind spots for improved teleoperation of a mobile robot. Beyond vision-based sensing, \cite{asokan2016armatron} used ARMatron, a wearable glove as a gesture recognition and control device to better communicate with a remote robot. Furthermore, disaster management scenarios also require better communication protocols for robots to avoid system failure. For example, a robot-drone-fog device team may communicate by creating an ad-hoc network in a search and rescue mission~\cite{dey2017semantic}. Finally, sensor fusion is also a critical requirement for many such tasks, as demonstrated by~\cite{lee2016drone} for robot-assisted victim search.

The end goals for most robotic applications require both \textit{detection} and \textit{action} processes to work in tandem. Robots need to have robust means of understanding their environment, and that understanding has to be consistently verified as the robot acts. As also discussed previously, situational awareness can be attained by semantic reasoning~\cite{cavaliere2017semantically, jeon2019relationship}. Furthermore, semantic understanding of the robot's surroundings also enhances the possibilities for interacting with those surroundings, including with humans.

\subsubsection{Augmented Reality}

Over the past decade, Augmented Reality (AR) applications have become a relatively common phenomenon. Augmented reality is also highly relevant to robotics and covered here because many of the capabilities - such as localization - are shared across both areas, and both rely on an enhanced understanding of the environment that moves beyond simple geometry.

\cite{billinghurst2015survey} recently reviewed AR in depth, covering its various components ranging from tracking and display technologies to design, interaction, and evaluation methods, while also highlighting future research directions. Other recent surveys and reviews include a summary of major milestones in \textit{mobile} AR since 1968~\cite{arth2015history}; opportunities and challenges facing AR~\cite{azuma2016most, van2010survey}; AR trends in education~\cite{bacca2014augmented} and browsing~\cite{grubert2011augmented}; AR in the medical area including developments and challenges~\cite{chen2017recent}, radiotherapy~\cite{cosentino2014overview}, image-guided surgery~\cite{kersten2013state}, neurosurgery~\cite{meola2017augmented}, endoscopic sugery~\cite{wang2016visualization}, laproscopic surgery~\cite{bernhardt2017status}, and health education~\cite{monkman2015see, zhu2014augmented}; and general trends in AR research spanning from 2008 to 2018~\cite{kim2018revisiting}, following a previous decade-span survey~\cite{zhou2008trends} and other older surveys~\cite{azuma2001recent, azuma1997survey}. As noted in~\cite{billinghurst2015survey}, earlier attempts at developing an AR prototype date back to 1968~\cite{sutherland1968head} and the current widespread use can be be witnessed in games, education and the marketing industry. While some of the initial work in the field of AR had been primarily based on sensors like gyroscopes, GPS, and magnetic compasses~\cite{azuma1999motion, feiner1997touring, baillot2001authoring, hollerer1999exploring, piekarski2001tinmith, thomas1998wearable, satoh2001townwear}, vision-based localization has gradually become one of the key areas of research focus, in order to achieve high accuracy and a more immersive experience~\cite{behringer1999registration, ribo2002hybrid, vacchetti2004combining, klein2003robust, klein2004sensor, reitmayr2006going}. 

With visual SLAM and high-level scene understanding strongly tied to AR, the latter has long been a part of the research motivation for the vision and robotics research community. Real-time dense visual SLAM methods, for example, KinectFusion~\cite{newcombe11kinectfusion, izadi2011kinectfusion} paved the way for accurate surface-based reconstruction, enabling more immersive AR experiences in real-time. Doing away with non-parametric representations~\cite{newcombe11kinectfusion} by instead using semantics in the form of known 3D object models, SLAM++~\cite{Salas-Moreno2013} demonstrated a context-aware AR application using an object-based SLAM pipeline, where virtual characters completed the task of path finding, obstacle avoidance, and sitting on real-world chairs. In subsequent work~\cite{salas2014dense}, the authors developed a real-time dense planar SLAM system and demonstrated its use in augmenting planar surfaces, for example, replacing carpets and styles on the floor and overlaying a Facebook ``wall'' web page on a real wall. Further to that work, \cite{whelan2016elasticfusion} proposed ElasticFusion: a surfels-based map-centric approach to real-time dense visual SLAM and light source detection, leading to a more realistic augmented reality rendering.

The application areas of AR are vast, for example, in education: teaching electromagnetism~\cite{ibanez2014experimenting}; in e-commerce: \textit{Webcam Social Shopper}~\cite{webcamSocialShopper}, a virtual dressing room software and IKEA's virtual placement AR system~\cite{ikea}; in retail stores: \textit{Magic Mirror}~\cite{magicMirror} by Charlotte Tilbury; in games: PokemonGo's popular pokemon interaction~\cite{pokemongo}; in navigation: \textit{Live View} by Google Maps~\cite{googleLiveView}; and in medical: VeinViewer, a subsurface structure visualizer~\cite{zeman2011projection}, TRAVEE, an augmented feedback system for neuromotor rehabilitation~\cite{caraiman2015architectural}, PalpSim, a needle insertion training system~\cite{coles2011integrating}, training for planning tumour resection~\cite{abhari2014training} and 3D intraoperative imaging and instrument navigation~\cite{elmi2016surgical}. Image/3D registration is one of the critical components for enabling an accurate AR application. In medical procedures, this task is difficult due to a limited field-of-view vision, rapid motion, organ deformation, lack of sufficient texture, occlusion, clutter, and the inability to add artificial markers. Visual SLAM techniques which are primarily aimed at autonomous navigation for mobile robots typically make assumptions that limit their transferability to other domains, leaving significant research to be done in medical applications~\cite{marmol2019dense, chen2018slam, chen2017real}. 

In an attempt to further advance this technology, a number of software platforms like ARToolKit~\cite{kato1999marker, artoolkitx}, Vuforia~\cite{vuforia}, ARCore~\cite{arcore} and EasyAR~\cite{easyar} have been developed that help in rapidly enabling mobile AR applications. Head-mounted display devices, for example, optical see-through devices like Google Glass~\cite{googleGlass} and HoloLens~\cite{hololens} are relatively mature AR devices in terms of a combined hardware and software solution. Such technology typically demands robust, accurate and real-time semantic scene understanding. Like autonomous vehicles, much progress has been made by making environment-specific scope assumptions.

\subsection{Critical Upcoming Enhancers}

Current capabilities in areas like scene understanding are in part due to advances that have occurred in parallel to research developments, for example, better hardware that led to the use of GPUs and the availability of large-scale datasets. Similarly, further progress in semantic mapping and representations will rely on some of the upcoming enablers and enhancers that will facilitate the development of systems capable of high-level semantic reasoning. These enablers include better computational resources and architectures at both hardware and software level; effective and more common usage of knowledge databases and the ability to better leverage the diverse range of existing sensor data; advances in the field of online, networked and cloud robotics and Internet of Things (IoT); and novel mechanisms for enabling human-robot interaction involving uncommon sensing and engaging capabilities. These aforementioned technologies are active areas of research and in the following sections, we highlight how they have been incorporated or potentially could be integrated in robotics to enable semantic reasoning, ultimately leading to more scalable, sophisticated and robust robotic operations.

\subsubsection{Computational Resources and Architectures}

The growing computational needs of deep learning-based techniques, including those developed for semantic scene understanding, have led to rapid development of improved hardware and computational resources. Notably, GPUs continue to advance, with recent hardware changes being increasingly focused on facilitating deep learning and related techniques, rather than just catering to the traditional consumer gaming, media production and modelling users. Key performance properties relevant to deep learning include memory bandwidth, total memory capacity, clock frequency, and architectural considerations that improve typical operations like inference. All these properties continue to advance rapidly from year to year. \cite{davison2018futuremapping} highlighted the gap between the performance requirements of embodied devices and their practical constraints, indicating the need for co-design of algorithms, processors and sensors. The high compute requirements of visual SLAM and 3D reconstruction algorithms often make extensive evaluation difficult, especially for typical university research labs, so evaluation is often limited to qualitative visualizations and accuracy estimation on a few datasets. This trend deters the uncovering of design choices at algorithm level that could potentially lead to better trade-offs between accuracy, compute time, power consumption and the quality of output. Hence, better design space exploration strategies are continuously being explored that involve both hardware and software, particularly taking into account the latter's low-level design~\cite{zia2016comparative, bodin2016integrating, nardi2017algorithmic, saeedi2018navigating, nardi2019practical}.

When optimizing for computational and power performance, data communication overheads (sensor to processor and processor-memory) play a major role by adding latency. \cite{dudek2005general} introduced a processor-per-pixel arrangement-based vision chip~\cite{moini1999vision} for low-level image processing tasks which are inherently pixel-parallel in nature. Such vision chips ensure that data is processed adjacent to the sensor, thus reducing data transfer costs. These vision chips belong to the class of Focal-Plane Sensor-Processor (FPSP) chips~\cite{zarandy2011focal}. For example, the SCAMP-5 vision chip~\cite{carey2013low} operates on image-wide register arrays enabling data (full image) transfer from sensor to processor array in one clock cycle (100 ns), delivering 655 GOPS at 1.2 W power consumption~\cite{martel2016vision}. Recently, \cite{wong2018analog} demonstrated the use of this chip with a novel energy-efficient CNN, resulting in significant gains in compute time (85\%) and energy (84\%) efficiency, while achieving high accuracy (90\%) at a very high frame rate (3000 fps) for a handwritten digit recognition task. 

Besides designing FPSP chips, neuroscience-driven vision chips~\cite{mead1990neuromorphic, aizawa1999computational} and the use of Field Programmable Gate Arrays (FPGAs) is increasingly becoming popular as a means to accelerate deep learning-based applications, as has been reviewed recently~\cite{ruiz2019field, mittal2020survey, blaiech2019survey, wang2018survey, shawahna2018fpga, venieris2018toolflows}. Neuromorphic vision chips have been discussed in detail in ~\cite{koch1996neuromorphic} and more recently in ~\cite{wu2018neuromorphic}, where the latter discussed both frame-driven and event-driven vision chips. \cite{schuman2017survey} comprehensively reviewed the neuromorphic computers, devices and model architectures that at some level model neuroscience to solve challenging machine learning problems. Neuromorphic computing provides opportunities to implement neural networks-based applications, particularly those based on Spiking Neural Networks (SNNs), directly in the hardware, thus offering significant advantages in terms of power, compute time and storage footprint. This hardware is typically implemented as FPGAs~\cite{tapiador2018event} or an Application-Specific Integrated Circuit (ASIC)~\cite{cruz2012energy}. Robotic applications, especially those requiring autonomous navigation competencies, have been explored with the use of spiking neural networks~\cite{tang2019spiking, huang2019flyintel, beyeler2015gpu, stewart2016serendipitous}. Furthermore, existing CNNs can be converted into SNNs and mapped to spike-based hardware architectures while retaining their original accuracy levels, as demonstrated on the object recognition task in~\cite{sengupta2019going, xing2019homeostasis, cao2015spiking}. This opens up opportunities to better leverage advances in these hardware platforms while also motivating researchers to bridge the gap between the state of the art in learning using ANNs and SNNs.

\subsubsection{Knowledge Repositories and Databases}

The availability of knowledge repositories, datasets and benchmarks has always proven to be an accelerator of research in both the computer vision and robotics community. While knowledge databases help in providing contextual information and designing semantically-informed systems, sensor data in the form of images, videos, depth and inertia help in developing data-driven learning-based systems.

With deep learning as such an integral component of most semantics-related research, it has become even more important to obtain both accurate and large-scale labeled data to enable \textit{deep} and \textit{supervised} learning, for example, in the case of object detection and semantic segmentation. The recent focus has also been on creating equivalent benchmark datasets for 3D semantic scene understanding. In Table~\ref{tab:datasets}, we list a sampling of datasets that are targeted at enabling semantic understanding by solving diverse tasks, for example, object detection in 2D~\cite{Song2015, neuhold2017mapillary} and 3D~\cite{Hua2016, geyer2020a2d2}, semantic segmentation in 2D~\cite{Cordts2016} and 3D~\cite{Silberman2012, chang2017matterport3d}, semantic place categorization~\cite{Zhou2018}, robotic manipulation~\cite{Aksoy2015a}, road semantics~\cite{wang2016torontocity}, multi-spectral semantics~\cite{Valada2017}, UAV-view semantics~\cite{nigam2018ensemble}, 3D object shapes~\cite{wu20153d}, multi-object tracking~\cite{gaidon2016virtual}, pedestrian intention prediction~\cite{liu2020spatiotemporal}, pedestrian locomotion forecasting~\cite{mangalam2020disentangling}, collision-free space detection (drivable vs non-drivable)~\cite{fan2020sne}, Near-InfraRed segmentation~\cite{pandey2020unsupervised}, and semantic mapping~\cite{Ruiz-Sarmiento2017}. This list of datasets is by no means comprehensive, and more details can be found in recent dataset review articles~\cite{feng2020deep, lateef2019survey, yu2018methods, yin2017use} and online~\cite{awesomeSemanticSegmentation}, although any single source does not provide a complete list, emphasizing the disconnect between the use of semantics for different tasks and applications. Table~\ref{tab:datasets} mainly highlights the diversity in semantic understanding tasks, which are typically related to each other but often evaluated in isolation. It suggests that any future general semantic capability for robots might also require a more unified approach to the development and use of these datasets~\cite{wang2016torontocity}.

\begin{table*}
\centering
\caption{Diversity in Datasets with Semantic Annotations}

\begin{tabular}{lp{0.3\textwidth}p{0.3\textwidth}}
\toprule
\textbf{Name} & \textbf{Annotation Type} & \textbf{Additional Information}\\
\midrule

NYUv2 (2012)~\cite{Silberman2012} & Dense RGB-D and object instance-wise & 1449 aligned RGB-D pairs from 464 indoor scene sequences \\

RueMonge (2014)~\cite{Riemenschneider2014} & Pixel-wise labels along with a 3D Mesh & 1 million 3D points outdoor forming several Haussmanian-style facades and 8 semantic classes. 
\\

MANIAC (2015)~\cite{Aksoy2015a} & Semantic annotations of manipulation activities & 8 unique activities, manipulating 30 objects \\

SUN RGB-D (2015)~\cite{Song2015} & Polygons and bounding boxes with object orientations (RGB-D) and scene category & 10,000 indoor RGB-D images with 146,617 2D polygons and 58,657 3D bounding boxes.
\\

SUNCG (2016)~\cite{Song2016} & Volumetric and object-level labels for synthetic indoor scenes & 45,000 scenes with realistic room and furniture layouts. \\

SceneNN (2016)~\cite{Hua2016} & Scene meshes labeled per-vertex and per-pixel with object poses and bounding boxes & 100 indoor scenes \\

Cityscapes (2016)~\cite{Cordts2016} & Pixel- and instance-wise labels & 5,000 fine and 20,000 coarse annotations for 30 semantic classes from outdoor scenes across 50 cities. \\

ScanNet (2017)~\cite{Dai2017} & 3D camera poses and dense and object instance-level labeled surface reconstructions (RGB-D) & 2.5M views in 1513 indoor scene sequences \\

SEMANTIC3D.NET (2017)~\cite{Hackel2017} & Labeled 3D point clouds & 4 billion manually labeled points and 8 semantic classes from urban outdoor scenes. \\

Freiburg Forest (2017)~\cite{Valada2017} & Multispectral and multimodal pixel-level labels & Multiple 4.7 km traverses with 6 outdoor semantic classes \\

Stanford 2D-3D-Semantics (2017)~\cite{Armeni2017} & Pixel- and instance-level labels with 3D meshes and point clouds (RGB-D) & Approximately 70,000 images and 695 million points from large-scale indoor spaces (6,000 m$^2$) \\

Places (2017)~\cite{Zhou2018} & Place-level labels & 10 million images with 400+ semantic place categories. \\

Robot@Home (2017)~\cite{Ruiz-Sarmiento2017} & Room-, object- and point-level labels of 3D indoor reconstructions (RGB-D) & 36 rooms (8 categories) and 1900 objects (57 categories) observed in 69,000 images captured from traverses of a mobile robot at home. \\

PrOD (2020)~\cite{hall2020probabilistic} & Pixel- and object-level labels and bounding boxes for synthetic images & Indoor image sequences captured from multiple robots in a domestic environment.\\

A2D2 (2020)~\cite{geyer2020a2d2} & Pixel- and object-level labels and 3D bounding boxes (RGB-D) & 41,280 frames with 38 semantic classes captured from multiple outdoor traverses.\\
\bottomrule
\end{tabular}
\label{tab:datasets} 
\end{table*}


Semantic-based approaches typically rely on pre-learnt material from carefully curated datasets. As the range, size and variety of these datasets improves, so likely will the outcomes. An alternative means of incorporating prior information is to make use of a pre-defined ontology and common knowledge with the help of knowledge graphs~\cite{wang2017knowledge, paulheim2017knowledge} and ontology languages~\cite{mcguinness2004owl, martin2004owl, janowicz2019sosa}.

In the context of robotics, several researchers have explored different ways of encoding the relevant knowledge and proposed a corresponding ontology. As highlighted in~\cite{prestes2013towards}, these ontologies have been applied in different ways: path planning and navigation~\cite{schlenoff2003using, habib1993map, balakirsky2004knowledge}, describing an environment~\cite{chella2002modeling}, task analysis for autonomous vehicles~\cite{barbera2004task}, task-oriented conceptualization~\cite{wood2004representation}, as meta-knowledge for learning from heuristics of a system~\cite{epstein2004metaknowledge}, policies to govern behavior~\cite{jung2004ontology}, describing robots and their capabilities~\cite{schlenoff2005robot}, control architecture concepts~\cite{dhouib2011control}, and characterizing sub-domains within robotics~\cite{hallam2006ontology}. Furthermore, given the ways of representing knowledge~\cite{davis1993knowledge, wood2012review, fischer2018dac}, its management~\cite{lemaignan2010oro} and processing~\cite{tenorth2009knowrob} frameworks have also been explored. Recently, \cite{gayathri2018ontology} reviewed ontology-based knowledge representations for robotic path planning.

The use of knowledge representation and ontology has been explored in semantic mapping for robots. \cite{galindo2005multi} proposed an ``anchoring'' process to link the spatial and semantic information hierarchies for a semantic map representation. For representing knowledge, authors used the NeoClassic~\cite{patel1996neoclassic} system to encode the semantic (conceptual) information hierarchy. \cite{zender2008conceptual} presented a hierarchical map representation where the highest level of abstraction forms the conceptual layer, representing knowledge in the form of abstract concepts (TBox with terminological knowledge) and instances of such concepts (ABox with assertion knowledge). Using the OWL-DL ontology of an indoor office environment, the conceptual map is defined using taxonomies of room types and objects found therein in the form of \textit{is-a} and \textit{has-a} relations. Beyond considering just the objects in the space, \cite{nuchter2008towards} implemented a constrained network solver using Prolog to encode the environment and in particular, the relationships between different planes (walls, floor, ceiling) as being parallel or orthogonal. Using the knowledge processing system, KnowRob~\cite{tenorth2009knowrob}, \cite{tenorth2010knowrob} presented a semantic mapping system, KnowRob-Map, that links objects and spatial information with two sources of knowledge: \textit{encyclopedic} knowledge, inspired by OpenCyc~\cite{lenat1995cyc}, comprising object classes and their inter-relation hierarchies and \textit{common-sense} knowledge, based on Open Mind Indoor Common Sense (OMICS)~\cite{gupta2004common}, comprising action-related knowledge about everyday objects. Riazuelo \textit{et al}.~\cite{Riazuelo2015} demonstrated the benefits of combining SLAM and knowledge-based reasoning through RoboEarth~\cite{mohanarajah2014rapyuta}, a semantic mapping system. In the proposed system, the RoboEarth knowledge base stores ``action recipes'' (hardware- and environment-agnostic abstract descriptions of a task), sets of object models, and robots' models in Semantic Robot Description Language (SRDL)~\cite{kunze2011towards}. For a requested action recipe (e.g. exploration), based on the robot's capabilities, an execution plan tailored to the robot is generated on the cloud. For plan execution, the robot downloads a set of object models expected in its environment for building a semantic map, which can be uploaded to RoboEarth for future use by other robots. Using the proposed system, the authors demonstrated two tasks: semantic mapping of a new environment and novel object searching based on semantic reasoning.

Addressing the gap between the use of a pre-defined knowledge database and the ability to gain new knowledge, \cite{ruiz2017building} presented the concept of a ``multiversal'' semantic map based on probabilistic symbol grounding. This approach enables reasoning through multiple ontological interpretations of a robot's workspace. \cite{kulvicius2013semantic} used additional linguistic cues for text-based image retrieval, showcasing the potential of using knowledge databases to improve retrieval accuracy. Beyond entity-level knowledge graphs, \cite{alemu2020healthaid} proposed HealthAidKB, a knowledge base with 71000 task frames structured hierarchically and categorically, mainly focused on procedural knowledge for queries that require a step-by-step solution to a problem at hand. Using Probabilistic Action Templates (PATs)~\cite{leidner2012things, younes2004ppddl1}, \cite{bauer2020probabilistic} developed a probabilistic effect prediction method based on semantic knowledge and physical simulation to predict a robot's action success. \cite{kulvicius2013semantic, Riazuelo2015} and a body of similar work~\cite{tenorth2011web, beetz2011robotic, tamosiunaite2011generalizing} attempt to achieve generalization and solve the open-set problem that persists especially in visual perception systems. Using prior knowledge and learning from various new knowledge sources is likely to be key to further advances in the semantic reasoning capability of robots.

\subsubsection{Online and the Cloud}

Cloud robotics~\cite{kuffner2010cloud} or online~\cite{goldberg2002beyond} and networked~\cite{goldberg1995desktop, mckee2008networked} robotics mainly refers to the use of internet or local communication networks to share resources and computation between multiple robots or applications. It is also closely related to the concept of Internet-of-Things (IoT)~\cite{davidson2004open, ashton2009internet, atzori2010internet}. Goldberg and Kehoe~\cite{goldberg2013cloud} reviewed five different ways in which cloud robotics and automation can be put to use: shared access to annotated sensor data, on-demand massively-parallel computational support, sharing of trial outcomes for collective learning, provision of open-source and open-access code, data, and hardware design, and on-demand human guidance for task support - all areas of relevance to semantics. Furthermore, \cite{kehoe2015survey} presented a concept of Robotics and Automation as a Service (RAaaS) that combines the ideas of Infrastructure as a Service (IaaS), Software as a Service (Saas) and Platform as a Service (PaaS). Wan \textit{et al}.~\cite{wan2016cloud} reviewed the development process of cloud robotics and its potential value in different applications: SLAM, navigation and grasping. \cite{hu2012cloud} designed a cloud robotics architecture comprised of two sub-systems: a communication framework (machine-to-machine and machine-to-cloud) and an elastic computing architecture. The latter is defined using three different models: peer-based, proxy-based and clone-based, where their suitability to a particular robotic application is based on key characteristics like robustness in network connections, interoperability and flexibility for mobility within the network. The authors also highlight the key challenges for such a cloud system in terms of computation, communication, optimization and security. Under the constraints of limited communication, \cite{carlone2010rao} presented a multi-robot SLAM system based on Rao-Blackwellized Particle Filters~\cite{doucet2000rao} that required only a small amount of data to be exchanged across robots. A number of recent articles have reviewed the current trends in cloud robotics~\cite{chowdhury2020approach, saha2018comprehensive, chen2018study, koubaa2019service, chinchali2019network, fosch2019cloud, botta2019cloud}. Saha \textit{et al}.~\cite{saha2018comprehensive} highlighted the increasing range of applications for cloud robotics in various areas: manufacturing, social, agriculture, medical and disaster management. 

Visual SLAM approaches can benefit significantly from an online and cloud based system. The Parallel Tracking and Mapping (PTAM) framework~\cite{klein2007parallel} developed by Klein and Murray has been one of the landmark works in the modern history of visual SLAM. It demonstrated the advantages of alternation and parallel computing (of tracking and mapping), leading to a robust real-time system with accuracy comparable to offline reconstruction. A number of present state-of-the-art visual SLAM and 3D reconstruction systems use similar principles~\cite{mur-atal15, bloesch2018codeslam, newcombe2011dtam, newcombe11kinectfusion}. However, the scale at which such a framework can be used is directly related to the compute resources and storage available on the embodied device, as well as the possibility of robot collaboration. Hence, online and cloud-based alternatives have been explored in the literature aimed at widely expanding the scope of using robots for various tasks. In~\cite{riazuelo2014c2tam}, the authors developed a Cloud framework for Cooperative Tracking and Mapping (C$^2$TAM) as an extension to PTAM~\cite{klein2007parallel}. In this framework place recognition and non-linear map optimization is performed on the cloud, while tracking and re-localization is handled by the robot client. A cloud-based framework not only provides computational benefits but also the opportunity to share work tasks and knowledge, for example, in multi-robot exploration and mapping~\cite{zhang2018cloud, choudhary2016multi, kim2010multiple, fox2006distributed}.

Some recent advances in the context of cloud computing and its use in robotics include reinforcement learning based resource allocation~\cite{liu2018reinforcement}, real-time object tracking over the internet using UAVs~\cite{koubaa2018dronetrack, koubaa2019dronemap}, FastSLAM 2.0~\cite{montemerlo2003fastslam} as a cloud service~\cite{ali2018fastslam} and cloud-based real-time multi-robot SLAM~\cite{zhang2018cloud}. One of the critical enablers of cloud robotics is network offloading where a number of challenges still remain to be solved~\cite{chinchali2019network}. In the context of robotic tasks based on semantic scene understanding, researchers have explored a number of ways to use a cloud-based paradigm. \cite{wu2019cloud} demonstrated the use of semantic databases and sub-databases on the the CloudStack~\cite{cloudstack}), comprised of text-based sub-database selection (tableware versus electric appliances) and vision-based representations of various objects. A semantic map of the environment is then constructed using ``belonging-annotation'' positional relationships between objects. In an industrial automation application, \cite{martin2017decentralized} proposed a decentralized robot-cloud communication architecture for autonomous transportation within a factory, particularly highlighting the benefits of their approach under communication and hardware failure conditions. More recently, \cite{ballotta2020sensor} presented an analytical optimization of the trade-off between communication and computational delay in a network of homogeneous sensors.

The knowledge repositories and relevant datasets discussed previously also lead to the use of online cloud services in order to improve knowledge sharing and enabling learning from other similar resources. The use of cloud services also means that a lot of heavy-lifting tasks can be done on the cloud including incremental learning~\cite{roy2020tree, zhang2020class}, large-scale retrieval~\cite{garg2020fast, chang2020pre}, intensive data association~\cite{ye2020efficient, chen2020graph} and non-convex optimization~\cite{yang2020one, yang2020graduated}. Furthermore, cloud infrastructure brings robots within the IoT and also enables effective collaborative task solving that involves multiple robots meaningfully interacting with each other and other machines or humans. This approach may, eventually, lead to solving literally city-scale complexity problems, like building and administering a smart city~\cite{kapitonov2019robotic, kim2017smart}.

\subsubsection{Human-robot interaction}

One of the key aims of equipping robots with rich semantic representations of the world is to enhance and naturalize human-robot interaction. The literature points towards the use of semantics to build expressive robots~\cite{breazeal2009role} that can use effective modalities to coordinate their verbal and non-verbal interaction with humans. This includes human-like motion of robotic arms~\cite{Fang2019}, semantic reasoning-based natural language understanding for service robots~\cite{Savage2019}, human-robot dancing~\cite{or2009towards,Jochum2019TonightWI}, expression of emotions (both facial and full body~\cite{miwa2004effective,Lai2018EmotionPreservingRL,giambattista2016expression}), and the ability to build and navigate in human-centric semantic spatial representations~\cite{zender2008conceptual,talbot2020robot}. More recently, \cite{moon2020object} proposed object-oriented semantic graphs based on a graph convolutional network to generate natural questions from the observed scene. Linked with inverse semantics~\cite{Tellex2015}, such systems enable a robot to ask the right questions in order to better communicate, ask for help and solve the task accordingly.

\section{DISCUSSION AND CONCLUSION}
\label{secn:conclusion}
Robotic perception, world modelling, and decision making have evolved beyond their early limited focus on geometry and appearance. As we have discussed, modern approaches increasingly incorporate semantic information, which enables a higher-level and richer understanding of the world. In return, a variety of new robotic applications have already emerged, with new applications on the horizon.

Still, many exciting directions for research remain. One open questions is about explicit or implicit representation of semantics: should we as researchers and algorithm designers \emph{enforce} semantic information being explicitly represented, or do we enable an algorithm to \emph{implicitly} learn task-relevant semantic concepts? Do we understand which semantic concepts are relevant for a particular robotic task, and is there a direct mapping between robot-relevant and human-relevant semantic concepts? In an age where explainability and the ``trustworthiness'' of autonomous systems is becoming increasingly important, how could we understand and interpret robot-learnt semantic concepts that bear no direct correspondence to any semantic concepts we are familiar with as humans? While it would be ``neat" if the concepts are the same across robots and humans, if optimal robot performance involves semantic learning and representations that aren't directly interpretable by humans, this is likely to be an ongoing area of research focus.

At the moment, most semantic representations assume a \emph{flat} structure. However, many semantic concepts can be naturally organised in a \emph{hierarchical} structure (or even into a more general, graph-like structure). This is maybe most obvious for object class labels (\texttt{object} $\rightarrow$ \texttt{indoor object} $\rightarrow$ \texttt{furniture} $\rightarrow$ \texttt{chair}), but might also be a useful representation for affordances, room or place categories, and other semantic domains. This hierarchical structure also extends spatially, where semantic concepts might be expressed on the scale of object parts, objects, functional ensembles of objects, rooms, buildings and city blocks. Hierarchical semantic knowledge is likely not to be a ``clean'' representation, but rather to involve levels where the distinction between levels can be quite fuzzy; take for example a barista's coffee machine, which is both a single, complex object and an ensemble of objects. Understanding the different aspects of a semantic hierarchy can be important for robotic applications, especially when faced with imperfect perception or incomplete knowledge. 

This discussion soon leads to more principled questions about the very nature of semantic concepts and how they can be represented to be accessible by robotic algorithms. Unless one follows the promises of pure and all-encompassing end-to-end learning, prior knowledge about semantic concepts needs to be modeled and represented in some form. This has already been investigated extensively in classical AI knowledge representation and reasoning research. An interesting middle ground is to try and incorporate hand-crafted semantics as priors into learning-based systems that can expand, continue to learn, or even re-learn semantics in a task-informed way. Given the wide range of potential applications of robotics with varying levels of complexity and operating requirements (especially around safety and reliability), a spectrum of approaches may be needed, informed by context; from pure end-to-end to hybrid, to entirely hand-crafted.

As noted in the survey's coverage of datasets relating to semantics research, there can be significant disconnects between different research subfields that touch on semantics. To achieve sufficient progress in application-focused areas like autonomous vehicles, much of the work in semantics has focused on domain-specific implementations, leading to higher performance levels but at the possible cost of generality for all of robotics. It will be interesting to see to what extent the insights gained from targeted semantics research in an area like autonomous driving will benefit the robotics field more broadly, just as it will be interesting to see whether it is possible to make enough progress by focusing on providing robots - or in this case autonomous vehicles - with only a targeted, constrained understanding of the world around them.

Like many other topics, progress in semantics-related research suffers to some extent from a disconnect across disciplinary boundaries, especially between robotics and computer vision. A related issue is the dominance of dataset-based evaluation, especially for tasks like semantic segmentation. Papers at the leading conferences, arguably predominantly in the computer vision discipline at conferences like CVPR, are dominated by research that achieves new levels of performance on these benchmark datasets. Although there is a promising trend towards evaluation in high fidelity simulation environments, performance on datasets and simulation alone is only one step towards safe and reliable deployment on robotic platforms, as has been revealed in application areas such as autonomous on-road vehicles. Bridging the divide between performance on datasets and performance on robots in closed-loop scenarios is likely to remain a major challenge for the foreseeable future, but also presents a unique opportunity for what we might coin ``active" semantics, where semantic learning and understanding are enhanced by active control of a robotic platform and its sensing modalities.

This survey has summarized the state-of-play with respect to semantics research: the fundamentals, and the increasing integration of semantics into systems addressing key robotic capabilities like mapping and interaction with the world. While much progress has been made in the quest to imbue robots with a richer and nuanced understanding of the world around them, there is much still to be done. Future research will benefit from developments in the technologies and datasets underpinning much semantics-based research, as well as from new conceptual approaches, including those discussed here. The use of semantics in robotics will also continue to be informed by humans. Human communication makes rich use of semantic concepts of various kinds; we formulate tasks, give instructions and feedback, and communicate expectations based on the \emph{meaning} of objects, affordances, or the wider spatial and temporal context. Incorporating semantics into robotics, especially by bridging classical approaches with modern, learning-based approaches could have great impact on future robotic applications, especially those where robots work closely with, for or around humans.


\bibliographystyle{IEEEtran}
\bibliography{megaRef.bib}

\pagebreak

\end{document}